\documentclass{article}

\usepackage[final]{neurips_2019}

\usepackage[utf8]{inputenc} %
\usepackage[T1]{fontenc}    %
\usepackage{hyperref}       %
\usepackage{url}            %
\usepackage{booktabs}       %
\usepackage{amsfonts}       %
\usepackage{nicefrac}       %
\usepackage{microtype}      %

\usepackage{amsmath,amssymb,amsthm}
\usepackage{amssymb,amsbsy}
\usepackage{mathtools}
\usepackage{bbm}
\usepackage{graphicx,xcolor}
\usepackage{cleveref}
\usepackage{subcaption}

\usepackage{algorithm, algorithmic}

\DeclareMathOperator*{\argmax}{arg\,max}
\DeclareMathOperator*{\argmin}{arg\,min}

\newtheorem{lem}{Lemma}
\newtheorem{prop}{Proposition}
\newtheorem{assumption}{Assumption}

\def\K{\mathcal{K}}

\def\D{\mathcal{D}}
\def\tr{\mathop{\text{tr}}\kern.2ex}

\def\E{{\mathbb E}}

\newcommand{\refp}[1]{(\ref{#1})}

\newcommand{\red}[1]{{#1}}

\newcommand{\norm}[1]{\left\|{#1}\right\|}

\title{Divergence-Augmented Policy Optimization}

\author{
    Qing Wang\thanks{The work was done when Qing Wang was at Tencent AI Lab. $\dagger$ Equal contribution.}~~$\dagger$ \\
    Huya AI \\
    Guangzhou, China \\
    \And
    Yingru Li$\dagger$ \\
    The Chinese University of Hong Kong \\
    Shenzhen, China \\
    \And 
    Jiechao Xiong \\
    Tencent AI Lab \\
    Shenzhen, China \\
    \And
    Tong Zhang \\
    The Hong Kong University of Science and Technology \\
    Hong Kong, China \\
}

\begin{document}

\maketitle

\begin{abstract}

In deep reinforcement learning, policy optimization methods need to deal with issues such as function approximation and the reuse of off-policy data. Standard policy gradient methods do not handle off-policy data well, leading to premature convergence and instability. This paper introduces a method to stabilize policy optimization when off-policy data are reused. The idea is to include a Bregman divergence between the behavior policy that generates the data and the current policy to
\red{ensure small and safe policy updates with off-policy data.}
\red{The Bregman divergence is calculated between the state distributions of two policies, instead of only on the action probabilities, leading to a divergence augmentation formulation.}
Empirical experiments on Atari games show that in the data-scarce scenario where the reuse of off-policy data becomes necessary, our method can achieve better performance than other state-of-the-art deep reinforcement learning algorithms.
\end{abstract}

\section{Introduction}\label{sec:intro}

In recent years, many algorithms based on policy optimization have been proposed for deep reinforcement learning (DRL), leading to great successes in Go, video games, and robotics \citep{silver2016mastering,mnih2016asynchronous, schulman2015trust, schulman2017proximal}.
Real-world applications of policy-based methods commonly involve function approximation and data reuse. Typically, the reused data are generated with an earlier version of the policy, leading to off-policy learning.
It is known that these issues may cause premature convergence and instability for policy gradient methods \citep{sutton2000policy,sutton2017introduction}.

A standard technique that allows policy optimization methods to handle off-policy data is to use importance sampling to correct trajectories from the behavior policy that generates the data to the target policy  (e.g.~Retrace \citep{munos2016safe} and V-trace \citep{espeholt2018scalable}). The efficiency of these methods depends on the divergence between the behavior policy and the target policy.
Moreover, to improve stability of training, one may introduce a regularization term (e.g.~Shannon-Gibbs entropy in \citep{mnih2016asynchronous}), or use a proximal objective of the original policy gradient loss (e.g.~clipping in  \citep{schulman2017proximal,wang2016sample}).
Although the well-adopted method of entropy regularization can stabilize the optimization process \citep{mnih2016asynchronous}, this additional entropy regularization alters the learning objective, and prevent the algorithm from converging to the optimal action for each state. Even for the simple case of bandit problems, the monotonic diminishing regularization may fail to converge to the best arm \citep{cesa2017boltzmann}.

In this work, we propose a method for policy optimization by adding a Bregman divergence term, which leads to more stable and sample efficient off-policy learning. 
\red{The Bregman divergence constraint is widely used to explore and exploit optimally in mirror descent methods \citep{nemirovsky1983problem}, in which specific form of divergence can attain the optimal rate of regret (sample efficiency) for bandit problems \citep{audibert2011minimax,bubeck2012regret}.}
In contrast to the traditional approach of constraining the divergence between target policy and behavior policy conditioned on each state \citep{schulman2015trust}, we consider the divergence over the joint state-action space. We show that the policy optimization problem with Bregman divergence on state-action space is equivalent to the standard policy gradient method with divergence-augmented advantage. Under this view, the divergence-augmented policy optimization method not only considers the divergence
on the current state but also takes into account the discrepancy of policies on future states, thus can provide a better constraint on the change of policy and encourage ``deeper'' exploration.

We experiment with the proposed method on the commonly used Atari 2600 environment from Arcade Learning Environment (ALE) \citep{bellemare2013arcade}. Empirical results show that divergence-augmented policy optimization method performs better than the state-of-the-art algorithm under data-scarce scenarios, i.e., when the sample generating speed is limited and samples in replay memory are reused multiple times. We also conduct a comparative study for the major effect of improvement on these games.

The article is organized as follows: we give the basic background and notations in Section \ref{sec:preliminaries}. The main method of divergence-augmented policy optimization is presented in Section \ref{sec:method}, with connections to previous works discussed in Section \ref{sec:related}.
Empirical results and studies can be found in Section \ref{sec:experiments}. We conclude this work with a short discussion in Section \ref{sec:conclusion}.

\section{Preliminaries}\label{sec:preliminaries}

In this section, we state the basic definition of the Markov decision process considered in this work, as well as the Bregman divergence used in the following discussions.

\subsection{Markov Decision Process}

We consider a Markov decision process (MDP) with infinite-horizon and discounted reward,
denoted by $\mathcal{M} = (\mathcal{S},\mathcal{A},P,r,d_0,\gamma)$,
where $\mathcal{S}$ is the finite state space, $\mathcal{A}$ is the finite action space,
$P: \mathcal{S}\times\mathcal{A} \rightarrow \Delta(\mathcal{S})$ is the transition function, where $\Delta(\mathcal{S})$ means the space of all probability distributions on $\mathcal{S}$. A reward function is denoted by $r:\mathcal{S}\times\mathcal{A} \rightarrow \mathbb{R}$.
The distribution of initial state $s_0$ is denoted by $d_0 \in \Delta(\mathcal{S})$. And a discount factor is denoted by $\gamma \in (0,1)$. %

A stochastic policy is denoted by $\pi: \mathcal{S} \rightarrow \Delta(\mathcal{A})$.
The space of all policies is denoted by $\Pi$.
We use the following standard notation of state-value $V^\pi(s_t)$, action-value $Q^\pi(s_t, a_t)$ and advantage $A^\pi(s_t, a_t)$, defined as
$V^\pi(s_t) = \E_{\pi|s_t}\sum_{l=0}^\infty \gamma^l r(s_{t+l},a_{t+l})$,
$Q^\pi(s_t,a_t) = \E_{\pi|s_t,a_t}\sum_{l=0}^\infty \gamma^l r(s_{t+l},a_{t+l})$, and
$A^\pi(s_t,a_t) = Q^\pi(s_t,a_t) - V^\pi(s_t)$,
where $\E_{\pi|s_t}$ means $a_l \sim \pi(a | s_l)$,
$s_{l+1} \sim P(s_{l+1} | s_l, a_l)$, $\forall l \ge t$, and
$\E_{\pi|s_t,a_t}$ means $s_{l+1} \sim P(s_{l+1} | s_l, a_l)$,
$a_{l+1} \sim \pi(a | s_{l+1})$, $\forall l \ge t$.
We also define the space of policy-induced state-action distributions under $\mathcal{M}$ as
\begin{align}\label{eq:delta_pi}
\Delta_\Pi = \{ & \mu \in \Delta(\mathcal{S}\times\mathcal{A}): \sum_{a'} \mu(s',a') = (1-\gamma) d_0(s') + \gamma \sum_{s,a} P(s'|s,a)\mu(s,a), \forall s' \in \mathcal{S} \}
\end{align}
We use the notation $\mu_\pi$ for the state-action distribution induced by $\pi$. On the other hand, for each $\mu \in \Delta_\Pi$, there also exists a unique policy $\pi_\mu(a|s) = \frac{\mu(s,a)}{\sum_b \mu(s,b)}$ %
which induces $\mu$. We define the state distribution $d_\pi$ as $d_\pi(s) = (1-\gamma) \E_{\tau|\pi}\sum_{t=0}^\infty \gamma^t \mathbf{1}(s_t=s)$. Then we have $\mu_\pi(s,a)$ = $d_\pi(s)\pi(a|s)$. We sometimes write $\pi_{\mu_t}$ as $\pi_t$ and $d_{\pi_t}$ as $d_t$ when there is no ambiguity.

In this paper, we mainly focus on the performance of a policy $\pi$ defined as
\begin{align}\label{eq:policy_perf}
J(\pi) = (1-\gamma) \E_{\tau|\pi}\sum_{t=0}^{\infty} \gamma^t r(s_t, a_t) = \E_{d_\pi, \pi} r(s,a)
\end{align}
where $\E_{\tau|\pi}$ means $s_0 \sim d_0, a_t \sim \pi(a_t|s_t), s_{t+1} \sim P(s_{t+1}|s_t, a_t), t\ge 0$. We use the notation $\E_{d,\pi} = \E_{s\sim d(\cdot), a\sim\pi(\cdot|s)}$
for brevity.

\subsection{Bregman Divergence}
We define Bregman divergence \citep{bregman1967relaxation} as follows (e.g.~Definition 5.3 in \citep{bubeck2012regret}). For $\D \subset \mathbb{R}^d$ an open convex set, the closure of $\D$ as $\bar{\D}$, we consider a \emph{Legendre} function $F : \bar{\D} \rightarrow \mathbb{R}$ defined as (1) $F$ is strictly convex and admits continuous first partial derivatives on $\D$, and (2) $\lim_{x\to\bar{\D}\setminus\D} \lVert \nabla F \rVert = +\infty$.
For function $F$, we define the Bregman divergence $D_F: \bar{\D}\times\D \rightarrow \mathbb{R}$ as
\[
D_F(x,y) = F(x) - F(y) - \langle \nabla F(y), x-y \rangle.
\]
The inner product is defined as $\langle x, y \rangle = \sum_i x_i y_i$.
For $\mathcal{K} \subset \bar{\D}$ and $\K\cap\D\ne\emptyset$, the Bregman projection
\[
z = \argmin_{x\in\mathcal{K}} D_F(x,y)
\]
exists uniquely for all $y\in\D$. Specifically, for $F(x) = \sum_i x_i \log(x_i) - \sum_i x_i$,
we recover the Kullback-Leibler (KL) divergence as
\begin{eqnarray*}
D_\text{KL}(\mu',\mu) &=& \sum_{s,a} \mu'(s,a) \log\frac{\mu'(s,a)}{\mu(s,a)} \\
\end{eqnarray*}
for $\mu, \mu' \in \Delta(\mathcal{S}\times\mathcal{A})$ and $\pi, \pi' \in \Pi$.
To measure the distance between two policies $\pi$ and $\pi'$, we also use the symbol for conditional ``Bregman divergence''\footnote{Note that $D_F^d$ may not be a Bregman divergence.} associated with state distribution $d$ denoted as
\begin{align}\label{eq:cond_bregman}
D_F^{d}(\pi', \pi) = \sum_s d(s) D_F(\pi'(\cdot|s), \pi(\cdot|s)).
\end{align}

\section{Method}\label{sec:method}
In this section, we present the proposed method from the motivation of mirror descent and then discuss the parametrization and off-policy correction we employed in the practical learning algorithm.

\subsection{Policy Optimization and Mirror Descent}

The mirror descent (MD) method \citep{nemirovsky1983problem} is a central topic in the optimization and online learning research literature. As a first-order method for optimization, the mirror descent method can recover several interesting algorithms discovered previously \citep{sutton2000policy,kakade2002natural,peters2010relative,schulman2015trust}. On the other hand, as an online learning method, the online (stochastic) mirror descent method can achieve (near-)optimal sample efficiency for a wide range of problems \citep{audibert2009minimax,audibert2011minimax,zimin2013online}.
In this work, following a series of previous works \citep{zimin2013online, neu2017unified}, we investigate the (online) mirror descent method for policy optimization. 
We denote the state-action distribution at iteration $t$ as $\mu_t$, and $\ell_t(\mu) = \langle g_t, \mu\rangle$ as the linear loss function for $\mu$ at iteration $t$. Without otherwise noted, we consider the negative reward as the loss objective $\ell_t(\mu) = -\langle r,\mu\rangle$, which also corresponds to the policy performance $\ell_t(\mu) \equiv -J(\pi_\mu)$ by Formula \refp{eq:policy_perf}. We consider the mirror map method associated with Legendre function $F$ as
\begin{eqnarray}
    \nabla F(\tilde{\mu}_{t+1}) &=& \nabla F(\mu_t) - \eta g_t \label{eq:mirrormap}\\
    \mu_{t+1} &\in& \Pi_{\Delta_\Pi}(\tilde{\mu}_{t+1}),
\end{eqnarray}
where $\tilde{\mu}_{t+1} \in \Delta(\mathcal{S}\times\mathcal{A})$ and $g_t = \nabla \ell_t(\mu_t)$.
It is well-known \citep{beck2003mirror} that an equivalent formulation of mirror map \refp{eq:mirrormap} is 
\begin{eqnarray}
\mu_{t+1} &=& \argmin_{\mu\in\Delta_\Pi} D_F(\mu, \tilde{\mu}_{t+1})\label{eq:mm} \\
&=& \argmin_{\mu\in\Delta_\Pi} D_F(\mu, \mu_t) + \eta \langle g_t, \mu \rangle,\label{eq:md}
\end{eqnarray}
The former formulation (\ref{eq:mm}) takes the view of non-linear sub-gradient projection in convex optimization,
while the later formulation (\ref{eq:md}) can be interpreted as a regularized optimization and is the usual definition of mirror descent \citep{nemirovsky1983problem, beck2003mirror, bubeck2015convex}. In this work, we will mostly investigate the approximate algorithm in the later formulation \refp{eq:md}.

\subsection{Parametric Policy-based Algorithm}
In the mirror descent view for policy optimization on state-action space as in Formula \refp{eq:md}, we need to compute the projection of $\mu$ onto the space of $\Delta_\Pi$. For the special case of KL-divergence on $\mu$, the sub-problem of finding minimum in \refp{eq:md} can be done efficiently, assuming the knowledge of transition function $P$ (See Proposition 1 in \citep{zimin2013online}). However, for a general divergence and real-world problems with unknown transition matrices, the projection in \refp{eq:md} is non-trivial to implement. In this section, we consider direct optimization in the (parametric) policy space without explicit projection. Specifically, we consider $\mu_\pi$ as a function of $\pi$, and $\pi$ parametrized as $\pi_\theta$. The Formula \refp{eq:md} can be written as
\begin{align}\label{eq:md-pi}
    \pi_{t+1} = \argmin_{\pi} D_{F}(\mu_\pi, \mu_t) + \eta \langle g_t, \mu_\pi \rangle.
\end{align}
Instead of solving globally, we approximate Formula \refp{eq:md-pi} with gradient descent on $\pi$. From the celebrated policy gradient theorem \citep{sutton2000policy}, we have the following lemma:
\begin{lem}(Policy Gradient Theorem \citep{sutton2000policy}) For $d_\pi$ and $\mu_\pi$ defined previously, the following equation holds for any state-action function $f: \mathcal{S}\times\mathcal{A} \rightarrow \mathbb{R}$:
\begin{align*}
\sum_{s,a} f(s,a) \nabla_\theta \mu_\pi(s,a) = \sum_{s,a} d_\pi(s) \mathcal{Q}^\pi(f)(s,a) \nabla_\theta \pi(a|s), %
\end{align*}
where $\mathcal{Q}^\pi$ is defined as an operator such that
\begin{align*}
\mathcal{Q}^\pi(f)(s,a) = \E_{\pi|s_t=s,a_t=a} \sum_{l=0}^\infty \gamma^l f(s_{t+l},a_{t+l}).
\end{align*}
\end{lem}
Decomposing the loss and divergence in two parts (\ref{eq:md-pi}), we have
\begin{align}
    \nabla_\theta \left\langle g_t, \mu_{\pi} \right\rangle = \left\langle d_\pi \mathcal{Q}^\pi(g_t), \nabla_\theta\pi(a|s)\right\rangle,\label{eq:q_g}
\end{align}
which is the usual policy gradient, and
\begin{align}
\nabla_\theta D_F(\mu_{\pi}, \mu_t) 
= \left\langle \nabla F(\mu_{\pi}) - \nabla F(\mu_t), \nabla_\theta \mu_\pi \right\rangle
= \left\langle d_\pi \mathcal{Q}^\pi\left(\nabla F(\mu_{\pi}) - \nabla F(\mu_t)\right), \nabla_\theta \pi(a|s) \right\rangle. \label{eq:md-pi2}
\end{align}
Similarly, we have the policy gradient for the conditional divergence (\ref{eq:cond_bregman}) as
\begin{align*}
\nabla_\theta D^{d_t}_F(\pi, \pi_t) =
\left\langle d_t (\nabla F(\pi) - \nabla F(\pi_t)), \nabla_\theta \pi(a|s) \right\rangle, %
\end{align*}
which does not have a discounted sum, since $d_t$ is fixed and independent of $\pi=\pi_\mu$.

\subsection{Off-policy Correction}

In this section, we discuss the practical method for estimating $\mathcal{Q}^\pi(f)$ under a behavior policy $\pi_t$. In distributed reinforcement learning with asynchronous gradient update, the policy $\pi_t$ which generated the trajectories may deviate from the policy $\pi_\theta$ currently being optimized. Thus off-policy correction is usually needed for the robustness of the algorithm (e.g.~V-trace as in IMPALA \citep{espeholt2018scalable}). Consider
\begin{align*}
\sum_{s,a} d_\pi(s) \mathcal{Q}^\pi(f)(s,a)\nabla_\theta \pi(a|s) 
&= \E_{(s,a)\sim\pi d_\pi}\mathcal{Q}^\pi(f)(s,a)\nabla_\theta \log\pi(a|s) \nonumber \\
&= \E_{(s,a)\sim\pi_t d_{\pi_t}}\frac{d_\pi(s)}{d_{\pi_t}(s)}\frac{\pi(a|s)}{\pi_t(a|s)}\mathcal{Q}^\pi(f)(s,a)\nabla_\theta \log\pi(a|s) %
\end{align*}
for $f = g_t$ or $f = \nabla F(\mu_\pi) - \nabla F(\mu_t)$. We would like to have an accurate estimation of $\mathcal{Q}^\pi(g_t)$ (\ref{eq:q_g}) and $\mathcal{Q}^\pi(\nabla F(\mu_\pi) - \nabla F(\mu_t))$ (\ref{eq:md-pi2}), and correct the deviation from $d_{\pi_t}$ to $d_\pi$ and $\pi_t$ to $\pi$.

For the estimation of $\mathcal{Q}^\pi(f)$ under a behavior policy $\pi_t$, possible methods include Retrace \citep{munos2016safe} providing an estimator of state-action value $\mathcal{Q}^\pi(f)$, and V-trace \citep{espeholt2018scalable} providing an estimator of state value $\E_{a\sim\pi} \mathcal{Q}^\pi(f) (s,a)$. In this work, we utilize the V-trace (Section 4.1 \citep{espeholt2018scalable}) estimation $v_{s_i} = v_i$ along a trajectory starting at $(s_i,a_i=s,a)$ under $\pi_t$. Details of multi-step Q-value estimation can be found in Appendix \ref{sec:exp_details}. With the value estimation $v_s$, the $\mathcal{Q}^\pi(g_t)$ is estimated with
\begin{align}
    \hat{A}_{s,a} = r_i + \gamma v_{i+1} - V_\theta(s_i). \label{eq:nstep-R}
\end{align}
We subtract a baseline $V_\theta(s_i)$ to reduce variance in estimation, as $\E_{\pi_t,d_t}\frac{\pi_\theta}{\pi_t}V_\theta(s)\nabla_\theta\log\pi_\theta = 0$.
For the estimation of $\mathcal{Q}^\pi(\nabla F(\mu_\pi) - \nabla F(\mu_t))$, we use the $n$-steps truncated importance sampling as
\begin{align}\label{eq:nstep-f}
    \hat{D}_{s,a} = f(s_i, a_i) + \sum_{j=1}^n \gamma^j (\prod_{k=0}^{j-1} c_{i+k})\rho_{i+j} f(s_{i+j}, a_{i+j}).
\end{align}
in which we use the notation $c_j = \min(\bar{c}_D,\frac{\pi_\theta(a_j|s_j)}{\pi_t(a_j|s_j)})$ and $\rho_j = \min(\bar{\rho}_D,\frac{\pi_\theta(a_j|s_j)}{\pi_t(a_j|s_j)})$.
The formula also corresponds to V-trace under the condition $V(\cdot)\equiv 0$. For RNN model trained on continuous roll-out samples, we set $n$ equals to the max-length till the end of roll-out.

For the correction of state distribution $d_\pi(s)/d_{\pi_t}(s)$, previous solutions include the use of emphatic algorithms as in \citep{sutton2016emphatic}, or through an estimate of state density ratio as in \citep{liu2018breaking}. However, in our experience, the estimation of density ratio will introduce additional error, which may lead to worse performance in practice. Therefore in this paper, we propose a different solution by restricting our attention to the correction of $\pi_t$ to $\pi$ via importance sampling and omitting the difference of $d_\pi/d_{\pi_t}$ in the algorithm. This introduces a bias in the gradient estimation, which we propose a new method to handle in this paper. Specifically, we show that although the omission of the state ratio introduces a bias in the gradient, the bias can be bounded by the regularization term of conditional KL divergence (see Appendix \ref{sec:analysis}). Therefore by explicitly adding an KL divergence regularization, we can effectively control the degree of off-policy bias caused by $d_\pi/d_{\pi_t}$ in that small regularization value implies a small bias. This approach naturally combines mirror descent with KL divergence regularization, leading to a more stable algorithm that is robust to off-policy data, as we will demonstrate by empirical experiments.

The final loss consists of the policy loss $L_\pi(\theta)$ and the value loss $L_v(\theta)$. To be specific, the gradient of policy loss is defined as
\begin{align}
\nabla_\theta L_\pi(\theta) = \E_{\pi_t,d_t}\frac{\pi}{\pi_t}(\hat{D}_{s,a}-\eta \hat{A}_{s,a})\nabla_\theta\log\pi. \label{eq:ploss} 
\end{align}
We can also use proximal methods like PPO \citep{schulman2017proximal} in conjunction with divergence augmentation. A practical implementation is elaborated later in Formula \refp{eq:ppo}.
In addition to the policy loss, we also update $V_\theta$ with value gradient defined as
\begin{align}
\nabla L_v(\theta) = \E_{\pi_t,d_t}\frac{\pi}{\pi_t} (V_\theta(s) - v_{s})\nabla_\theta V_\theta(s), \label{eq:vloss}
\end{align}
where $v_{s}=v_{s_i}$ is the multi-step value estimation with V-trace.
The parameter $\theta$ is then updated with a mixture of policy loss and value loss
\begin{align}
    \theta \leftarrow \theta - \alpha_{t}(\nabla_\theta L_\pi(\theta) + \red{b} \nabla_\theta L_v(\theta)), \label{eq:update}
\end{align}
in which $\alpha_{t}$ is the current learning rate, and \red{$b$} is the loss scaling coefficient.
The algorithm is summarized in Algorithm \ref{alg:dapo}. 

\begin{algorithm}[tb]
   \caption{Divergence-Augmented Policy Optimization (DAPO)}
   \label{alg:dapo}
\begin{algorithmic}
  \STATE {\bfseries Input: } $D_F(\mu', \mu)$, total iteration $T$, batch size $M$, learning rate $\alpha_{t}$.
  \STATE {\bfseries Initialize :} randomly initiate $\theta_0$
  \FOR{$t=0$ \TO $T$}
    \STATE (in parallel) Use $\pi_t = \pi_{\theta_t}$ to generate trajectories.
    \FOR{$m=1$ \TO $M$}
        \STATE Sample $(s_i,a_i) \in \mathcal{S}\times\mathcal{A}$ w.p.~$ d_t\pi_t$.
        \STATE Estimate state value $v_{s_i}$ (e.g.~by V-trace).
        \STATE Calculate Q-value estimation $\hat{A}_{s,a}$ \refp{eq:nstep-R} and divergence estimation $\hat{D}_{s,a}$ \refp{eq:nstep-f}.
        \STATE $\quad \hat{A}_{s,a} = r_i + \gamma v_{i+1} - V_\theta(s_i)$,
        \STATE $\quad \hat{D}_{s,a} = f(s_i, a_i) + \sum_{j=1}^n \gamma^j (\prod_{k=0}^{j-1} c_{i+k})\rho_{i+j} f(s_{i+j}, a_{i+j})$.
        \STATE Update $\theta$ with respect of policy loss (\ref{eq:ploss}, optionally \ref{eq:ppo}) and value loss \refp{eq:vloss}
        \STATE $\quad \theta \leftarrow \theta - \alpha_{t}(\nabla_\theta L_\pi(\theta) + \red{b} \nabla_\theta L_v(\theta)).$
    \ENDFOR
    \STATE Set $\theta_{t+1} = \theta$.
  \ENDFOR
\end{algorithmic}
\end{algorithm}

\section{Related Works}\label{sec:related}

The policy performance in Equation \refp{eq:policy_perf} and the well-known policy difference lemma \citep{kakade2002optimal} serve a fundamental role in policy-based reinforcement learning (e.g~TRPO, PPO \citep{schulman2015trust, schulman2017proximal}).
The gradient with respect to the policy performance and policy difference provides a natural direction for policy optimization. And to restrict the changes in each policy improvement step, as well as encouraging exploration at the early stage, the constraint-based policy optimization methods try to limit the changes in the policy by constraining the divergence between behavior policy and current policy. The use of entropy maximization in reinforcement learning can be dated back to the work of \citet{williams1991function}. And methods with relative entropy regularization include \citet{peters2010relative,schulman2015trust}.
The relationship between these methods and the mirror descent method has been discussed in \citet{neu2017unified}.
With notations in this work, consider the natural choice of $F$ as the \emph{negative Shannon entropy} defined as $F(x) = \sum_i x_i \log(x_i)$, the $D_F(\cdot,\cdot)$ becomes the KL-divergence $D_\text{KL}(\cdot,\cdot)$. By the equivalence of sub-gradient projection (\ref{eq:mm}) and mirror descent (\ref{eq:md}), the mirror descent policy optimization with KL-divergence can be written as
\begin{align}
\begin{aligned}
\mu_{t+1} &= \argmin_{\mu\in\Delta_\Pi} D_\text{KL}(\mu, \tilde{\mu}_{t+1}) = \argmin_{\mu\in\Delta_\Pi} D_\text{KL}(\mu, \mu_t) + \eta \langle g_t, \mu \rangle.
\end{aligned}\label{eq:mdkl}
\end{align}
Under slightly different settings, this learning objective is the regularized version of the constrained optimization problem considered in Relative Entropy Policy Search (REPS) \citep{peters2010relative}; And for $\ell_t(\mu)$ depending on $t$, the Equation \refp{eq:mdkl} can also recover the O-REPS method considered in \citet{zimin2013online}. On the other hand, as the KL-divergence (and Bregman divergence) is asymmetric, we can also replace the $D_F(x,y)$ in either formulation (\ref{eq:mm}, \ref{eq:md}) with reverse KL $D_\text{KL}(y,x)$, which will result in different iterative algorithms (as the reverse KL is no longer a Bregman divergence, the equivalence of Formula \refp{eq:mm} and \refp{eq:md} no longer holds). Consider replacing $D_F(\mu,\tilde{\mu}_{t+1})$ with $D_\text{KL}(\tilde{\mu}_{t+1}, \mu)$ in sub-gradient projection (\ref{eq:mm}), we have the ``mirror map'' method with reverse KL as
\begin{align}
\mu_{t+1} = \argmin_{\mu\in\Delta_\Pi} D_\text{KL}(\tilde{\mu}_{t+1}, \mu),\label{eq:mmrkl}
\end{align}
which is essentially the MPO algorithm \citep{abdolmaleki2018maximum} under a probabilistic inference perspective, and MARWIL algorithm \citep{qing2018exponentially} when learning from off-policy data.
Similarly, consider the replacement of $D_F(\mu, \mu_t)$ with $D_\text{KL}(\mu_t,\mu)$ in mirror descent (\ref{eq:md}), we have the ``mirror descent'' method with reverse KL as
\begin{align}
\mu_{t+1} = \argmin_{\mu\in\Delta_\Pi} D_\text{KL}(\mu_t, \mu) + \eta \langle g_t, \mu \rangle,\label{eq:mdrkl}
\end{align}
which can approximately recover the TRPO optimization objective \citep{schulman2015trust} (if the relative entropy between two state-action distributions $D_\text{KL}(\mu_t, \mu)$ in \refp{eq:mdrkl} is replaced by the conditional entropy $D_\text{KL}^{d_t}(\pi_t, \pi)$, also see Section 5.1 of \citet{neu2017unified}).

\red{Besides, we note that there are other choices of constraint for policy optimization as well. For example, in \citep{lee2018maximum,chow2018path,lee2019tsallis}, a Tsallis entropy is used to promote sparsity in the policy distribution. And in \citep{belousov2017fdivergence}, the authors generalize KL, Hellinger distance, and reversed KL to the class of $f$-divergence. In preliminary results, we found divergence based on $0$-potential \citep{audibert2011minimax, bubeck2012regret} is also promising for policy optimization. We left this for future research.}

For multi-step KL divergence regularized policy optimization, we note that the formulation also corresponds to the KL-divergence-augmented return considered previously in several works (\citet{fox2015taming}, Section 3 of \citet{schulman2017equivalence}), although in \citet{schulman2017equivalence} the authors use a fixed behavior policy instead of $\pi_t$ as in ours. More often, the Shannon-entropy-augmented return can be dated back to earlier works \citep{kappen2005path,todorov2007linearly,ziebart2008maximum,nachum2017bridging}, and is a central topic in ``soft'' reinforcement learning \citep{haarnoja2017reinforcement,haarnoja2018soft}.

The mirror descent method is originally introduced by the seminal work of \citet{nemirovsky1983problem} as a convex optimization method. 
Also, the online stochastic mirror descent method has alternative views, e.g.~Follow the Regularized Leader \citep{mcmahan2011follow}, and Proximal Point Algorithm \citep{rockafellar1976monotone}.
For more discussions on mirror descent and online learning, we refer interested readers to the work of \citet{cesa2006prediction} and \citet{bubeck2012regret}.

\section{Experiments}\label{sec:experiments}

\begin{figure}
\centering
  \includegraphics[width=0.9\textwidth]{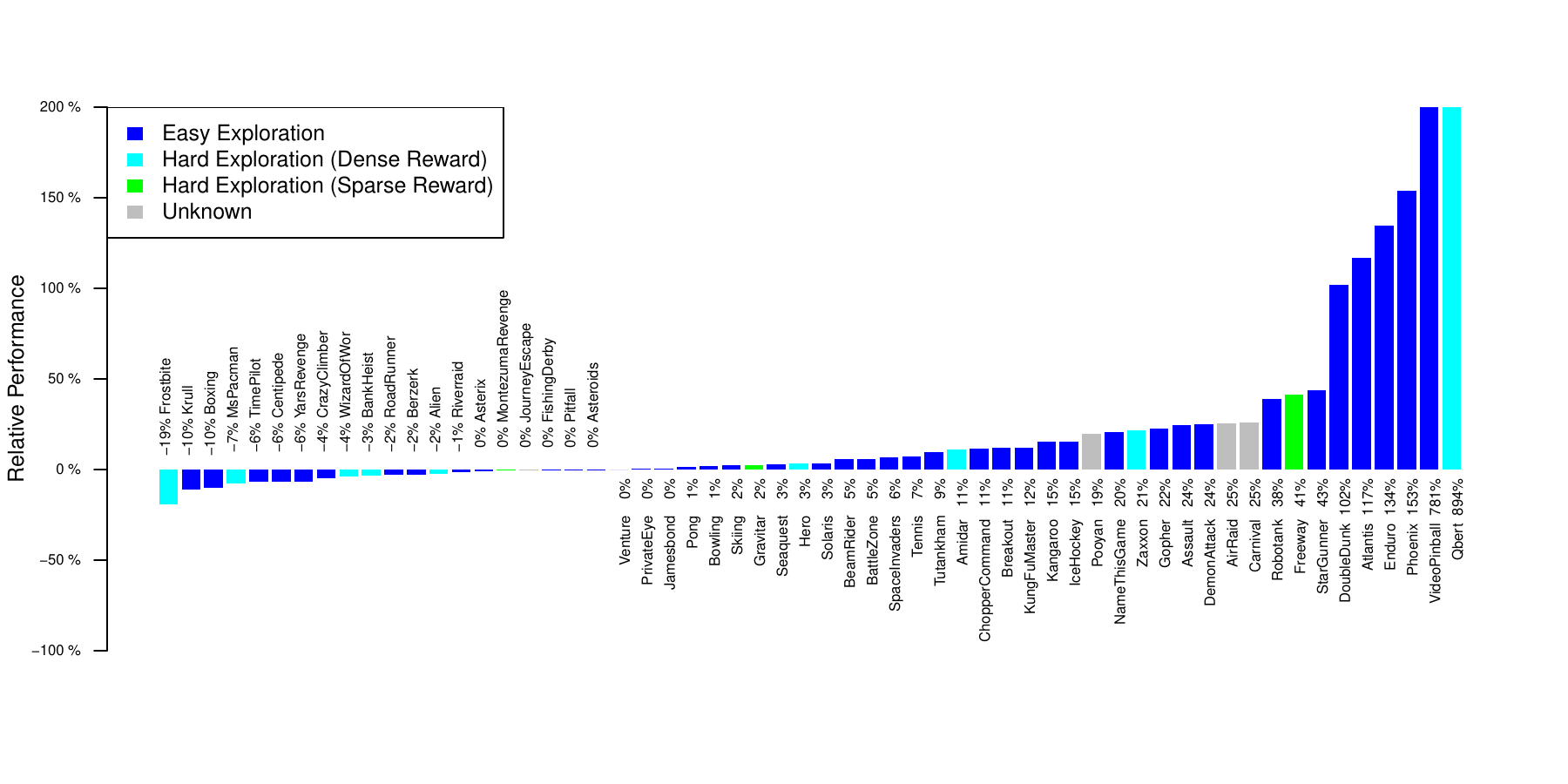}
  \vspace{-2em}
  \caption{Relative score improvement of PPO+DA compared with PPO on 58 Atari environments. The relative performance is calculated as a $\frac{\text{proposed}-\text{baseline}}{\max(\text{human},\text{baseline})-\text{random}}$ \citep{wang2016dueling}. The Atari games are categorized
  according to Figure 4 of \citep{oh2018self}. }
  \label{fig:relative_score}
\end{figure}

\red{In the experiments, we test the exploratory effect of divergence augmentation comparing with entropy augmentation, and the empirical difference between multi-step and 1-step divergence.} For the experiments, we mainly consider the DAPO algorithm \refp{alg:dapo} associated with the conditional KL divergence (see $R_C$ and $D_C$ in \citep{neu2017unified}).
For $F(\mu) = \sum_{s,a} \mu(s,a) \log\frac{\mu(s,a)}{\sum_b \mu(s,b)}$, we have the gradient in \refp{eq:md-pi2} as
\begin{equation*}
    \nabla F(\mu_\pi) - \nabla F(\mu_t) = \log\frac{\pi}{\pi_t}.
\end{equation*}
The multi-step divergence augmentation term as in \refp{eq:nstep-f} is then calculated as
\begin{equation*}
    \hat{D}^\text{KL}_{s,a} = \log\frac{\pi(a_i|s_i)}{\pi_t(a_i| s_i)} + \sum_{j=1}^n\gamma^j(\prod_{k=1}^{j-1}c_{i+k})\rho_{i+j}\log\frac{\pi(a_{i+j}| s_{i+j})}{\pi_t(a_{i+j}| s_{i+j})}.
\end{equation*}
As a baseline, we also implement the PPO algorithm with a V-trace \citep{espeholt2018scalable} estimation of advantage function $A^\pi$ for target policy\footnote{In the original PPO \citep{schulman2017proximal} they use $\hat{A}$ as the advantage estimation of behavior policy $A^{\pi_t}$.}. Specifically, we consider the policy loss as: 
\begin{equation}\label{eq:ppo}
    L^\text{PPO}_\pi(\theta) = \E_{\pi_t,d_t} \min(\frac{\pi_\theta}{\pi_t} A_{s,a}, \text{clip}(\frac{\pi_\theta}{\pi_t}, 1-\epsilon, 1+\epsilon) A_{s,a} ),
\end{equation}
where we choose $\epsilon=0.2$ and the advantage is estimated by $R_{s,a}$. We also tested the DAPO algorithm with PPO, with the advantage estimation $A_{s,a}$ in \refp{eq:ppo} replaced with $\hat{A}_{s,a} - \frac{1}{\eta}\hat{D}_{s,a}$ defined in \refp{eq:nstep-R} and \refp{eq:nstep-f}. We will refer to this algorithm as PPO+DA in the following sections.

\subsection{Algorithm Settings}
The algorithm is implemented with TensorFlow \citep{abadi2016tensorflow}. 
For efficient training with deep neural networks, we use the Adam \citep{kingma2014adam} method for optimization. The learning rate is linearly scaled from 1e-3 to 0. The parameters are updated according to a mixture of policy loss and value loss, with the loss scaling coefficient $c = 0.5$. In calculating multi-step $\lambda$-returns $R_{s,a}$ and divergence $D_{s,a}$, we use fixed $\lambda = 0.9$ and $\gamma = 0.99$. The batch size is set to 1024, with roll-out length set to 32, resulting in 1024/32=32 roll-outs in a batch. The policy $\pi_t$ and value $V_t$ is updated every 100 iterations ($M$ = 100 in Algorithm \ref{alg:dapo}). With our implementation, the training speed is about 25k samples per second, and the data generating speed is about 220 samples per second for each actor, resulting in about 3500 samples per second for a total of 16 actors. \red{Note that the PPO results may not be directly comparable with other works \citep{schulman2017proximal,espeholt2018scalable,xu2018meta}, mainly due to the different number of actors used.} \red{Unless} otherwise noted, each experiment is allowed to run 16000 seconds (about 4.5 hours), corresponding a total of 60M samples generated and 400M samples (with replacement) trained. Details of experimental settings can be found in Appendix \ref{sec:exp_details}.

\subsection{Empirical Results}

\begin{figure}
\vspace{-1em}
\centering
\begin{subfigure}[b]{.32\linewidth}
  \includegraphics[width=\textwidth]{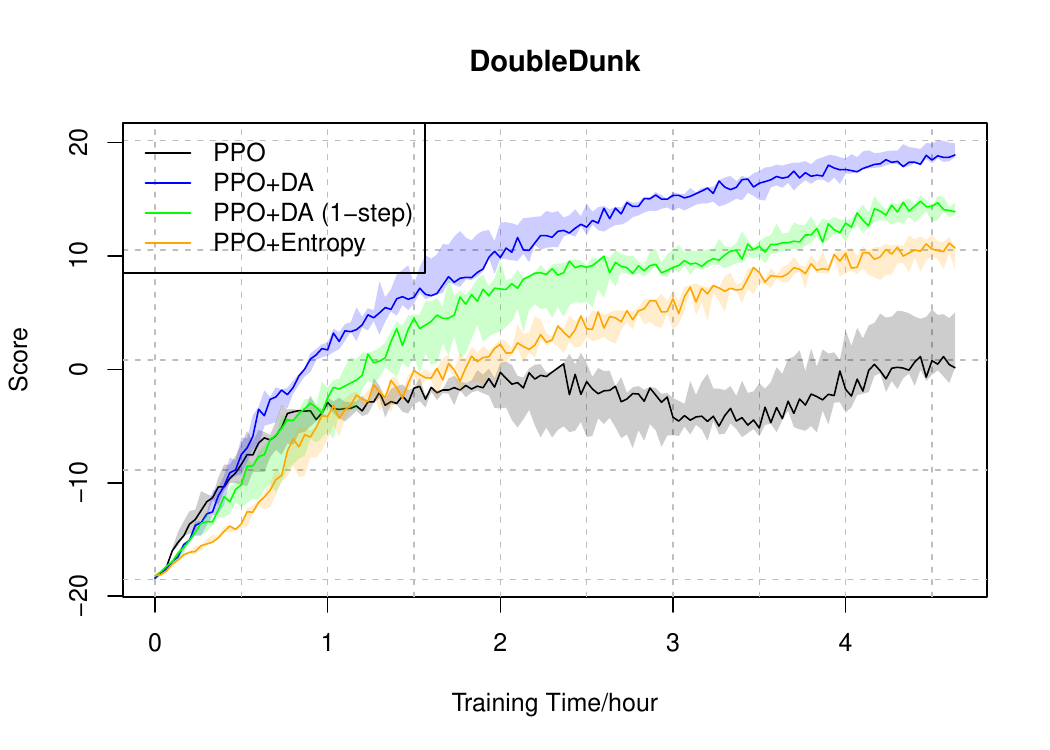}
\end{subfigure}
\begin{subfigure}[b]{.32\linewidth}
  \includegraphics[width=\textwidth]{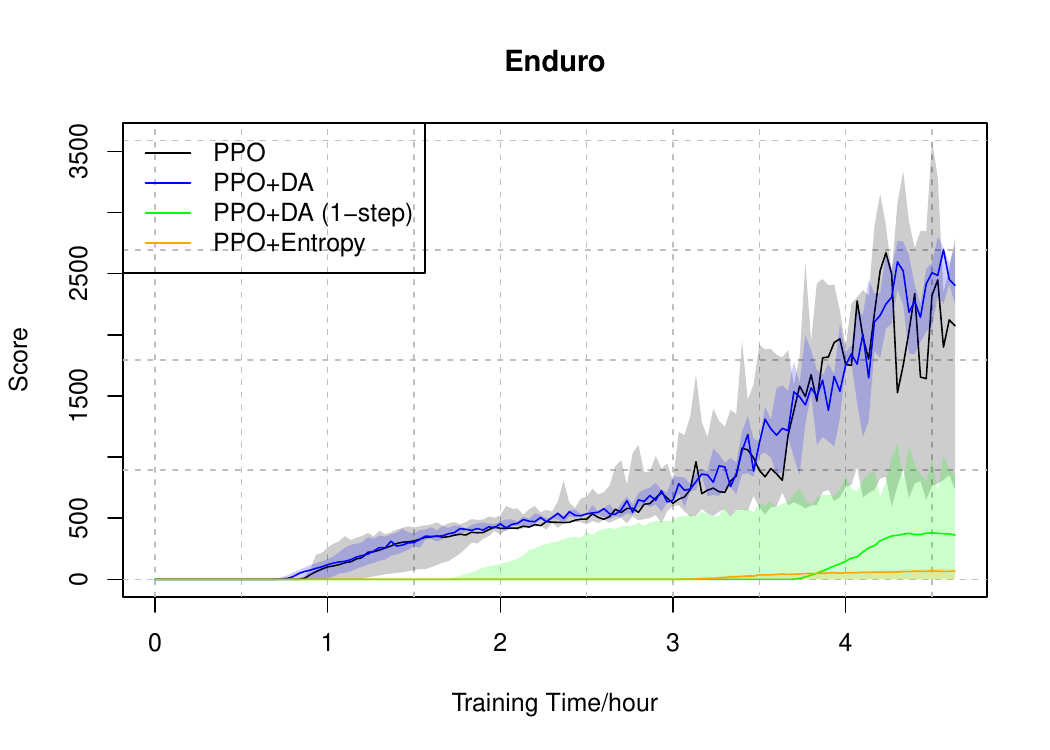}
\end{subfigure}
\begin{subfigure}[b]{.32\linewidth}
  \includegraphics[width=\textwidth]{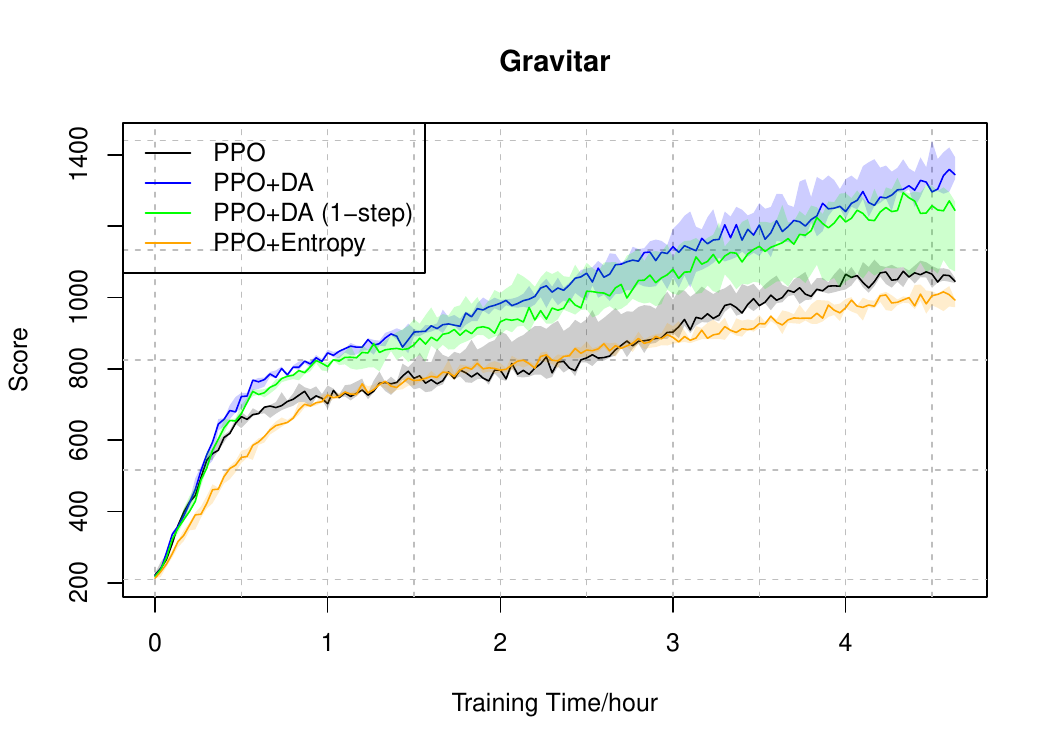}
\end{subfigure}
\begin{subfigure}[b]{.32\linewidth}
  \includegraphics[width=\textwidth]{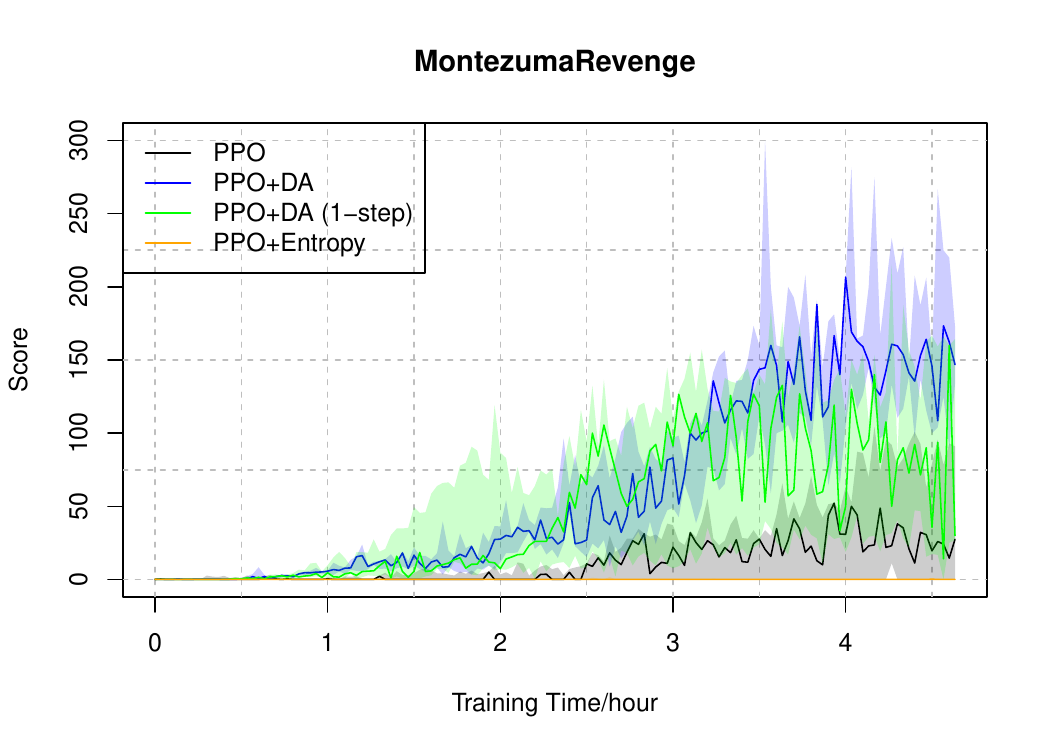}
\end{subfigure}
\begin{subfigure}[b]{.32\linewidth}
  \includegraphics[width=\textwidth]{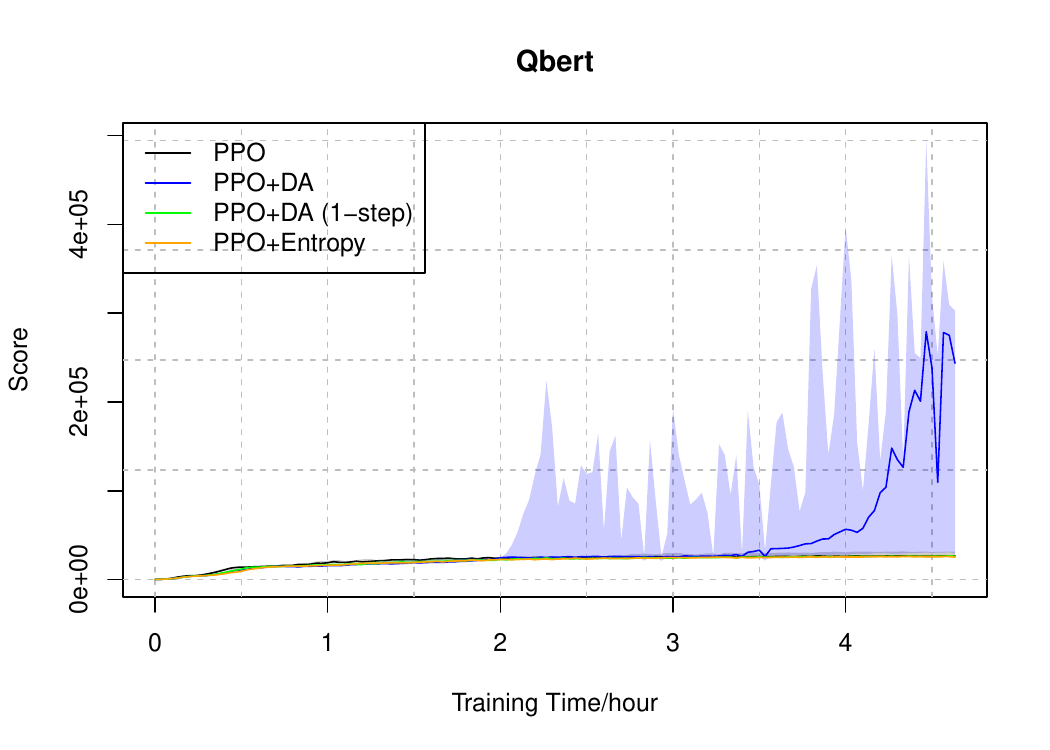}
\end{subfigure}
\begin{subfigure}[b]{.32\linewidth}
  \includegraphics[width=\textwidth]{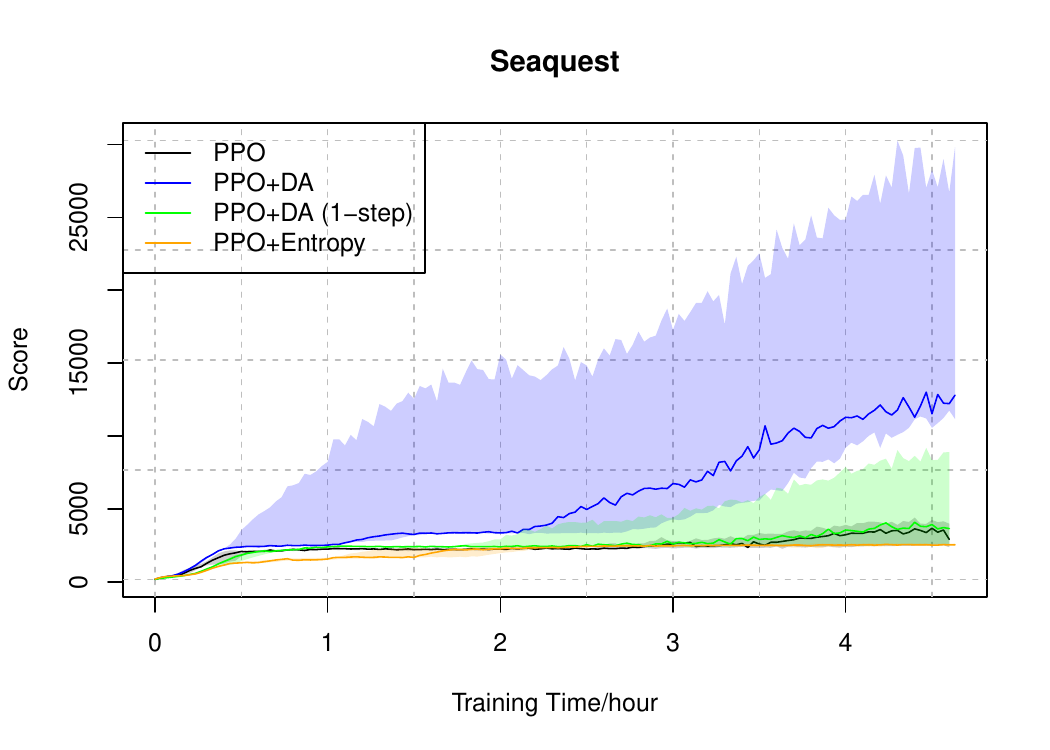}
\end{subfigure}
  \caption{Performance comparison of selected  environments of Atari games. The performance of 
  \textbf{PPO},
  \textcolor[rgb]{0.00,0.00,1.00}{PPO+DA},
  \textcolor[rgb]{0.00,1.00,0.00}{PPO+DA (1-step)}, and
  \textcolor[rgb]{0.96,0.57,0.11}{PPO+Entropy} are plotted in different colors. \red{The score for each game is plotted on the y-axis with running time on the x-axis, as the algorithm is paralleled asynchronously in a distributed environment.}
  For each line in the plots, we run the experiment 5 times with the same parameters and environment settings. The median scores are plotted in solid lines, while the regions between 25\% and 75\% quantiles are shaded with respective colors.}
  \label{fig:compare_ppo}
\end{figure}

We test the algorithm on 58 Atari environments and calculate its relative performance with PPO \citep{schulman2017proximal}. 
The empirical performance is plotted in Figure \ref{fig:relative_score}. We run PPO and PPO+DA with the same environmental settings and computational resources. The relative performance is calculated as $\frac{\text{proposed} - \text{baseline}}{\max(\text{human},\text{baseline}) - \text{random}}$ \citep{wang2016dueling}. We also categorize the game environments into easy exploration games and hard exploration games \citep{oh2018self}. We see that with a KL-divergence-augmented return, the algorithm PPO+DA performs better than the baseline method, especially for the games that may have local minimums and require deeper exploration. We plot the learning curves of PPO+DA (in \textcolor[rgb]{0.00,0.00,1.00}{blue}) comparing with PPO (in \textbf{black}) and other baseline methods on 6 typical environments in Figure \ref{fig:compare_ppo}. Detailed learning curves for PPO and PPO+DA for the complete 58 games can be found in Figure \ref{fig:compare_full} in the Appendix.

\subsubsection{Divergence augmentation vs Entropy augmentation}
We test the effect of divergence augmentation in contrast to the entropy augmentation (plotted in \textcolor[rgb]{0.96,0.57,0.11}{orange} in Figure \ref{fig:compare_ppo}). Entropy augmentation can prevent premature convergence and encourage exploration as well as stabilize policy during optimization. However, the additional entropy may hinder the convergence to the optimal action, as it alters the original learning objective. We set $f(s,a)$ as $\log\pi(a|s)$ in Formula \refp{eq:nstep-f}, and experiment the algorithm with $\frac{1}{\eta} = 0.5, 0.1, 0.01, 0.001$, in which we found that $\frac{1}{\eta} = 0.1$ performs best.
From the empirical results, we see that divergence-augmented PPO works better, while the entropy-augmented version may be too conservative on policy changes, resulting in inferior performance on these games.

\subsubsection{Multi-step divergence vs 1-step divergence}
In Figure \ref{fig:compare_ppo}, we also test the PPO+DA algorithm with its 1-step divergence-augmented counterpart (plotted in \textcolor[rgb]{0.00,1.00,0.00}{green}). We rerun the experiments with the parameter $\bar{c}_D$ (Formula \refp{eq:nstep-f}) set to $0$, which means we only aggregate the divergence on the current state and action $f(s_i, a_i)$, without summing up future discounted divergence $f(s_{i+j}, a_{i+j})$. This method also relates to the conditional divergence defined in Formula \refp{eq:cond_bregman}, and shares more similarities with previous works on regularized and constrained policy optimization methods \citep{schulman2015trust,achiam2017constrained}.
We see that with multi-step divergence augmentation, the algorithm can achieve high scores, especially on games requiring deeper exploration like Enduro and Qbert. We \red{hypothesize} that the accumulated divergence on future states can encourage the policy to explore more efficiently.

\section{Conclusion}\label{sec:conclusion}
In this paper, we proposed a divergence-augmented policy optimization method to improve the stability of policy gradient methods when it is necessary to reuse off-policy data. We showed that the proposed divergence augmentation technique can be viewed as imposing Bregman divergence constraint on the state-action space, which is related to online mirror descent methods. Experiments on Atari games showed that in the data-scarce scenario, the proposed method works better than other state-of-the-art algorithms such as PPO. Our results showed that the technique of divergence augmentation is effective when data generated by previous policies are reused in policy optimization.

\bibliographystyle{abbrvnat}
\bibliography{rl}

\newpage
\appendix

\newcommand{\temp}{\frac{1}{\alpha}}
\newcommand{\KL}{\operatorname{KL}}
\newcommand{\piref}{\overline{\pi}}
\newcommand{\kl}[2]{D_{\mathrm{KL}}\left[#1 ~ \middle \| ~ #2\right]}
\newcommand{\tv}[2]{D_{TV}(#1 \ \| \ #2)}

\newcommand{\given}{\:\vert\:}
\newcommand{\givenb}{\:\middle\vert\:}

\newcommand{\T}{\mathcal{T}}
\newcommand{\Tpi}{\mathcal{T}_{\pi}}
\newcommand{\Tpilam}{\mathcal{T}_{\pi,\lambda}}
\newcommand{\Qpi}{\ensuremath{Q^{\pi}}}
\newcommand{\Api}{\ensuremath{A^{\pi}}}
\newcommand{\Vpi}{\ensuremath{V^{\pi}}}

\newcommand{\Var}{\mathrm{Var}}
\newcommand{\Cov}{\mathrm{Cov}}
\newcommand{\Ea}[1]{\E\left[#1\right]}
\newcommand{\Eb}[2]{\E_{#1}\left[#2\right]}
\newcommand{\Vara}[1]{\Var\left[#1\right]}
\newcommand{\Varb}[2]{\Var_{#1}\left[#2\right]}
\newcommand{\lone}[1]{\norm{#1}_1}
\newcommand{\Prob}[1]{\operatorname{Pr}\left[#1\right]}
\newcommand{\pibol}{\pi^{\mathcal{B}}}
\newcommand{\pibolq}{\pibol_Q}
\newcommand{\pibolth}{\pibol_{\qth}}

\newcommand{\BlackBox}{\rule{1.5ex}{1.5ex}}  

\newcommand{\eqdef}{\stackrel{\text{def}}{=}}
\newcommand{\R}{\mathbb{R}}
\newcommand{\Eg}{\psi}
\newcommand{\Exp}{\mathbf{E}}
\newcommand{\calP}{\mathcal{P}}
\newcommand{\calF}{\mathcal{F}}

\newcommand{\Efp}[1]{(\Ef{#1})}
\newcommand{\Ed}[1]{\textcolor[rgb]{1.00,0.00,0.00}{#1}}

\section{Details of the algorithm} \label{sec:exp_details}

\subsection{Environment Settings}
We evaluate the algorithm on the Atari 2600 video games from Arcade Learning Environment (ALE) \citep{bellemare2013arcade}, which is widely used as a standard benchmark for deep reinforcement learning, especially for distributed training \citep{horgan2018distributed,espeholt2018scalable}. There are at least two evaluation protocols, One is the ``human starts'' protocol \citep{hasselt2016deep}, in which each episode for evaluation is started at a state sampled from human play. The other is the ``no-ops start'' protocol \citep{mnih2015human}, in which the starting state of an episode for evaluation is generated by playing a random length of no-op actions in the environment. We adopt the later evaluation protocol both for training and testing in this work. The environment is based on the OpenAI Gym \citep{greg2016openai}. We mostly follow the environment settings as in \citep{mnih2015human}. The environment is randomly initialized by playing a random number (no more than 30) of no-op actions. During playing, each action is repeated for 4 contiguous frames, and every 4th frame is taken a pixel-wise max over the previous frame, and then returned as the screen observation. The size of the raw screen is 210$\times$160 pixels with 128 colors. The colored image is firstly converted to gray-scale and then resized to 84$\times$84 pixels represented by integers from 0 to 255, followed by a scaling to floats between 0 to 1. We also use the ``episodic life'' trick in the training phase: For games with a life counter, the loss of life is marked as an end for the current episode. The rewards are clipped with a $sgn()$ function, such that positive rewards are represented by 1, negative rewards as -1, and 0 otherwise. For some games (e.g.~Atlantis) we observe that there is a maximum limit of 100000 steps for each episode, corresponding to 400000 raw frames. In addition to the settings above, we also reset the environment if no reward is received in 1000 steps, to prevent the environment from accidental stuck.

\subsection{Network Structure}

For comparable results, we use a similar network structure as in \citep{mnih2015human}. The first layer consists of 32 convolution filters of 8$\times$8 with stride 4 and applies a ReLU non-linearity. And the second layer convolves the image with 64 filters of 4$\times$4 with stride 2 followed by a ReLU rectifier. The third layer has 64 filters of 3$\times$3 with stride 1 followed by a rectifier. The final hidden layer is fully-connected and consists of 512 ReLU units. The output layer is two-headed, representing $\pi_\theta(\cdot|s)$ and $V_\theta(s)$ respectively. For the vanilla model, the input is stacked with 4 frames; while for RNN model, we put an additional LSTM \citep{hochreiter1997long} layer with 256 cells after the fully-connected layer, which is similar to the previous works \citep{espeholt2018scalable}.

\subsection{Training Framework}

For large-scale training with the interested algorithms, we adopt an ``Actor-Learner'' style \citep{horgan2018distributed,espeholt2018scalable} distributed framework. In our distributed framework, actors are responsible for generating massive trajectories with current policy; while learners are responsible for updating policy with the data generated by the actors. To be specific, in its main loop, an actor runs a local environment with actions from current local policy and caches the generated data at local memory. The running policy is updated periodically to the latest policy at the learner; while the generated data is sent to the learner asynchronously. At the learner side, the learner keeps at most 20 latest episodes generated by each actor respectively, in a FIFO manner. Each batch of samples for training are randomly sampled from these trajectories with replacement. We deploy the distributed training framework on a small cluster. The learner runs on a GPU machine and occupies an M40 card, while actors run in 16 parallel processes on 2.5GHz Xeon Gold 6133 CPUs. 

\subsection{Multi-step Return}
In this work, we utilize the V-trace (Section 4.1 \citep{espeholt2018scalable}) estimation $v_{s_i} = v_i$ along a trajectory starting at $(s_i,a_i=s,a)$ under $\pi_t$ defined recursively as
\begin{align}
    v_j = \left\{
    \begin{aligned}
    &V_\theta(s_j) + \delta_j V_\theta + \gamma c_j(v_{j+1} - V_\theta(s_{j+1})), & i\le j<n\\
    &r_j + \gamma \lambda_V v_{j+1} + \gamma (1-\lambda_V) V_t(s_{j+1}), & n\le j<T
    \end{aligned}\right.
    \label{eq:nstep-V} 
\end{align}

where $\delta_j V_\theta = \rho_j(r_j + \gamma V_\theta(s_{j+1}) - V_\theta(s_j))$, $\rho_j = \min(\bar{\rho}_V,\frac{\pi_\theta(a_j|s_j)}{\pi_t(a_j|s_j)})$, $c_j = \min(\bar{c}_V,\frac{\pi_\theta(a_j|s_j)}{\pi_t(a_j|s_j)})$, and $0\le\lambda_V\le 1$. The state value estimation function at iteration $t$ is denoted as $V_t(\cdot) = V_{\theta_t}(\cdot)$. The definition of \refp{eq:nstep-V} can be seen as following V-trace algorithm along the roll-out (for which we have $\pi_\theta(a_j| s_j)$ and $V_\theta(s_j)$) for $i\le j< n$, and switch to TD($\lambda$) until a terminal time $T$ (which is estimated offline as we only have $V_t(s_j)$ instead of $\pi_\theta(a_j|s_j)$ and $V_\theta(s_j)$ for $n\le j < T$). It is noted that TD($\lambda$) also corresponds to V-trace in on-policy settings (Remark 2, \citep{espeholt2018scalable}). 

\subsection{Hyper-parameters}
The default hyper-parameters used in our experiments are given in Table \ref{table:hyperparameters}.
\begin{table}
  \caption{Hyper-parameters}
  \label{table:hyperparameters}
  \centering
  \begin{tabular}{ll}
    \toprule
    Name     & Value \\
    \midrule
    Batch size          & 1024 \\
    Replay memory size  & 16384 ($2^{14}$) \\
    $\lambda$           & 0.9 \\
    Rollout length      & 32 \\
    Burn-in samples     & 1024 \\
    Learning rate       & 0.001 to 0 \\
    $\bar{c}_D$ (Formula \ref{eq:nstep-f}) & 0.5 \\
    $\bar{\rho}_D$ (Formula \ref{eq:nstep-f}) & 1.0 \\
    $\bar{c}_V$ (Formula \ref{eq:nstep-V}) & 1.0 \\
    $\bar{\rho}_V$ (Formula \ref{eq:nstep-V}) & 1.0 \\
    $\epsilon$ (Formula \ref{eq:ppo}) & 0.2 \\
    $1/\eta$            & 0.5 \\
    $b$ (Formula \ref{eq:update}) & 0.5 \\
    Optimizer           & Adam \\
    \bottomrule
  \end{tabular}
\end{table}

\setcounter{equation}{0}
\section{Theoretical Analysis} \label{sec:analysis}
In this section, we provide some theoretical analysis of the parametrized algorithm considered in this work. We show that the bias introduced by omitting the state ratio can be bounded by the divergence up to a constant factor.
\subsection{Error bound of the biased gradient}
Without ambiguity, we define $\pi=\pi_\theta$, $\pi_t=\pi_{\theta_t}$ for brevity, the policy value as
\[
V(\theta) = \langle r, \mu_{\pi_\theta} \rangle = \E_{s,a\sim d_\pi \pi} r(s,a),
\] and the conditional KL-divergence for parametrized policies as
\[
D(\theta, \theta_t) = \E_{s,a\sim d_\pi \pi} \log\frac{\pi(a|s)}{\pi_{t}(a|s)}.
\]
The algorithm iteratively maximizes the following equation with SGD steps
\begin{align}
\theta_{t+1} \approx \argmax_{\theta} f(\theta,\theta_t) \equiv V(\theta) - \lambda D(\theta,\theta_t). \label{eq:prox}
\end{align}
Denote $A^{\pi}(s,a)$ as the advantage function of reward $r$ following target policy $\pi$, and $\mathbf{A}^{\pi}_{\pi_t}(s,a)$ as the advantage function of pseudo reward $(\log\pi - \log\pi_t)$ following target policy $\pi$,
the gradient of \eqref{eq:prox} is
\[
\nabla_\theta f(\theta, \theta_t) = \E_{(s,a)\sim d_{\pi_t} \pi_t} \frac{d_\pi(s)}{d_{\pi_t}(s)} \frac{\pi(a|s)}{\pi_t(a|s)} (A^{\pi}(s,a) - \lambda \mathbf{A}^{\pi}_{\pi_t}(s,a)) \nabla_\theta\log\pi(a|s).
\]
In the actual implementation, we omit the state ratio of $d_\pi / d_{\pi_t}$, resulting in a biased gradient
\[
g(\theta, \theta_t) = \E_{(s,a)\sim  d_{\pi_t} \pi_t} \frac{\pi(a|s)}{\pi_t(a|s)} (A^{\pi}(s,a) - \lambda \mathbf{A}^{\pi}_{\pi_t}(s,a)) \nabla_\theta\log\pi(a|s).
\]
In the following proposition, for target policy $\pi \equiv \pi_{\theta}$ and reference policy $\tilde{\pi} \equiv \pi_{\tilde{\theta}}$, we show that the error $\delta(\theta, \tilde{\theta}) \equiv \norm{\nabla f(\theta, \tilde{\theta})- g(\theta,\tilde{\theta})}$ introduced by omitting the state ratio can be bounded by the conditional KL divergence. To be rigorous, we make the following assumptions:
\begin{assumption}[Universal boundedness] For all $\theta, \tilde{\theta} \in \Theta$, the set for policy parametrization, there exist non-negative constants $\zeta_1,\zeta_2$ such that,
\[
\max_{s} \E_{a\sim \pi}\norm{\nabla_{\theta} \ln \pi(a|s)} \leq \zeta_1,
\]
\[
\max \left\{\max_{s,a} |A^{\pi}(s,a)|, \max_{s,a} |A^{\pi}(s,a) -\mathbf{A}^{\pi}_{\tilde{\pi}}(s,a)| \right\} \leq \zeta_2.
\]
\end{assumption}
Then we have the following proposition
\begin{prop}
Under Assumption 1, the norm of the gradient bias can be bounded by the conditional KL-divergence:
\[
\delta(\theta,\tilde{\theta})^2 \le c D(\theta, \tilde{\theta}).
\]
\end{prop}

\begin{proof}
The proof provided here is based on the perturbation theory. We firstly define the symbols and notations we used in the proof. Consider the difference on each state denoted as
\begin{align}
\Delta_{\theta,\tilde{\theta}}(s) = \E_{a \sim \pi}
 \norm{\left[ A^{\pi}(s,a) - \lambda \mathbf{A}^{\pi}_{\tilde{\pi}}(s,a)\right] \nabla_\theta \ln \pi(a|s)}. \label{eq:delta_diff}
\end{align}
By triangular inequality and H\"{o}lder's inequality, we have
\[
    \delta(\theta,\tilde{\theta}) =
    \norm{\nabla_\theta f(\theta, \tilde{\theta}) - g(\theta,\tilde{\theta})} \leq
    \langle |d_\pi - d_{\tilde{\pi}}|, \Delta_{\theta,\tilde{\theta}}\rangle \leq
    \norm{d_\pi - d_{\tilde{\pi}}}_1
    \norm{\Delta_{\theta,\tilde{\theta}}}_{\infty}.
\]
Let $P_\pi \in \R^{|S|\times |S|}$ be the transition matrix associated with policy $\pi$
$$P_\pi (s'|s) = \sum _a P(s'|s,a) \pi(a|s).$$ The discounted state distribution can be written as
\[
    d_\pi = (1-\gamma) \sum_{t=0}^{\infty} (\gamma P_\pi)^t d_0= (1-\gamma) (I - \gamma P_{\pi})^{-1}d_0,
\]
where $d_0$ is the initial state distribution. For policies $\pi$ and $\tilde{\pi}$, consider the matrices $G \equiv (I - \gamma P_{\pi})^{-1}$ and $\tilde{G} \equiv (I -\gamma P_{\tilde{\pi}})^{-1}$, we have
\[
    G^{-1} - \tilde{G}^{-1} = (I - \gamma P_\pi) - (I - \gamma P_{\tilde{\pi}}) = \gamma (P_{\tilde{\pi}} - P_{\pi}).
\]
Multiplying by $\tilde{G}$ and $G$ on the left and right side respectively, we have
\[
    \tilde{G} - G = \gamma \tilde{G}(P_{\tilde{\pi}} - P_{\pi})G \label{eq:perturb-A}.
\]
The difference of state distribution can be bounded as
\begin{align*}
\norm{d_\pi - d_{\tilde{\pi}}}_1 & = \norm{(1-\gamma) (G - \tilde{G}) d_0}_1\\
    & = \norm{\gamma (1-\gamma) \tilde{G}(P_{\pi} - P_{\tilde{\pi}}) {G}d_0}_1 \\
    & = \norm{\gamma \tilde{G} (P_\pi - P_{\tilde{\pi}})d_\pi}_1\\
    & \leq \gamma \norm{\tilde{G}}_1 \norm{(P_\pi - P_{\tilde{\pi}})d_\pi}_1\\
    & = \gamma \norm{\sum_{t=0}^{\infty}(\gamma P_{\tilde{\pi}})^t}_1 \norm{(P_\pi - P_{\tilde{\pi}})d_\pi}_1\\
    & \leq \gamma \sum_{t=0}^{\infty}\gamma^t \norm{P_{\tilde{\pi}}^t}_1 \norm{(P_\pi - P_{\tilde{\pi}})d_\pi}_1.
\end{align*}
Since the transition matrix, $P_{\tilde{\pi}}$ is a left stochastic matrix \citep{asmussen2003markov},
\begin{align*}
    \norm{d_\pi - d_{\tilde{\pi}}}_1 &\leq \gamma \sum_{t=0}^{\infty}\gamma^t \norm{(P_\pi - P_{\tilde{\pi}})d_\pi}_1\\
    & = \gamma (1-\gamma)^{-1} \norm{(P_\pi - P_{\tilde{\pi}})d_\pi}_1,
\end{align*}
we have that
\begin{align*}
    \norm{(P_\pi - P_{\tilde{\pi}})d_\pi}_1 &=
    \sum_{s} \left|\sum_{s',a} \left(P(s'|s,a)
    \left(\pi(a|s) -\tilde{\pi}(a|s)\right) \right) d_\pi(s)\right|\\
    &\leq \sum_{s,s'} \left|\sum_a \left(P(s'|s,a)
    \left(\pi(a|s) -\tilde{\pi}(a|s)\right) \right) \right|d_\pi(s) \\
    &\leq \sum_{s,s',a} P(s'|s,a) \left|
    \left(\pi(a|s) -\tilde{\pi}(a|s) \right) \right|d_\pi(s)\\
    & = \sum_{s,a} \left|
    \left(\pi(a|s) -\tilde{\pi}(a|s) \right) \right|d_\pi(s) \sum_{s'} P(s'|s,a)\\
    & = \E_{s\sim d_\pi} \norm{\pi(\cdot|s) -\tilde{\pi}{(\cdot|s)}}_1\\
    & = 2 \E_{s\sim d_\pi} D_{\operatorname{TV}}\left(\pi(\cdot|s) ,\tilde{\pi}(\cdot|s)\right)\\
    & \leq \E_{s\sim d_\pi} \sqrt{2D_{\operatorname{KL}}\left(\pi(\cdot|s) ,\tilde{\pi}(\cdot|s)\right)}\\
    & \leq \sqrt{2\E_{s\sim d_\pi}D_{\operatorname{KL}}\left(\pi(\cdot|s) ,\tilde{\pi}(\cdot|s)\right)}.
\end{align*}
of which the first two inequalities follow from the triangular inequality, the relationship between $D_{\operatorname{TV}}$ and $D_{\operatorname{KL}}$ is deduced by Pinsker's inequality \citep{csiszar2011information}, and the last inequality is by Jensen's inequality with concavity. 

From the definition of $D(\theta,\tilde{\theta})$ and \refp{eq:delta_diff}, we could get
\begin{align*}
    \delta(\theta,\tilde{\theta})& \leq
    \norm{d_\pi - d_{\tilde{\pi}}}_1
    \norm{\Delta_{\theta,\tilde{\theta}}}_{\infty}\\
    & \leq \frac{\gamma}{1-\gamma} \norm{(P_\pi - P_{\tilde{\pi}})d_\pi}_1 \norm{\Delta_{\theta,\tilde{\theta}}}_{\infty}\\
    & \leq \frac{\gamma}{1-\gamma}\max_s\Delta_{\theta,\tilde{\theta}}(s) \sqrt{2D(\theta, \tilde{\theta})}.
\end{align*}
The final result follows from squaring both sides
\begin{align*}
    \delta(\theta,\tilde{\theta})^2
    & \leq 2\left(\frac{\gamma}{1-\gamma}\max_s\Delta_{\theta,\tilde{\theta}}(s) \right)^2 D(\theta, \tilde{\theta}) \\
    & \leq 2\left(\frac{\gamma}{1-\gamma}\zeta_1\zeta_2 \right)^2 D(\theta, \tilde{\theta}) \\
    & = c D(\theta,\tilde{\theta}).
\end{align*}

\end{proof}

\section{Additional Empirical Results}
For the algorithm performance as summarized in Figure \ref{fig:relative_score}, we provide the comparison results for each game in details. \red{We also provide experimental results with 64 actors for interested readers.}
\begin{figure*}
\includegraphics[width=0.245 \textwidth]{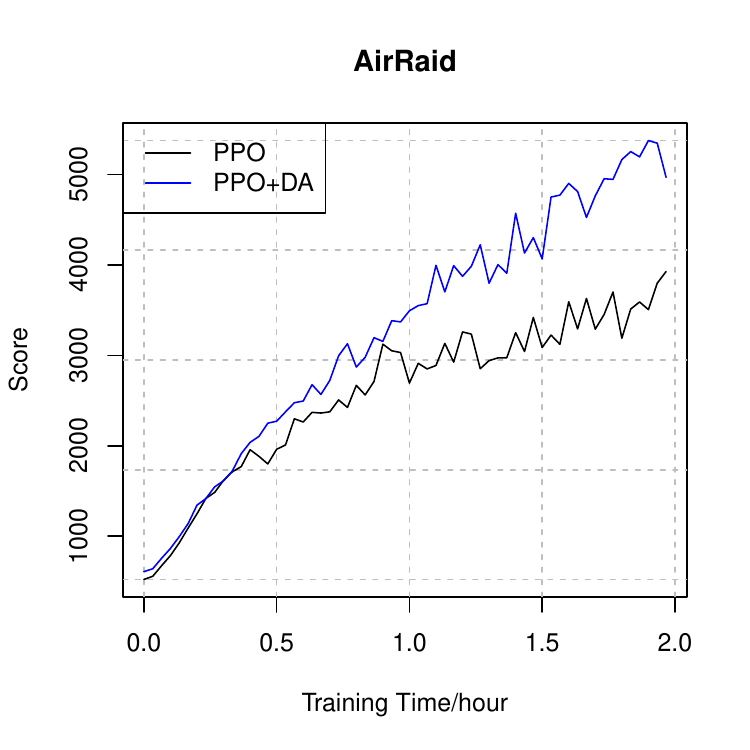}
\includegraphics[width=0.245 \textwidth]{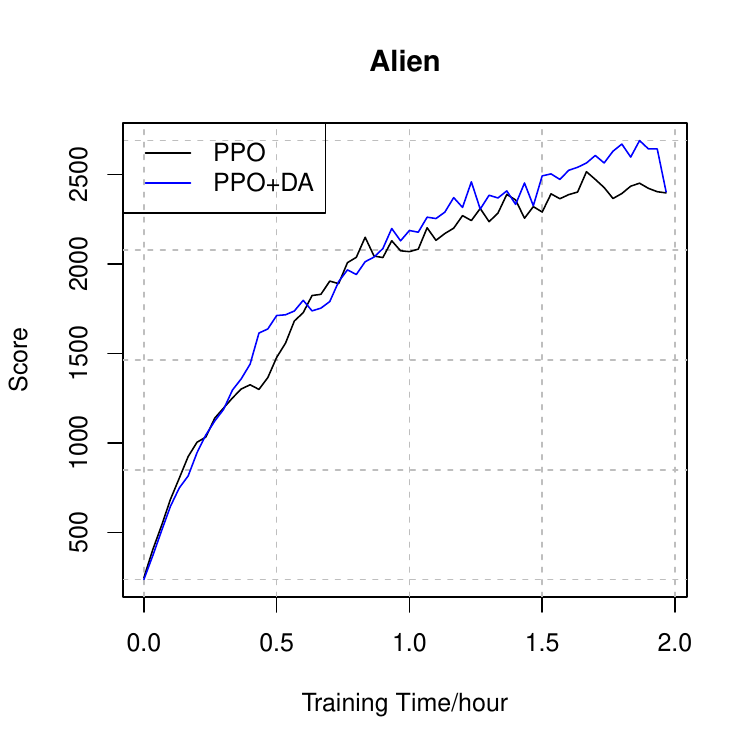}
\includegraphics[width=0.245 \textwidth]{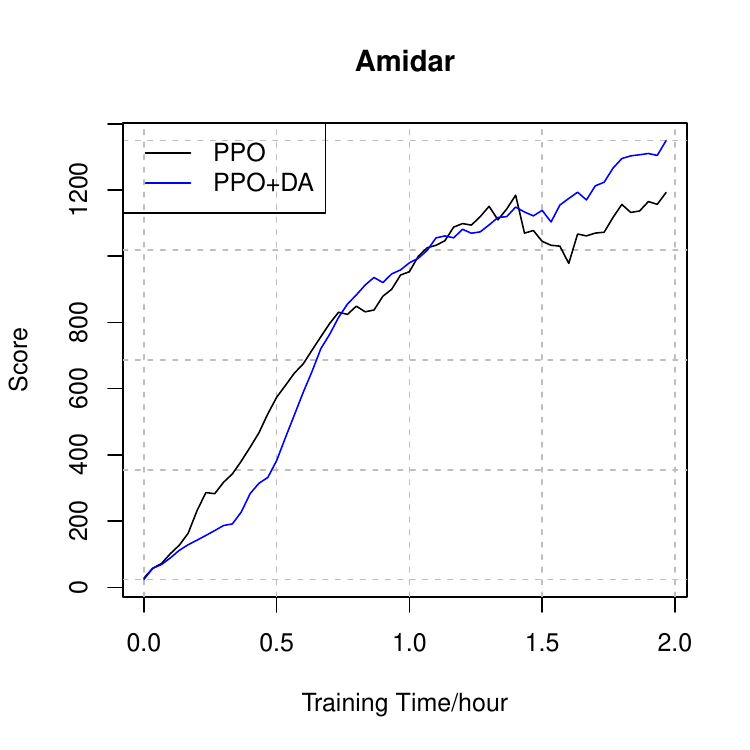}
\includegraphics[width=0.245 \textwidth]{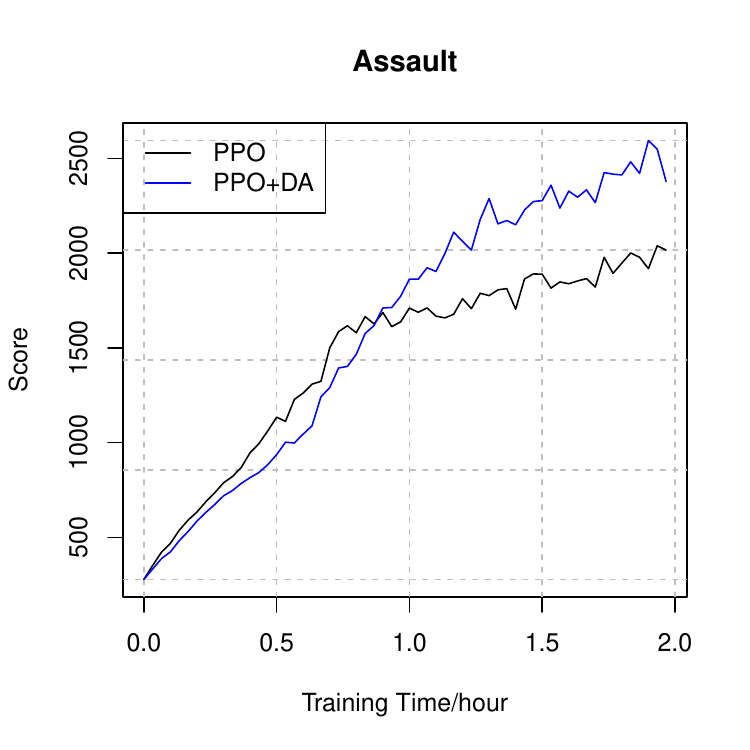}
\includegraphics[width=0.245 \textwidth]{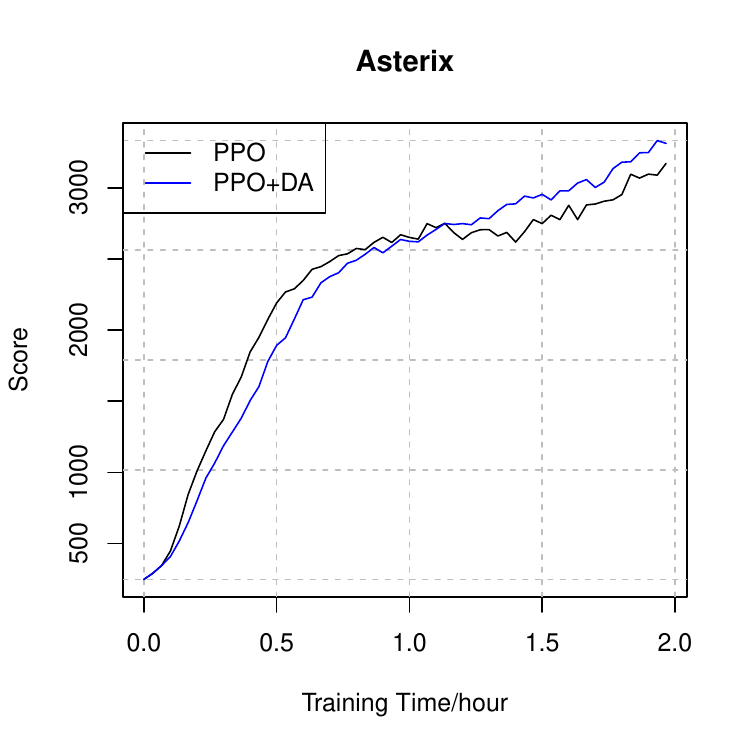}
\includegraphics[width=0.245 \textwidth]{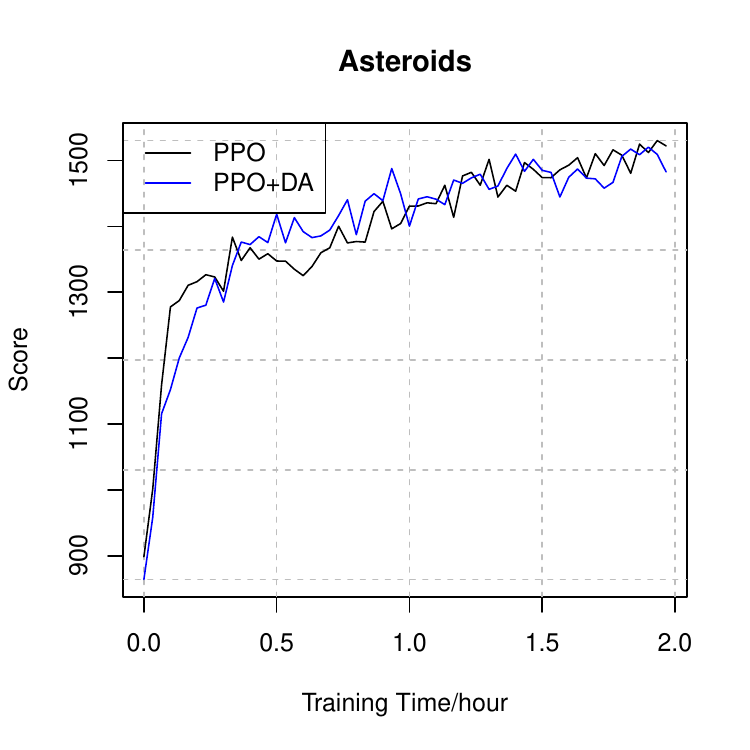}
\includegraphics[width=0.245 \textwidth]{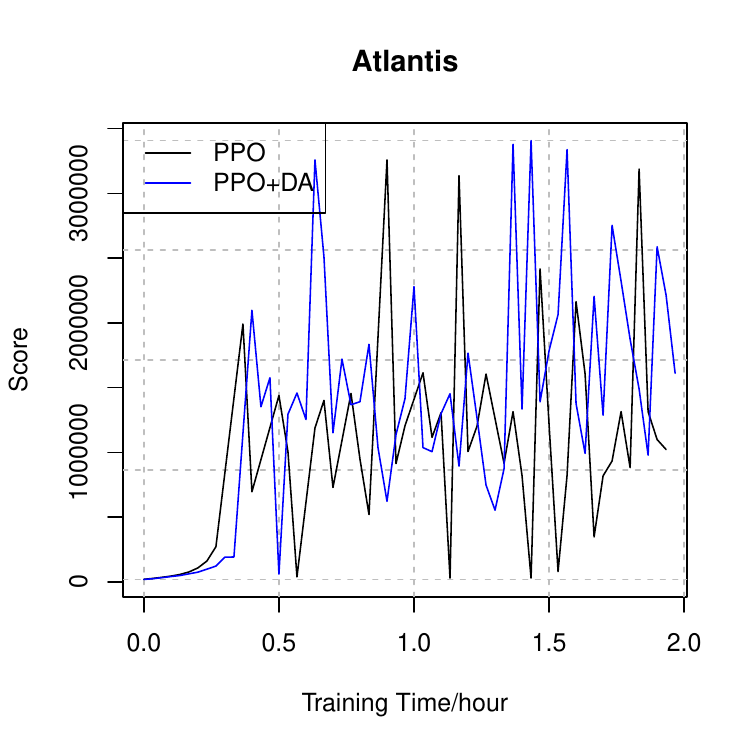}
\includegraphics[width=0.245 \textwidth]{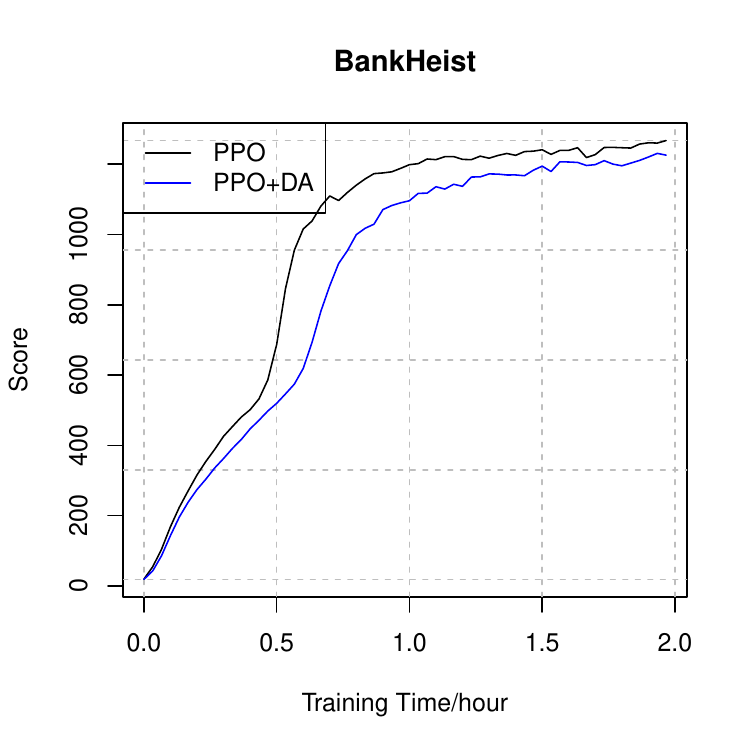}
\includegraphics[width=0.245 \textwidth]{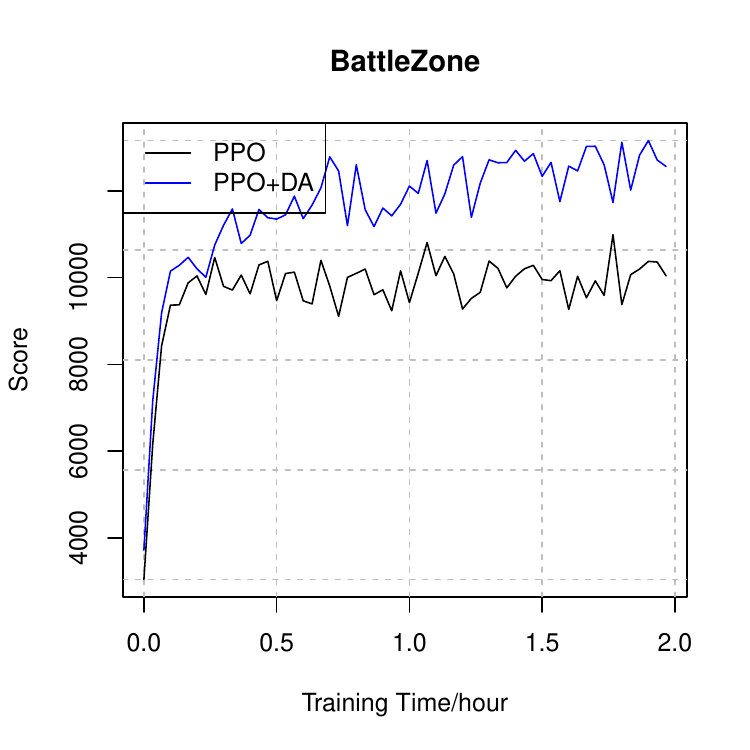}
\includegraphics[width=0.245 \textwidth]{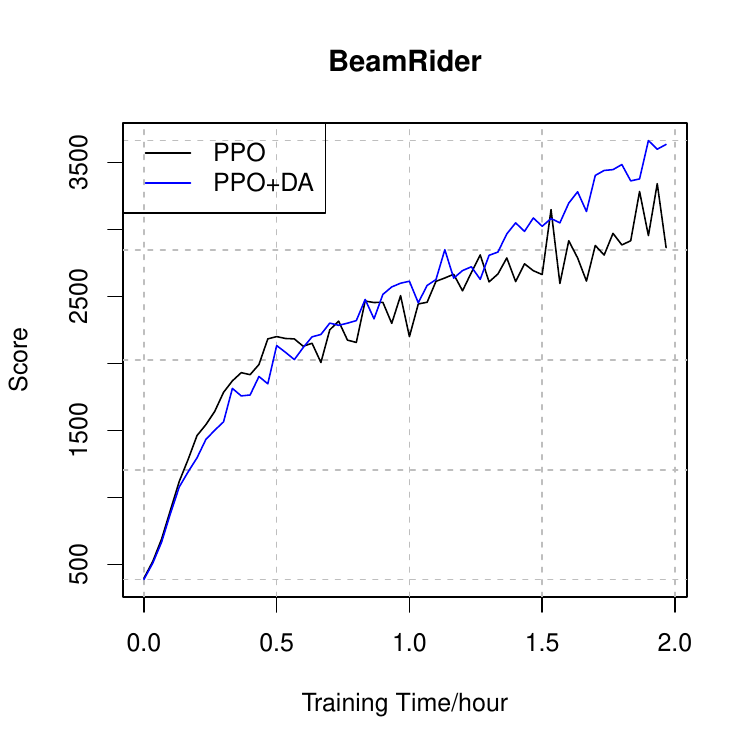}
\includegraphics[width=0.245 \textwidth]{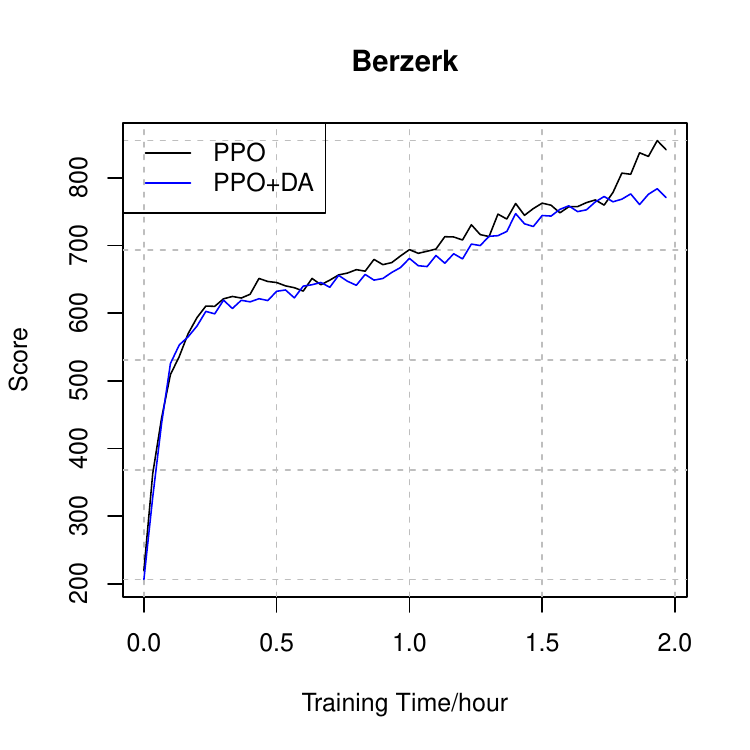}
\includegraphics[width=0.245 \textwidth]{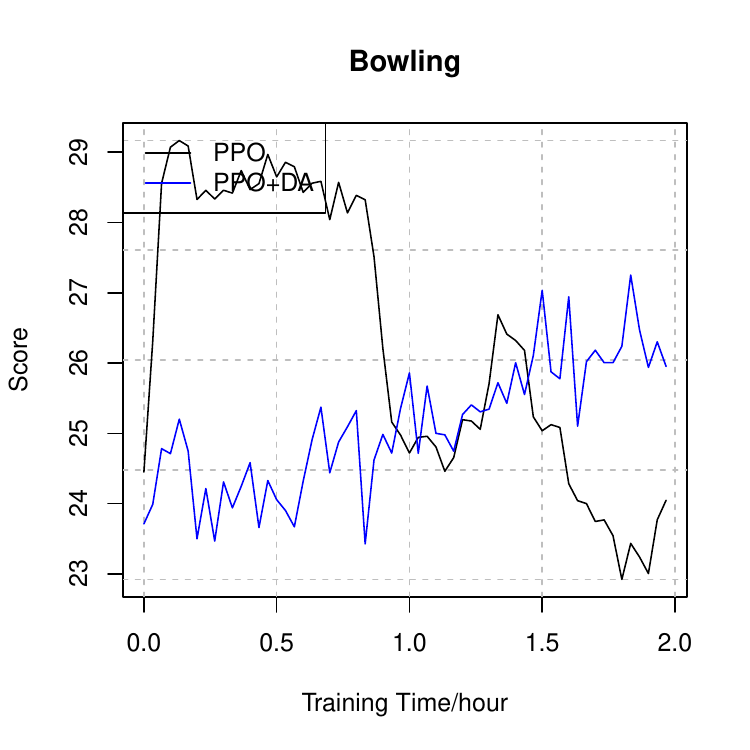}
\includegraphics[width=0.245 \textwidth]{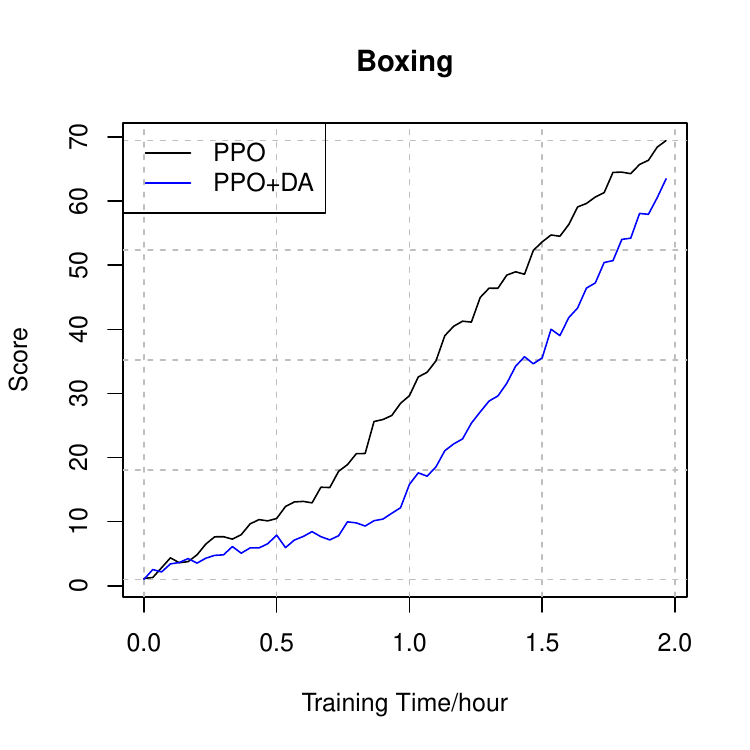}
\includegraphics[width=0.245 \textwidth]{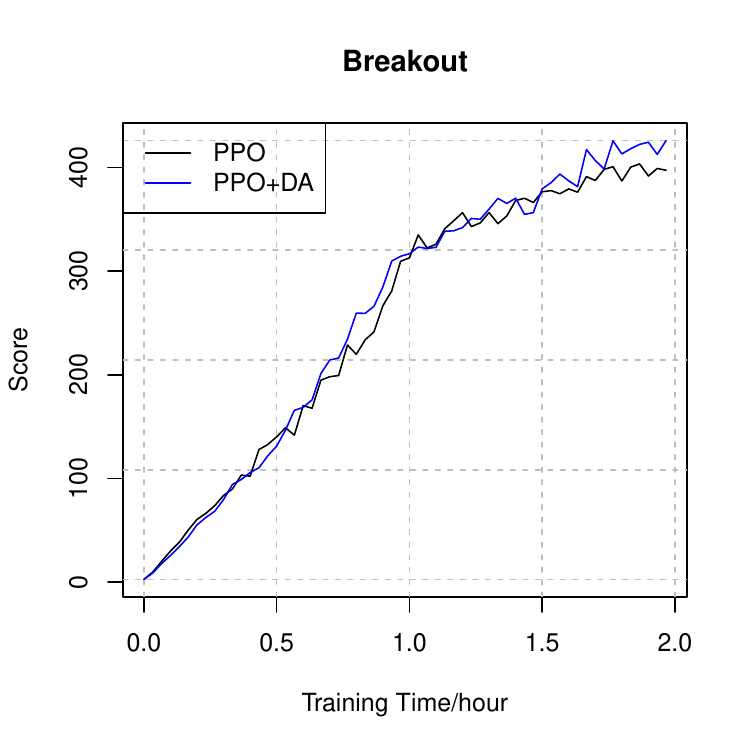}
\includegraphics[width=0.245 \textwidth]{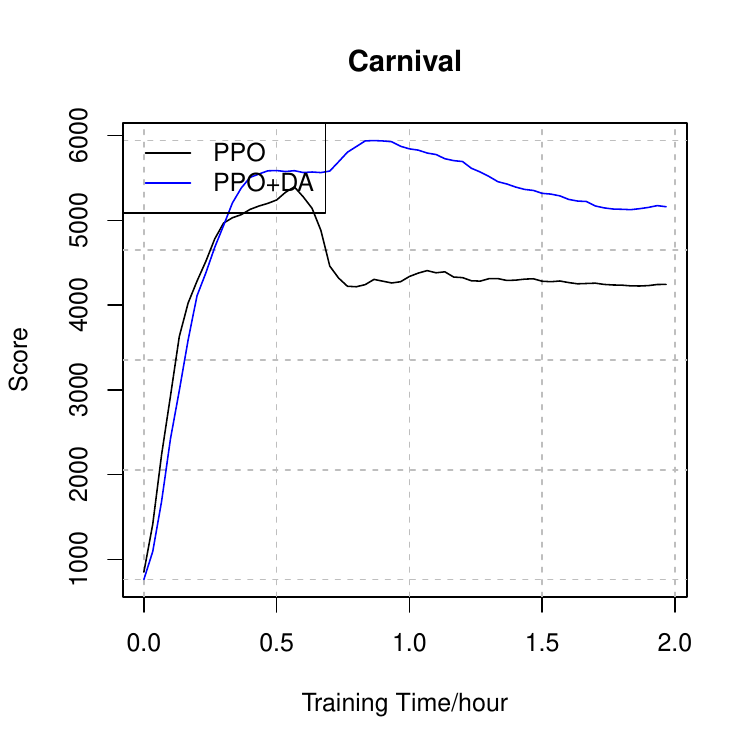}
\includegraphics[width=0.245 \textwidth]{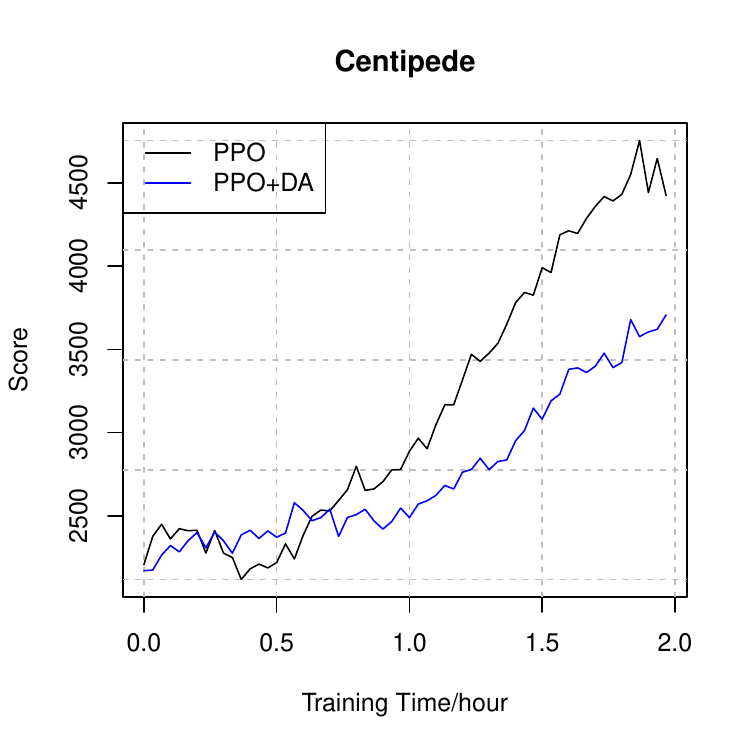}
\includegraphics[width=0.245 \textwidth]{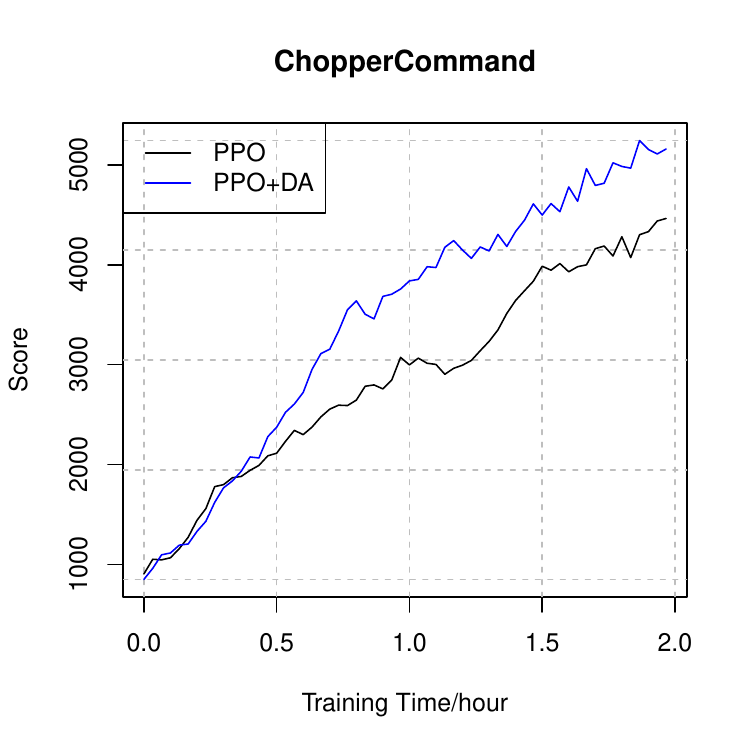}
\includegraphics[width=0.245 \textwidth]{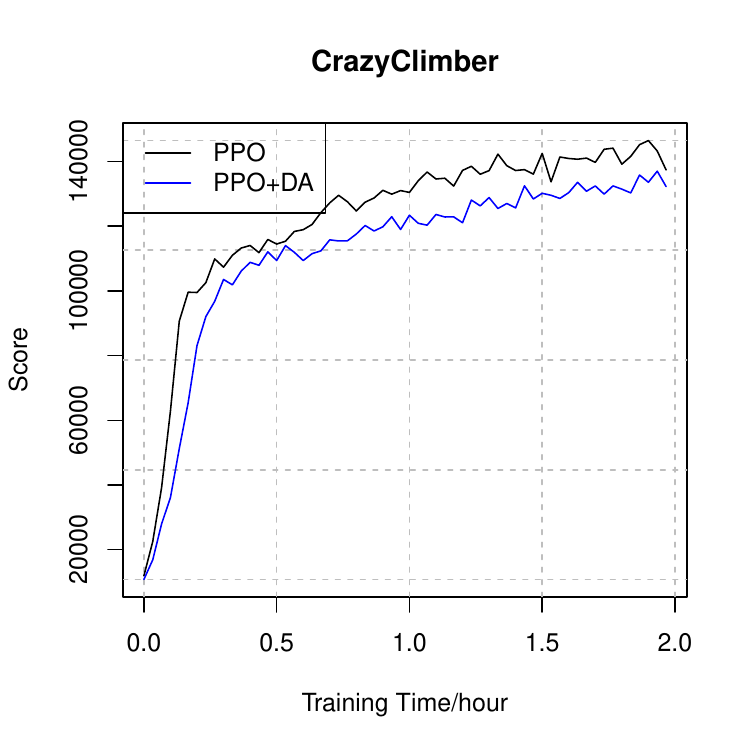}
\includegraphics[width=0.245 \textwidth]{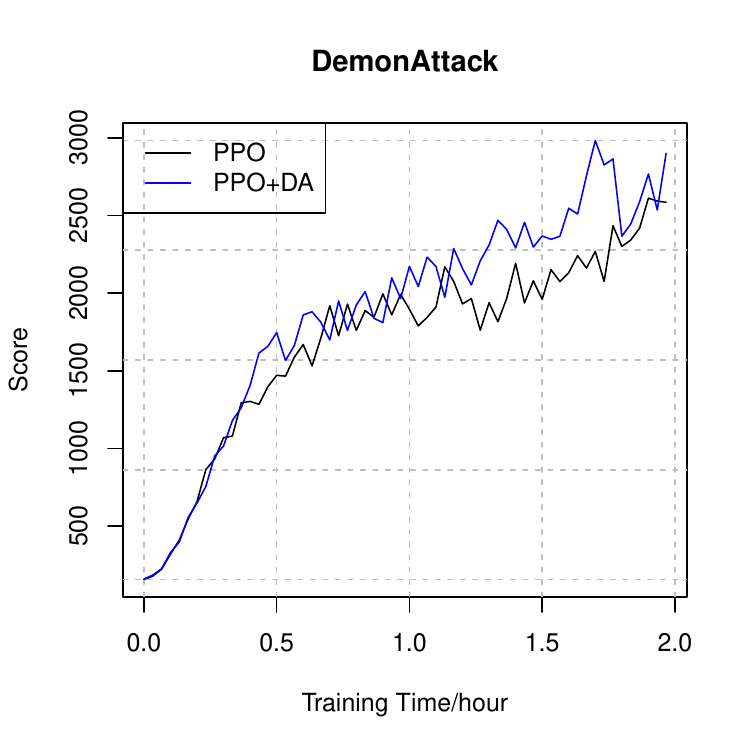}
\includegraphics[width=0.245 \textwidth]{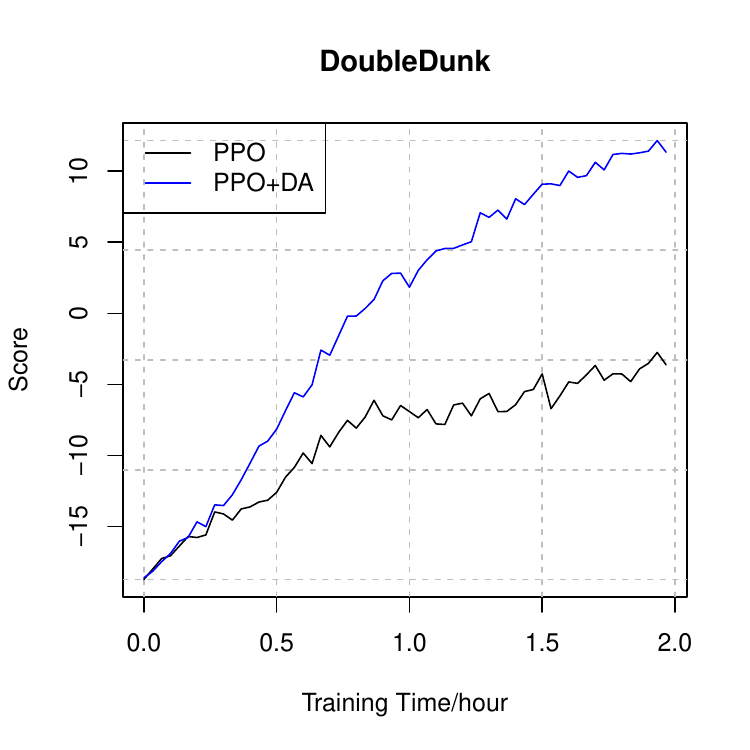}
\includegraphics[width=0.245 \textwidth]{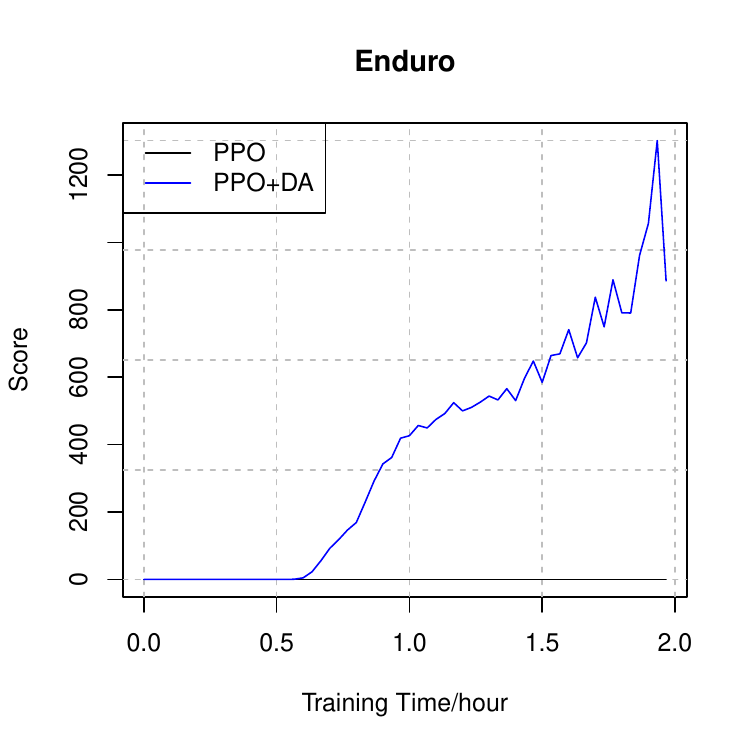}
\includegraphics[width=0.245 \textwidth]{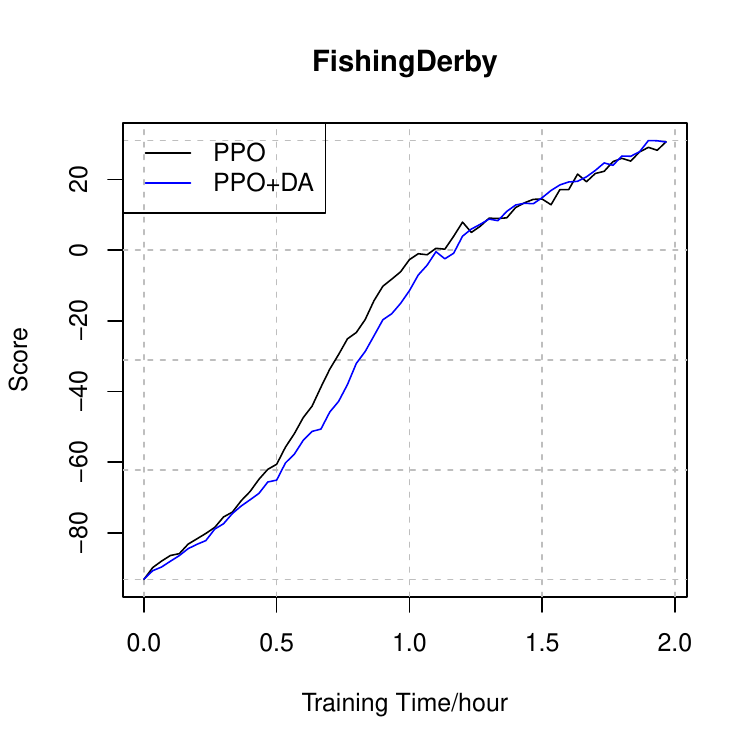}
\includegraphics[width=0.245 \textwidth]{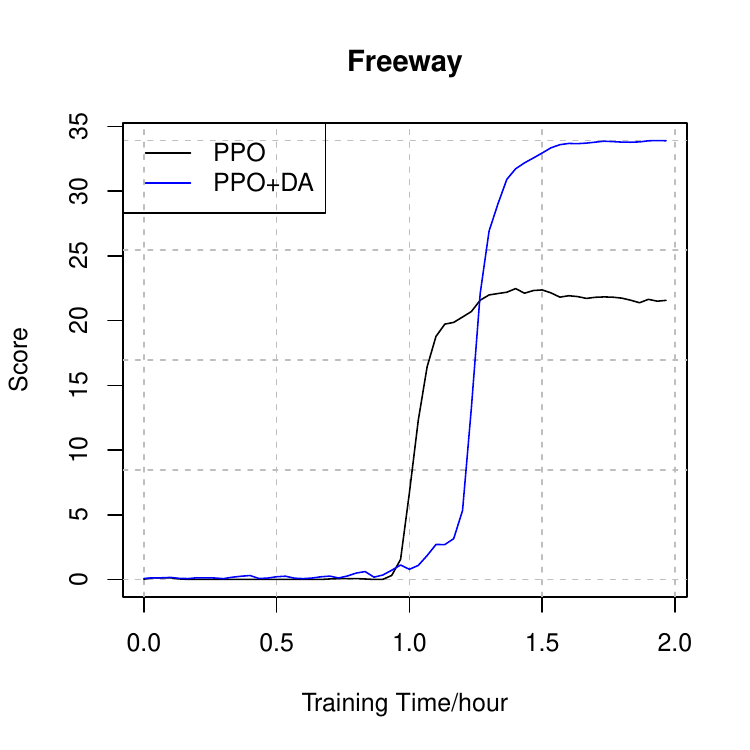}
\includegraphics[width=0.245 \textwidth]{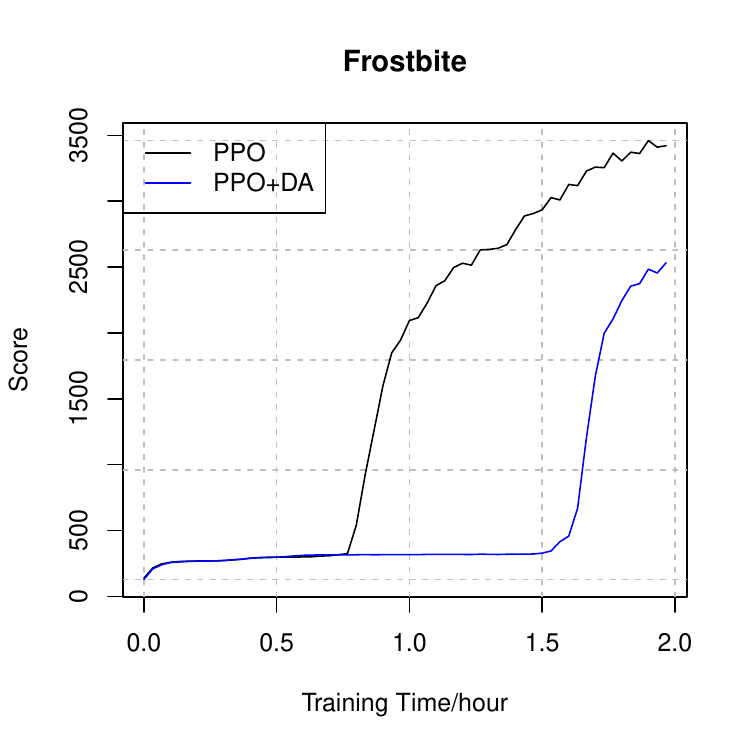}
\end{figure*}

\begin{figure*}
\includegraphics[width=0.245 \textwidth]{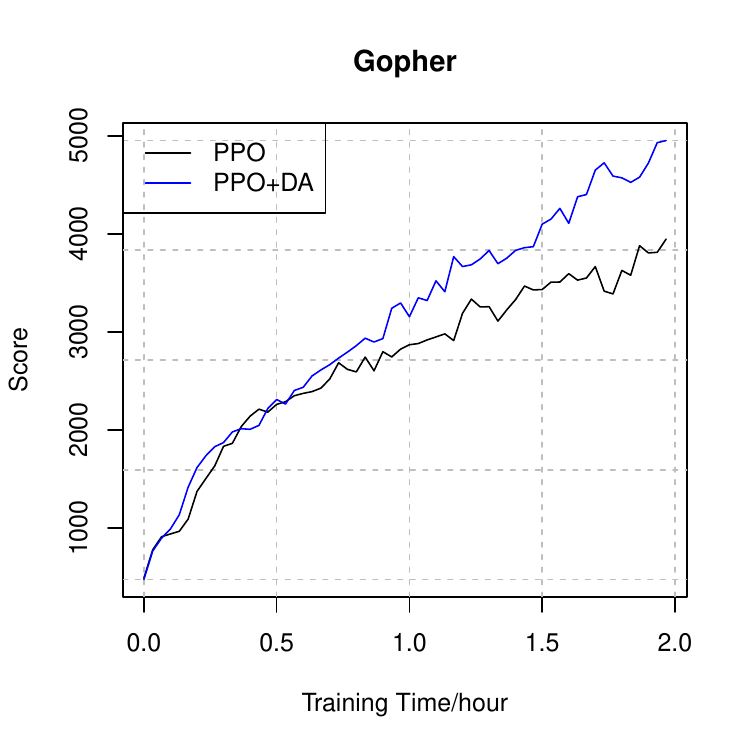}
\includegraphics[width=0.245 \textwidth]{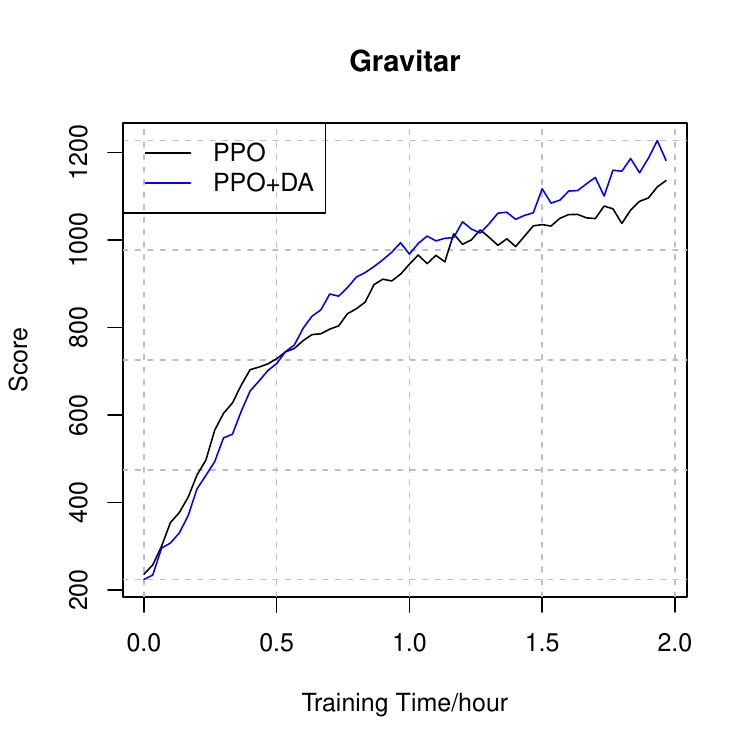}
\includegraphics[width=0.245 \textwidth]{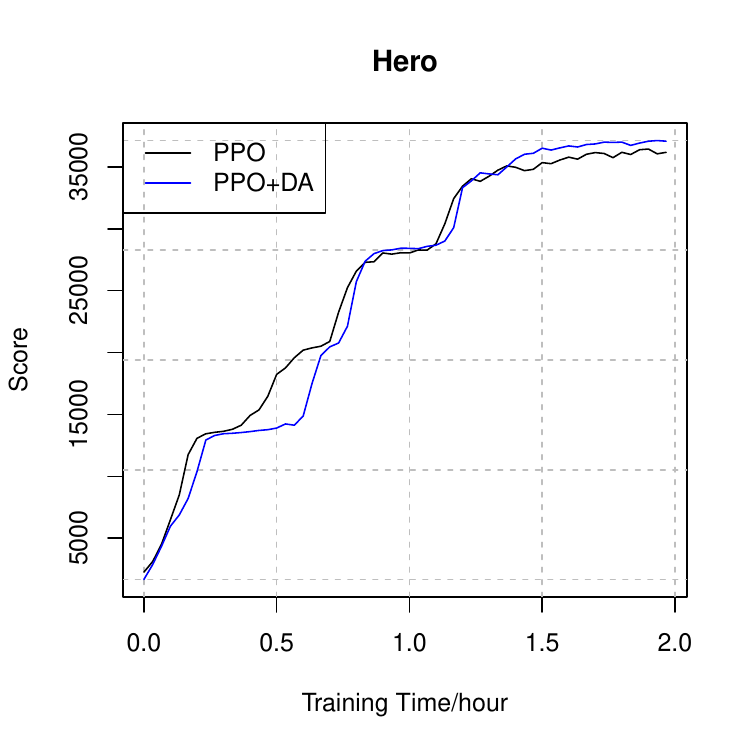}
\includegraphics[width=0.245 \textwidth]{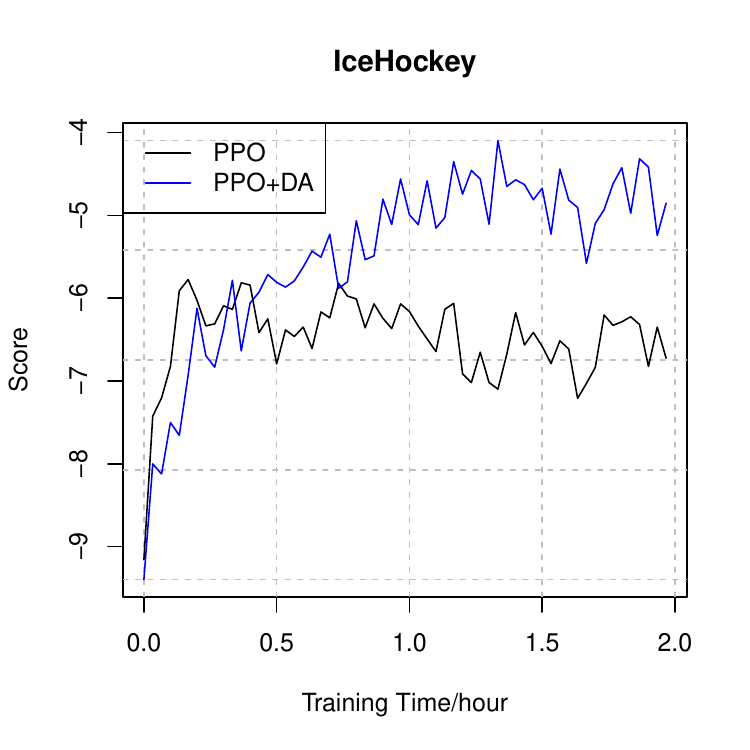}
\includegraphics[width=0.245 \textwidth]{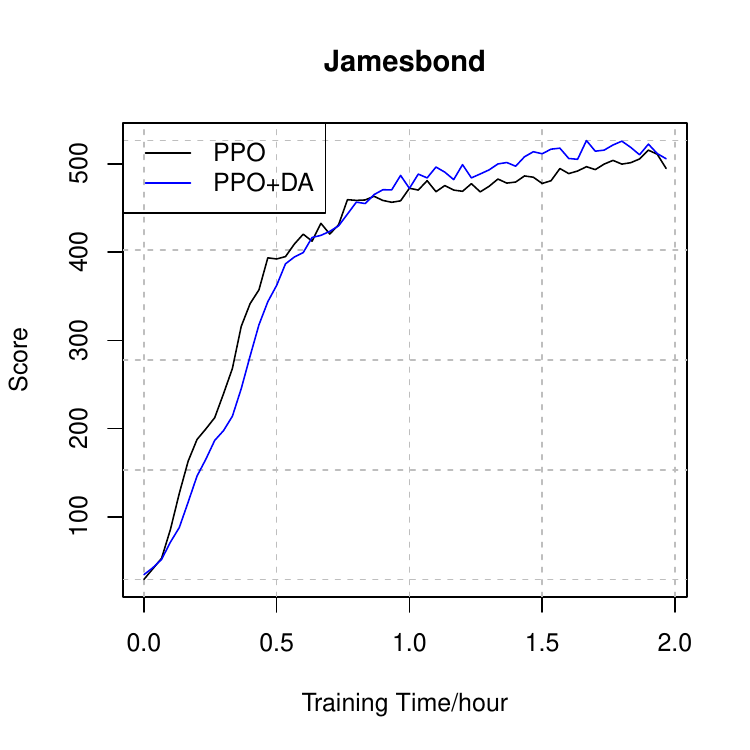}
\includegraphics[width=0.245 \textwidth]{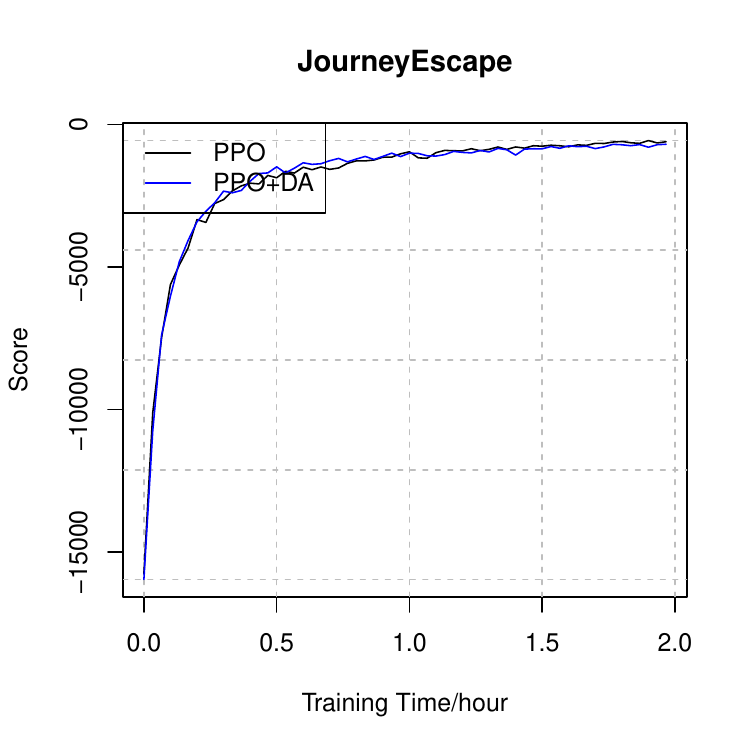}
\includegraphics[width=0.245 \textwidth]{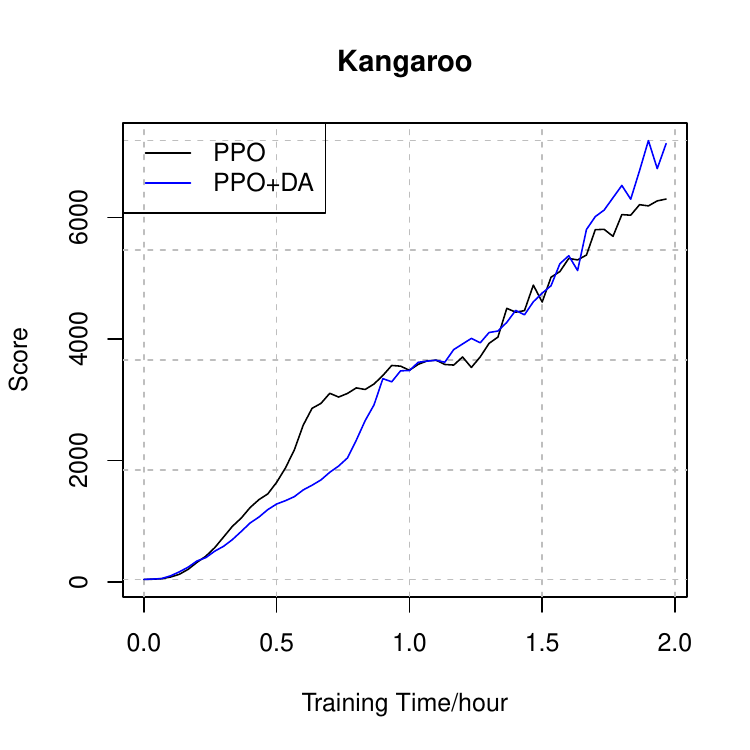}
\includegraphics[width=0.245 \textwidth]{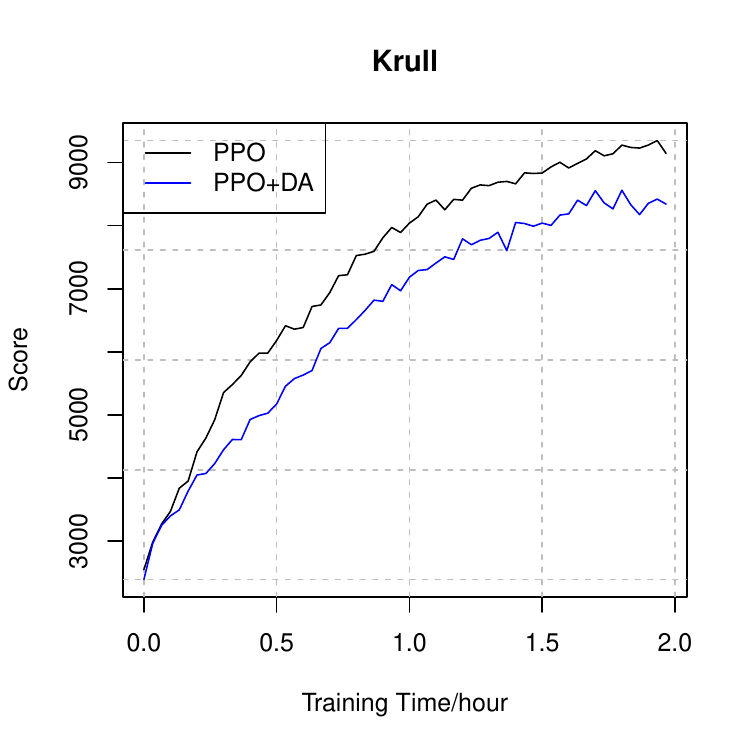}
\includegraphics[width=0.245 \textwidth]{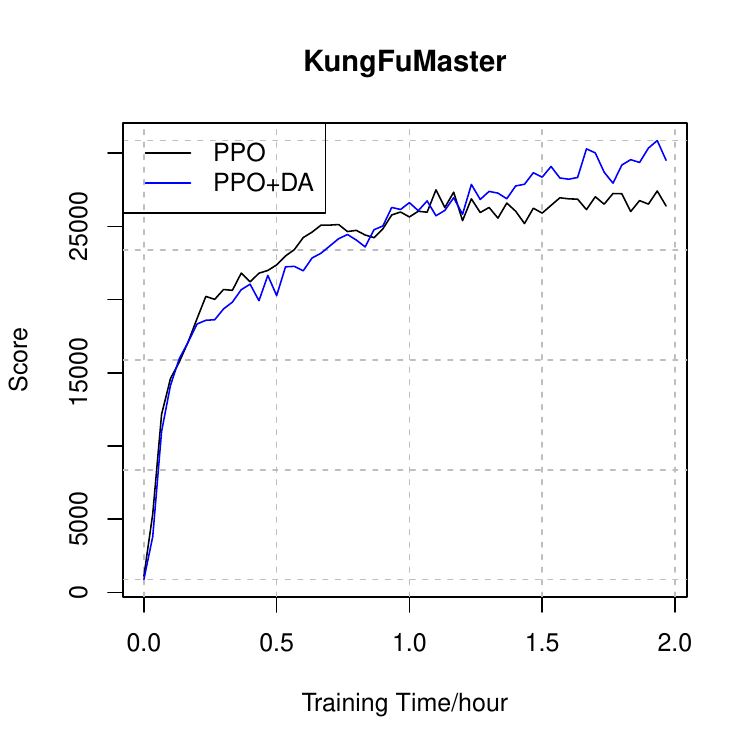}
\includegraphics[width=0.245 \textwidth]{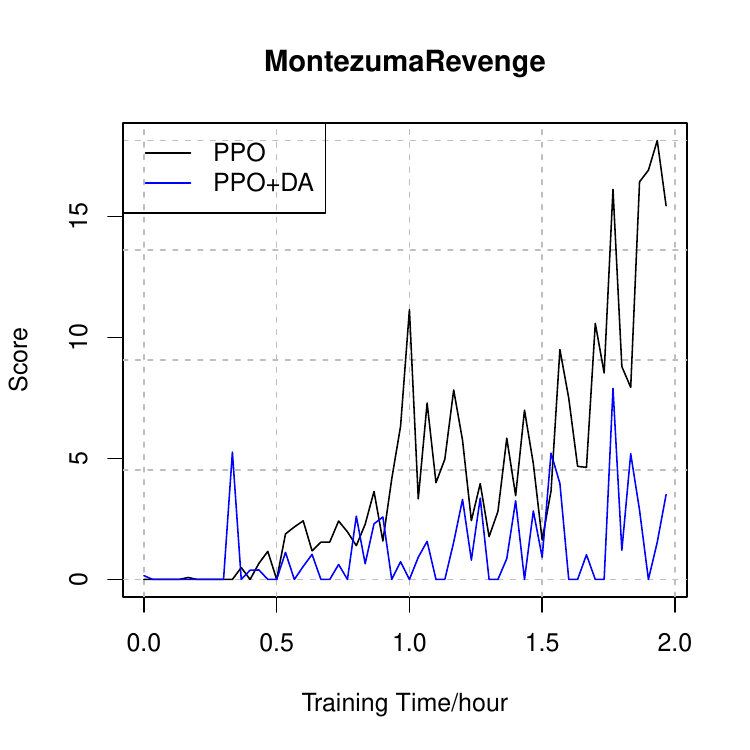}
\includegraphics[width=0.245 \textwidth]{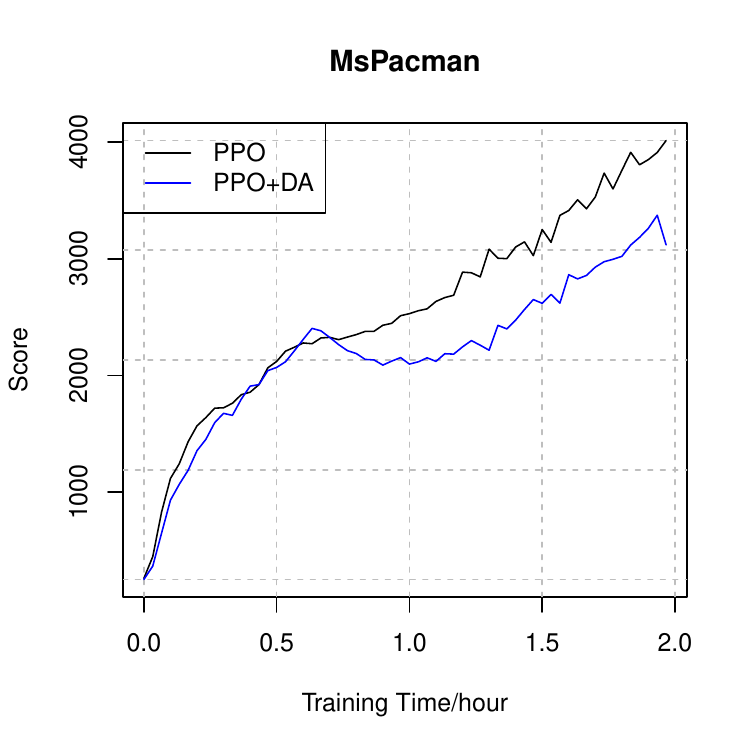}
\includegraphics[width=0.245 \textwidth]{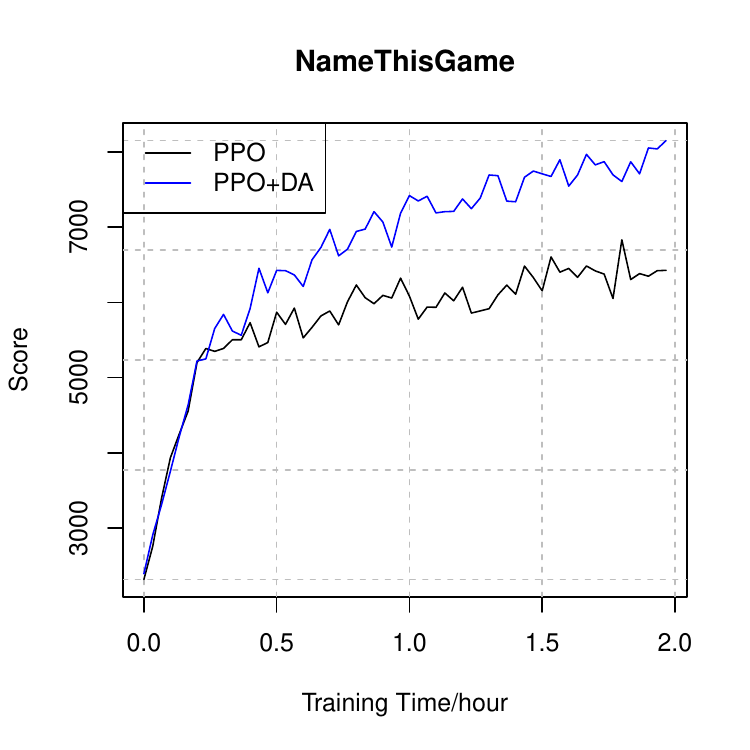}
\includegraphics[width=0.245 \textwidth]{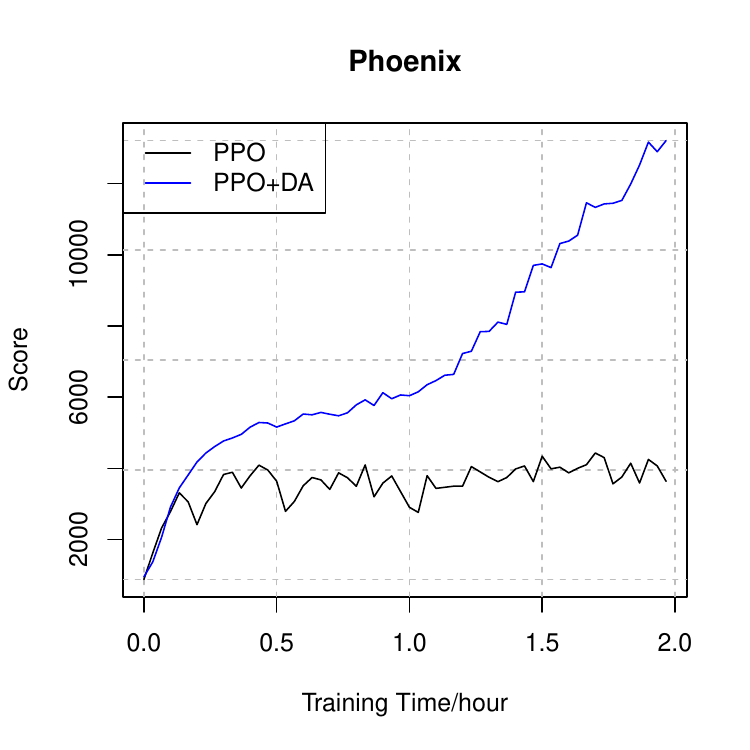}
\includegraphics[width=0.245 \textwidth]{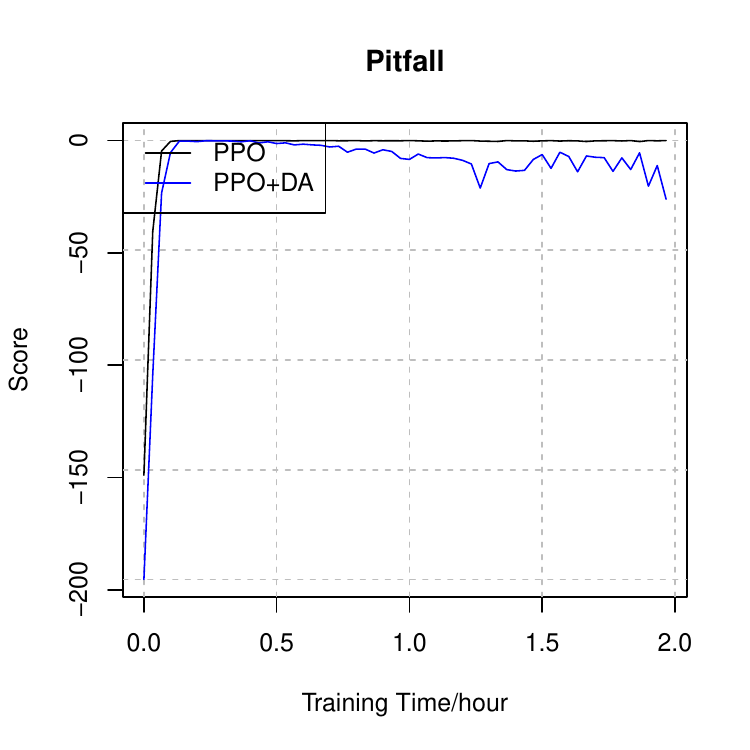}
\includegraphics[width=0.245 \textwidth]{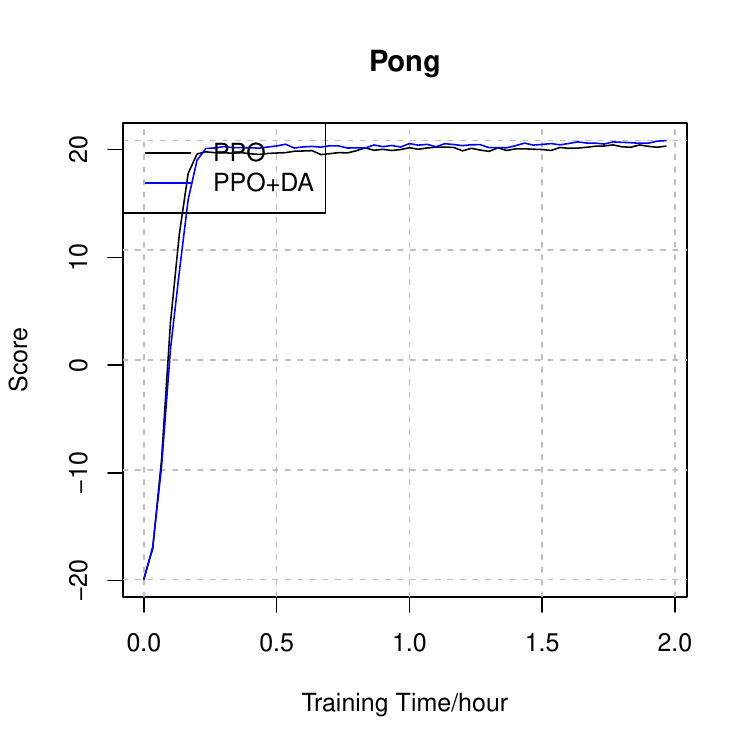}
\includegraphics[width=0.245 \textwidth]{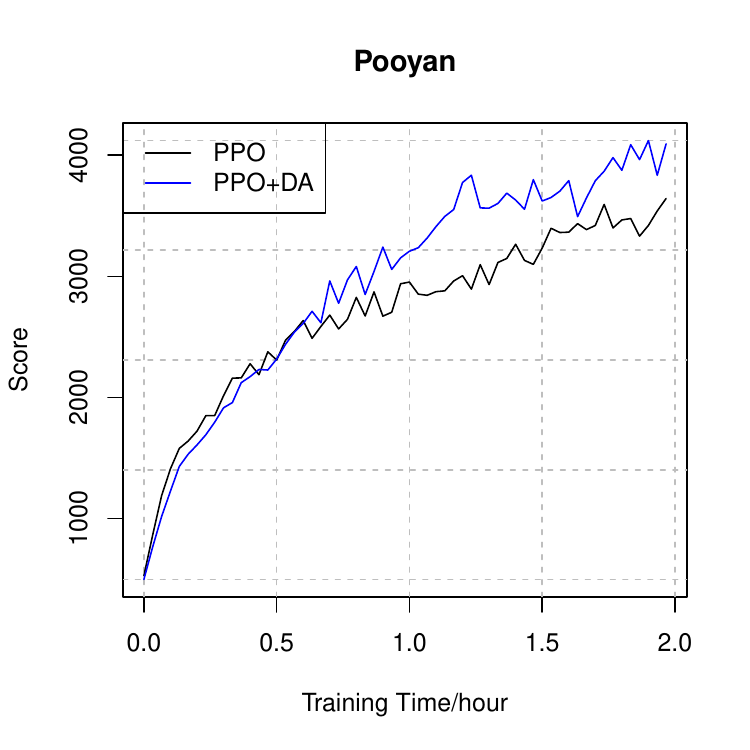}
\includegraphics[width=0.245 \textwidth]{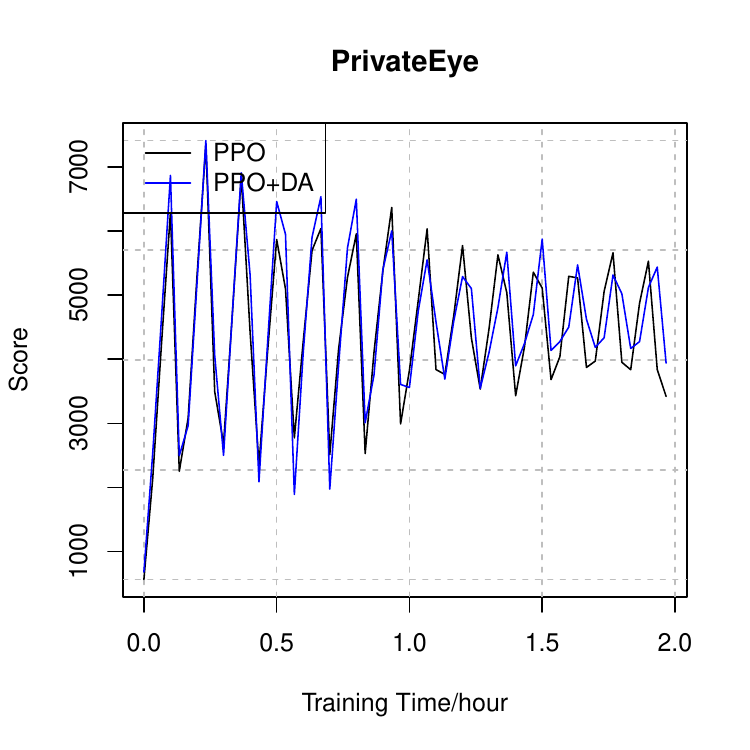}
\includegraphics[width=0.245 \textwidth]{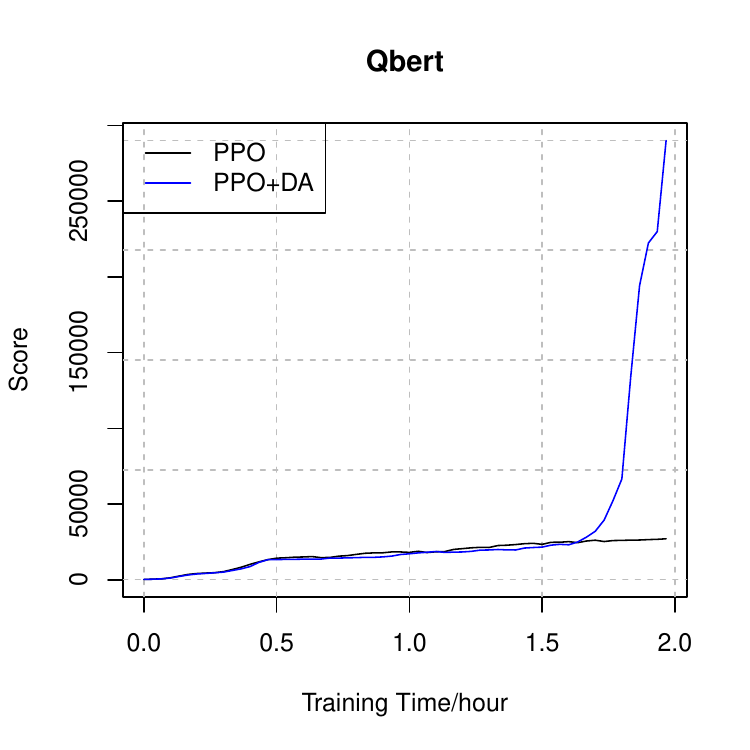}
\includegraphics[width=0.245 \textwidth]{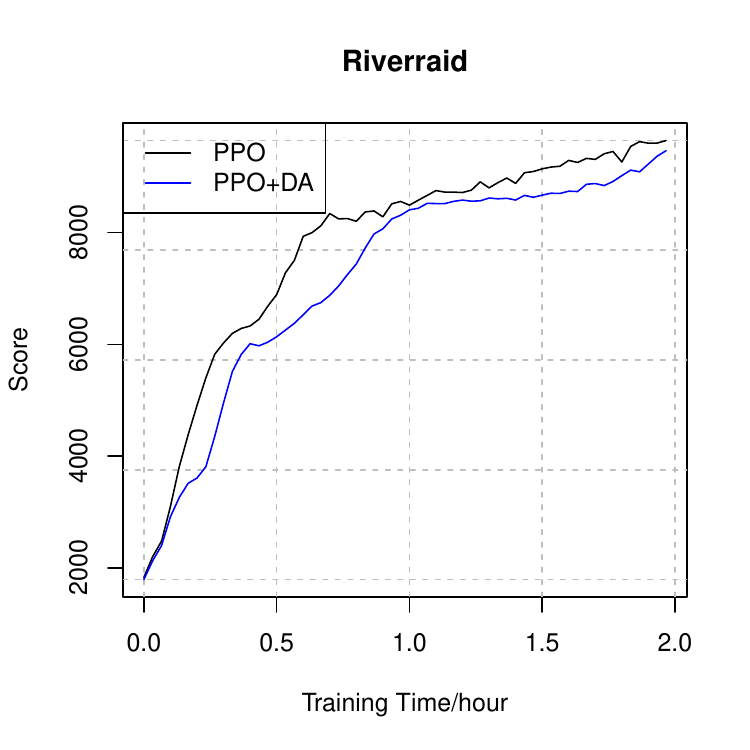}
\includegraphics[width=0.245 \textwidth]{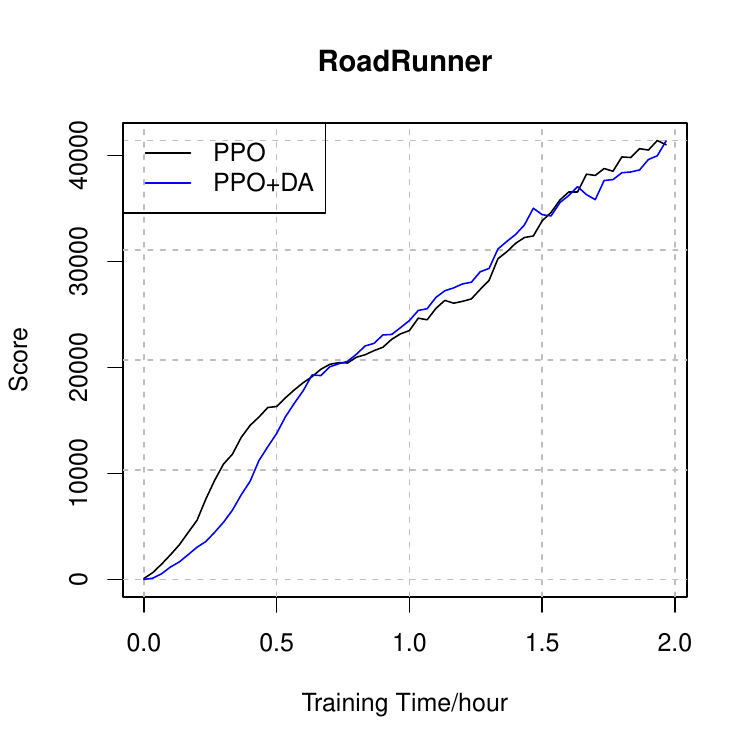}
\includegraphics[width=0.245 \textwidth]{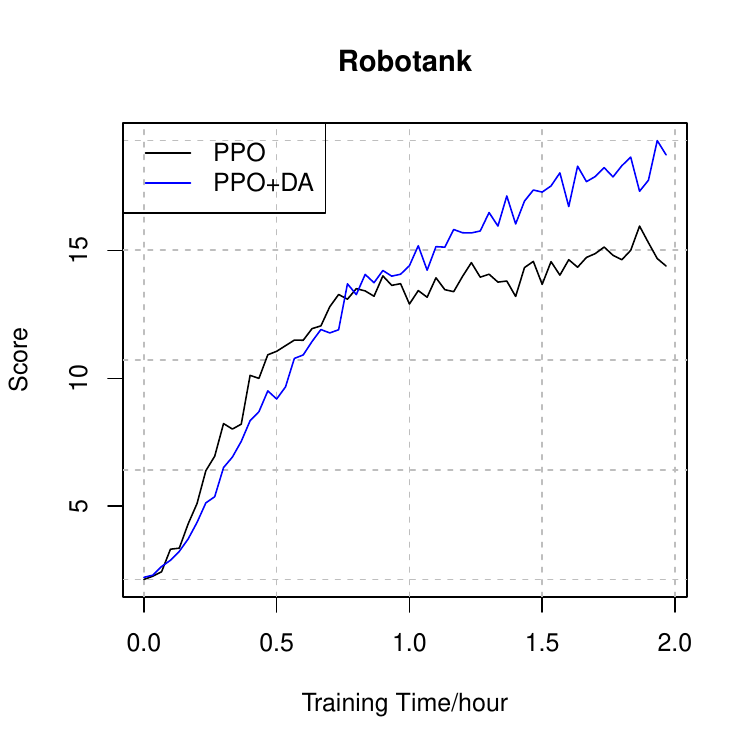}
\includegraphics[width=0.245 \textwidth]{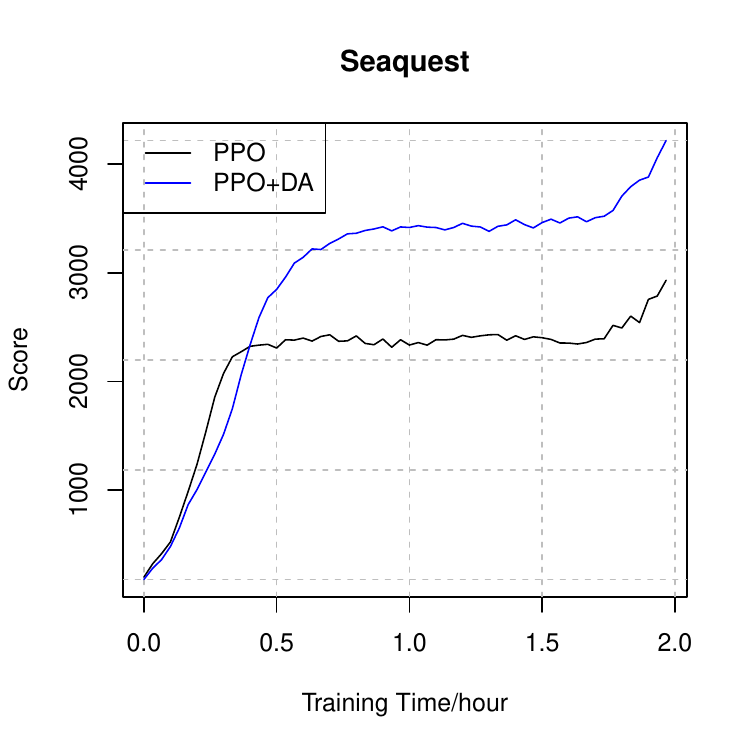}
\includegraphics[width=0.245 \textwidth]{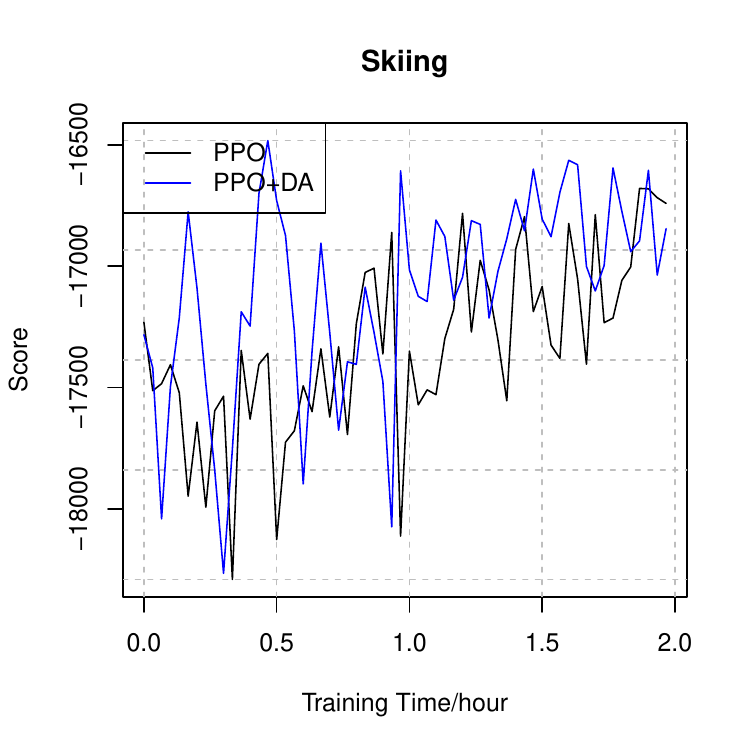}
\includegraphics[width=0.245 \textwidth]{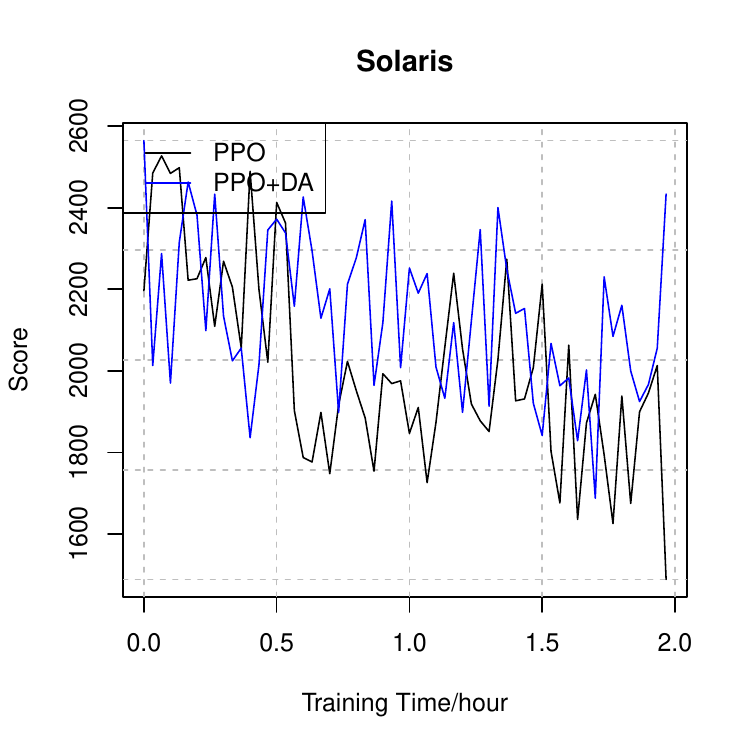}
\end{figure*}

\begin{figure*}[!t]
\includegraphics[width=0.245 \textwidth]{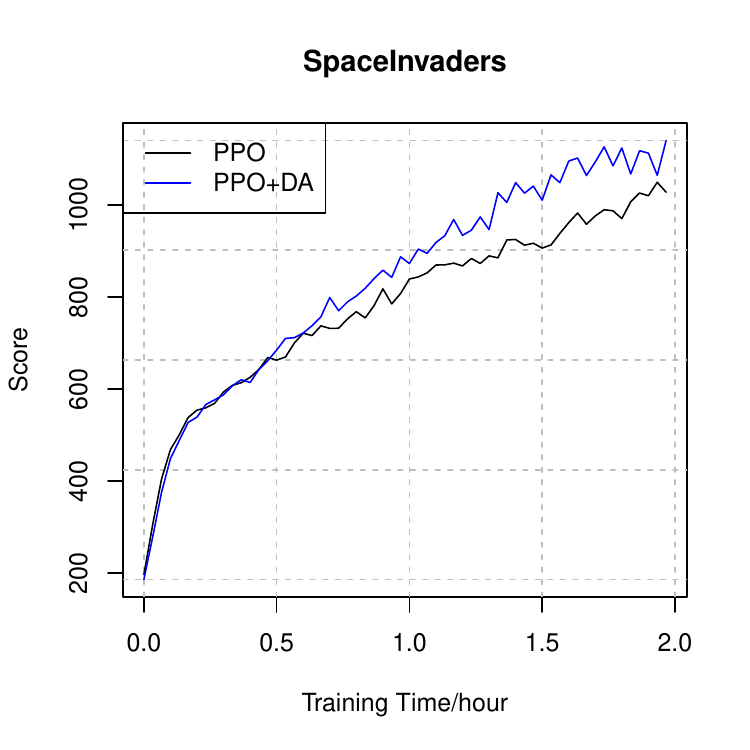}
\includegraphics[width=0.245 \textwidth]{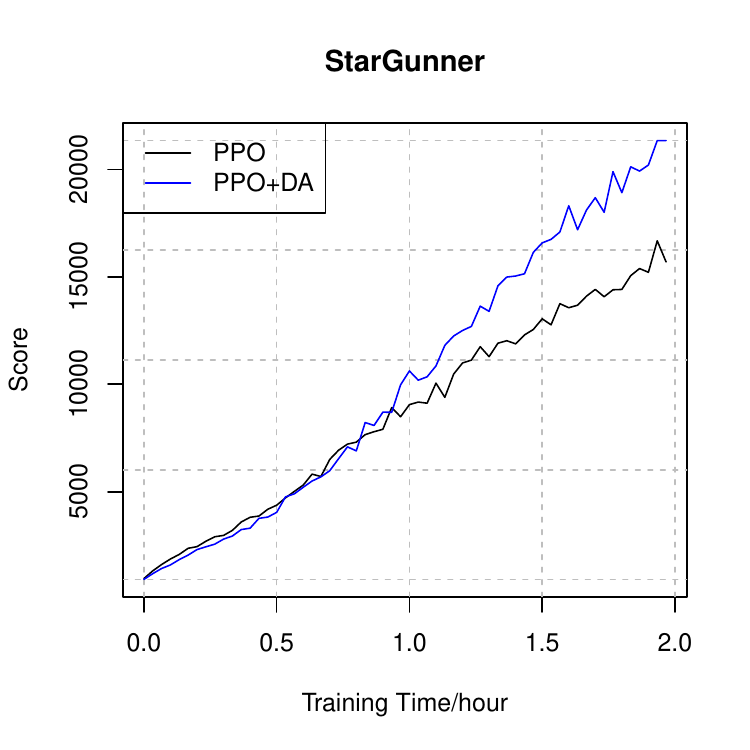}
\includegraphics[width=0.245 \textwidth]{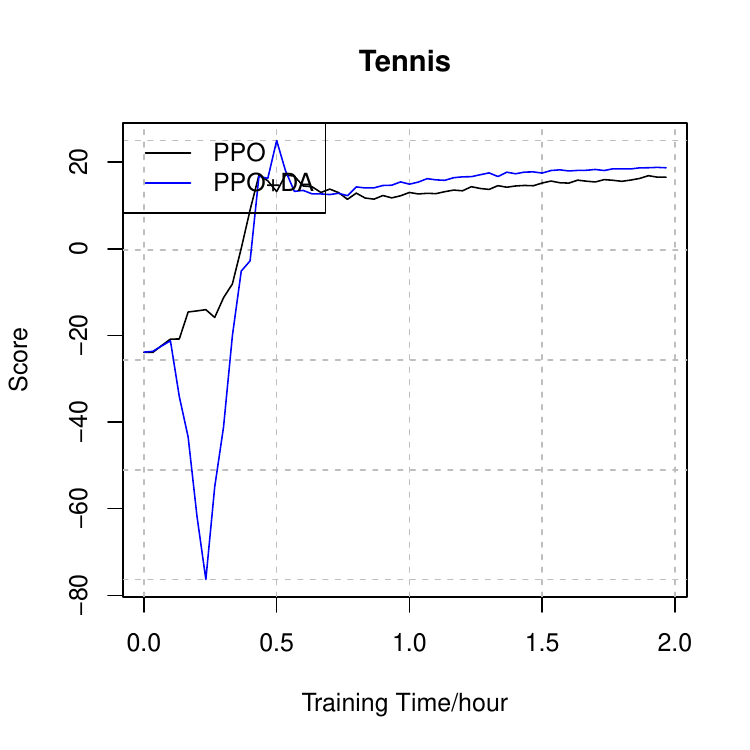}
\includegraphics[width=0.245 \textwidth]{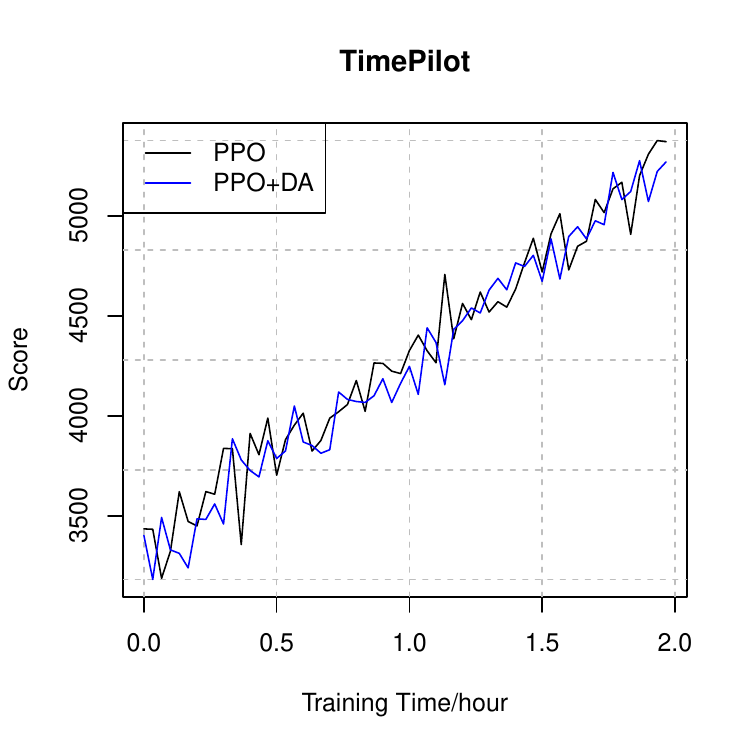}
\includegraphics[width=0.245 \textwidth]{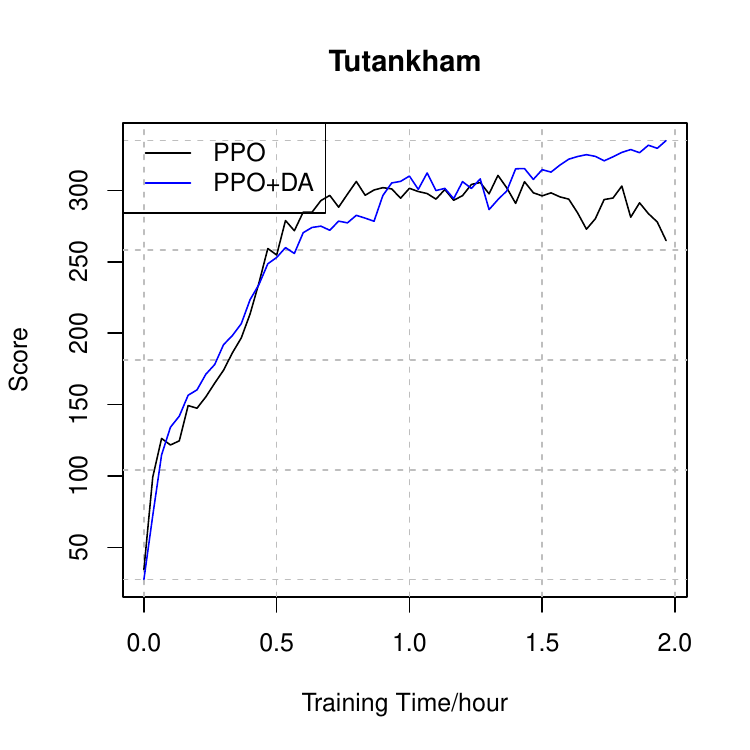}
\includegraphics[width=0.245 \textwidth]{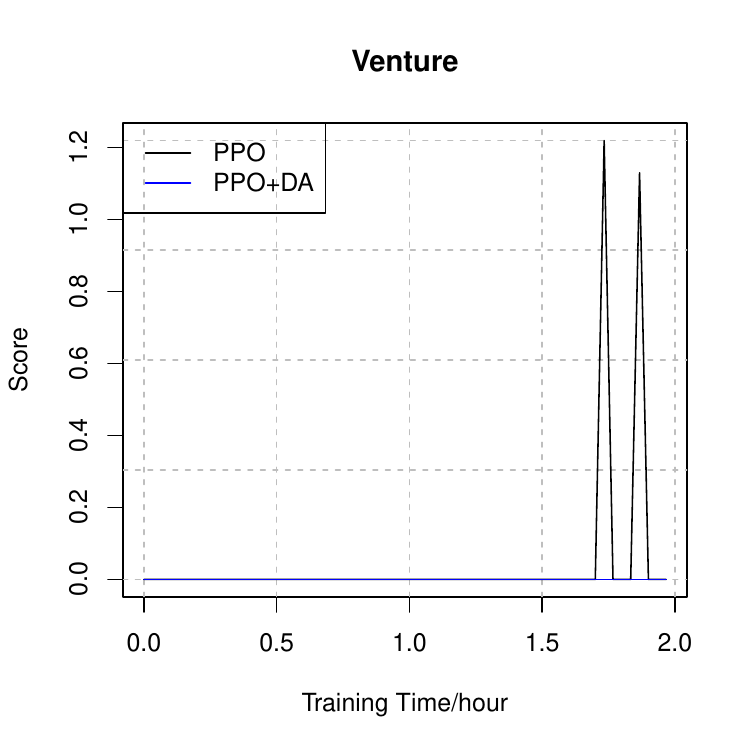}
\includegraphics[width=0.245 \textwidth]{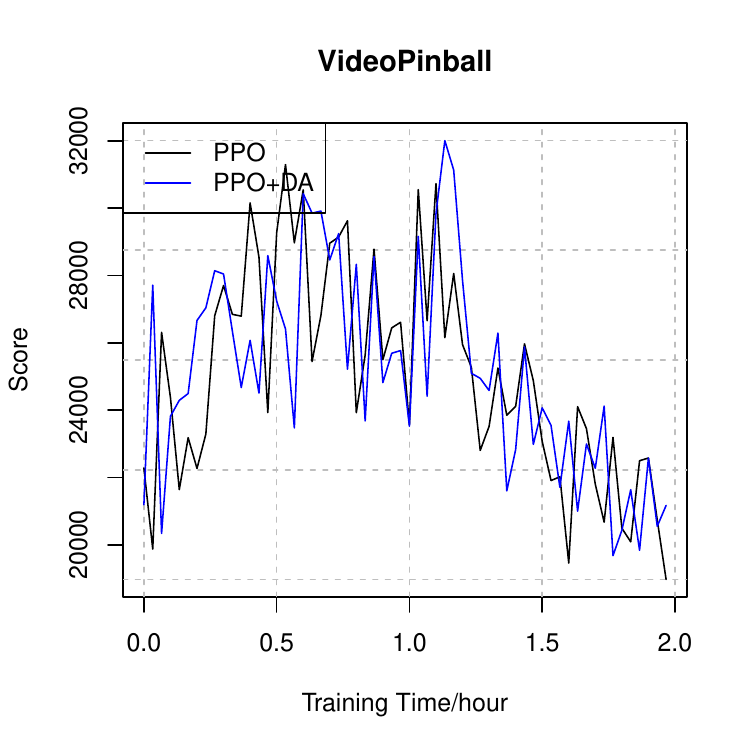}
\includegraphics[width=0.245 \textwidth]{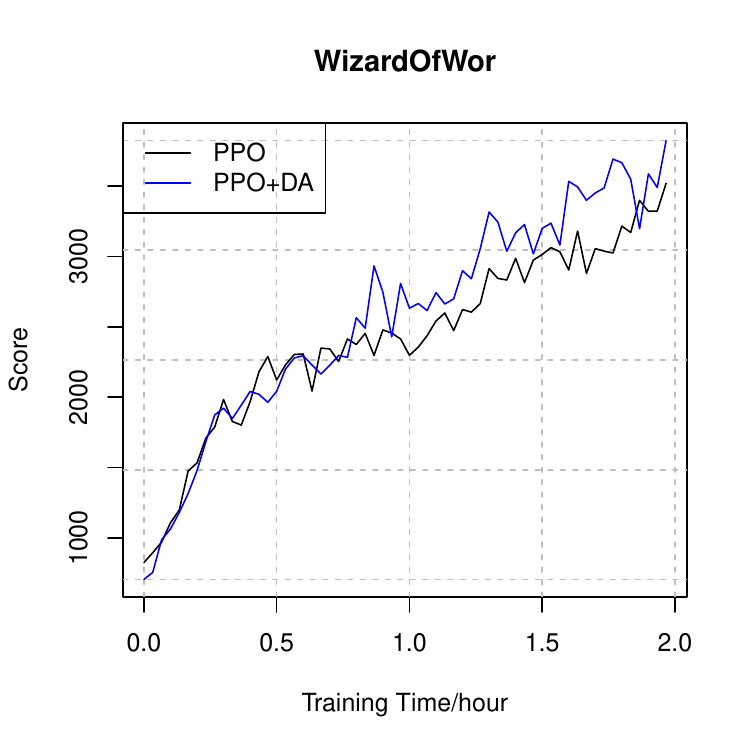}
\includegraphics[width=0.245 \textwidth]{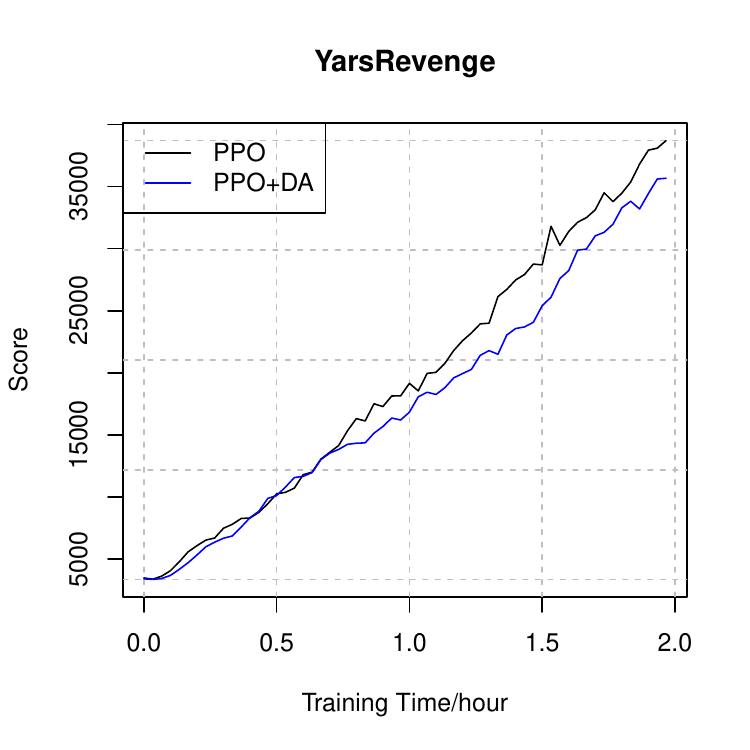}
\includegraphics[width=0.245 \textwidth]{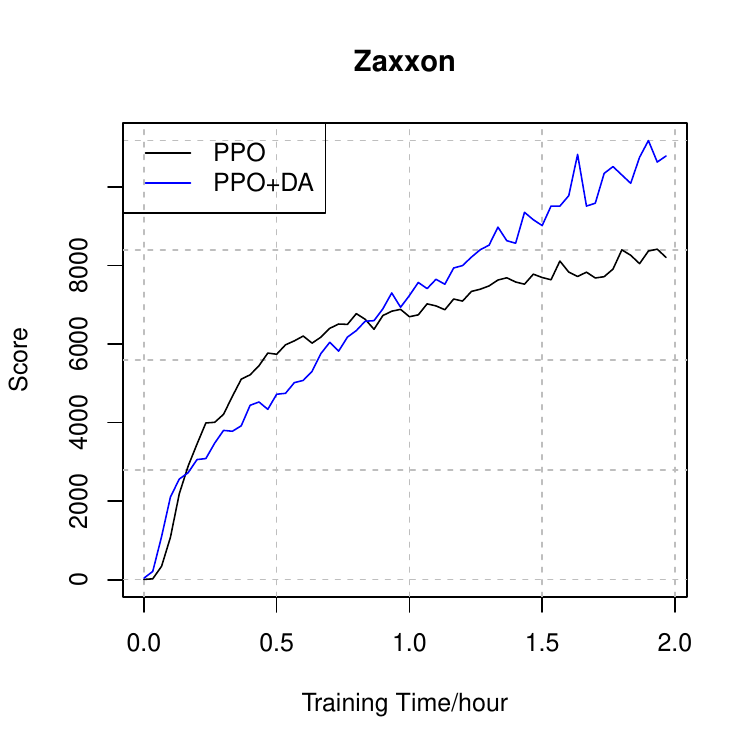}
  \caption{Performance comparison of PPO+DA with PPO on 58 Atari games. Each experiment is allowed to run for 2 hours as a limited time.}
  \label{fig:compare_full}
\end{figure*}
\newpage
\begin{figure*}
\includegraphics[width=0.245 \textwidth]{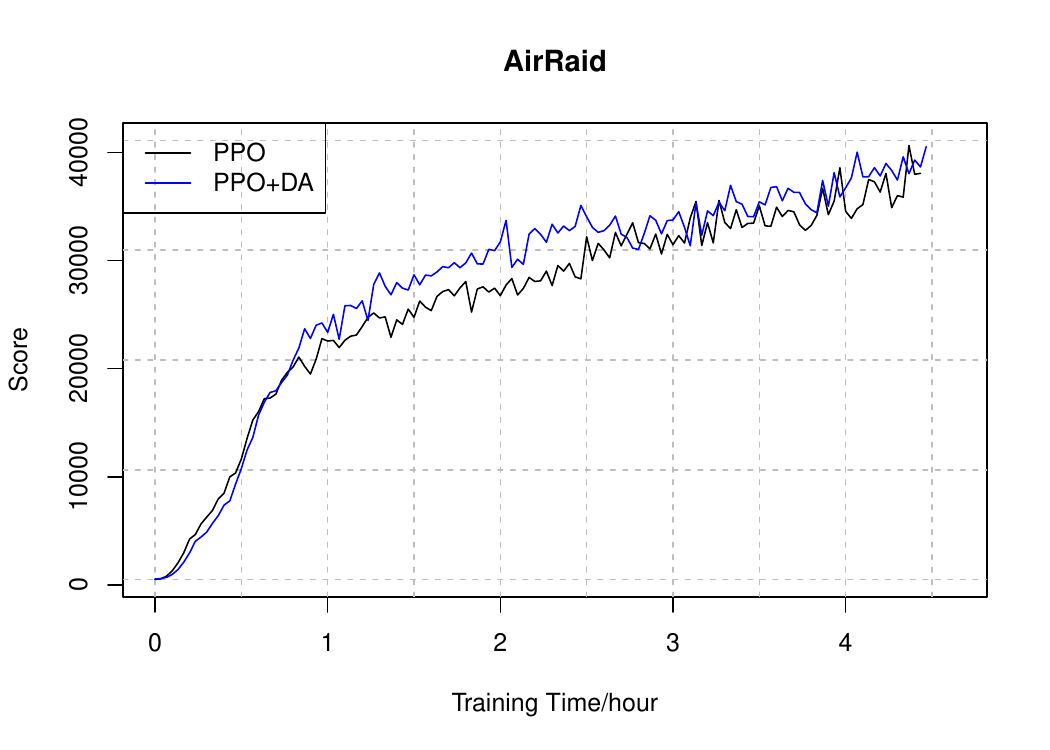}
\includegraphics[width=0.245 \textwidth]{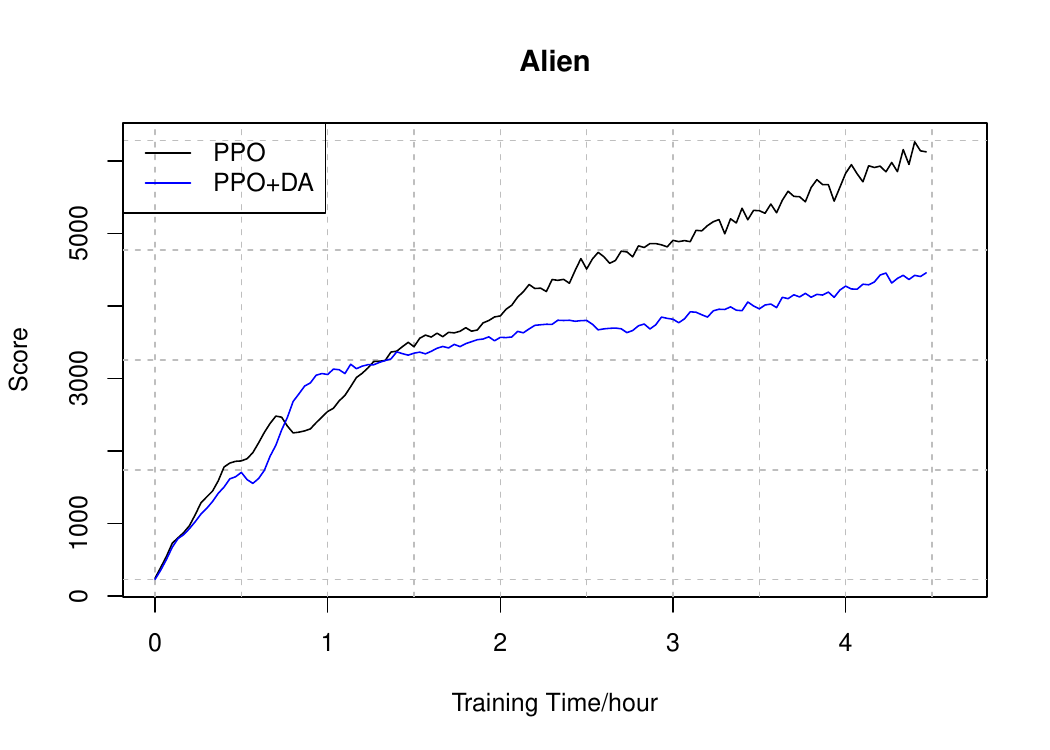}
\includegraphics[width=0.245 \textwidth]{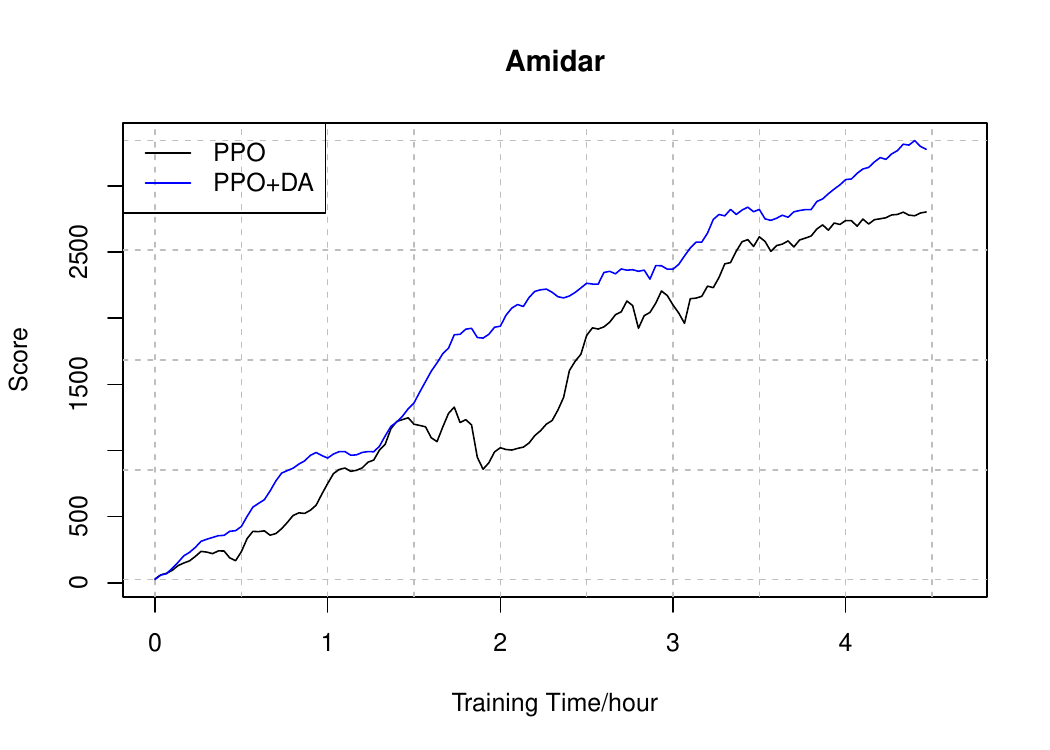}
\includegraphics[width=0.245 \textwidth]{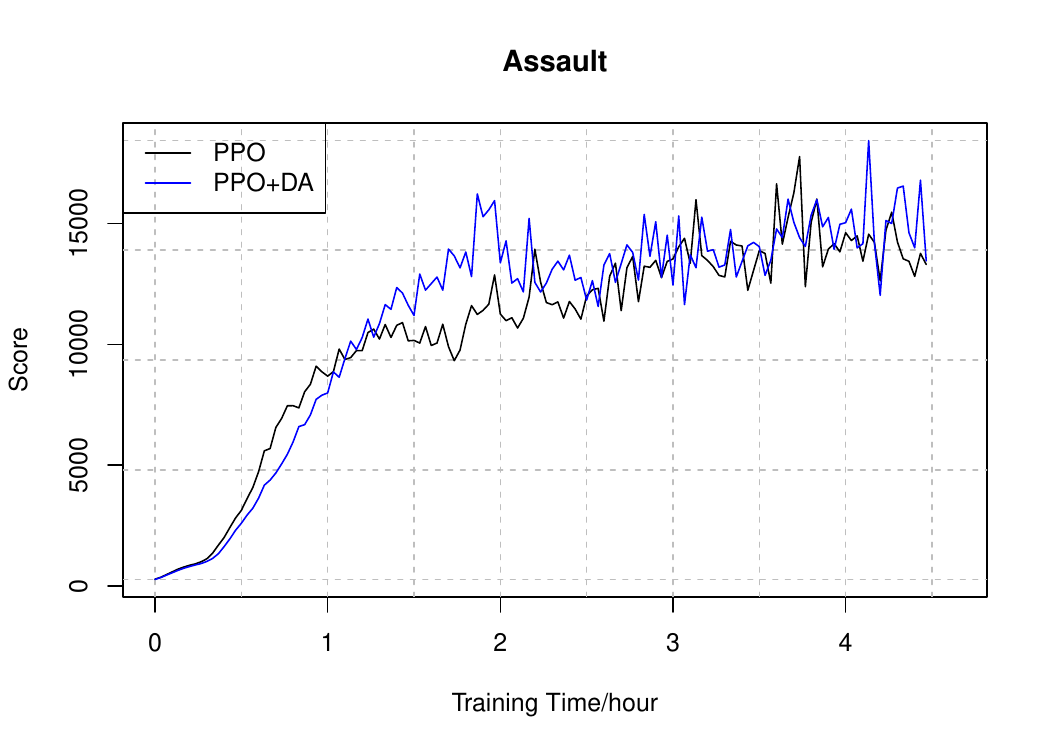}
\includegraphics[width=0.245 \textwidth]{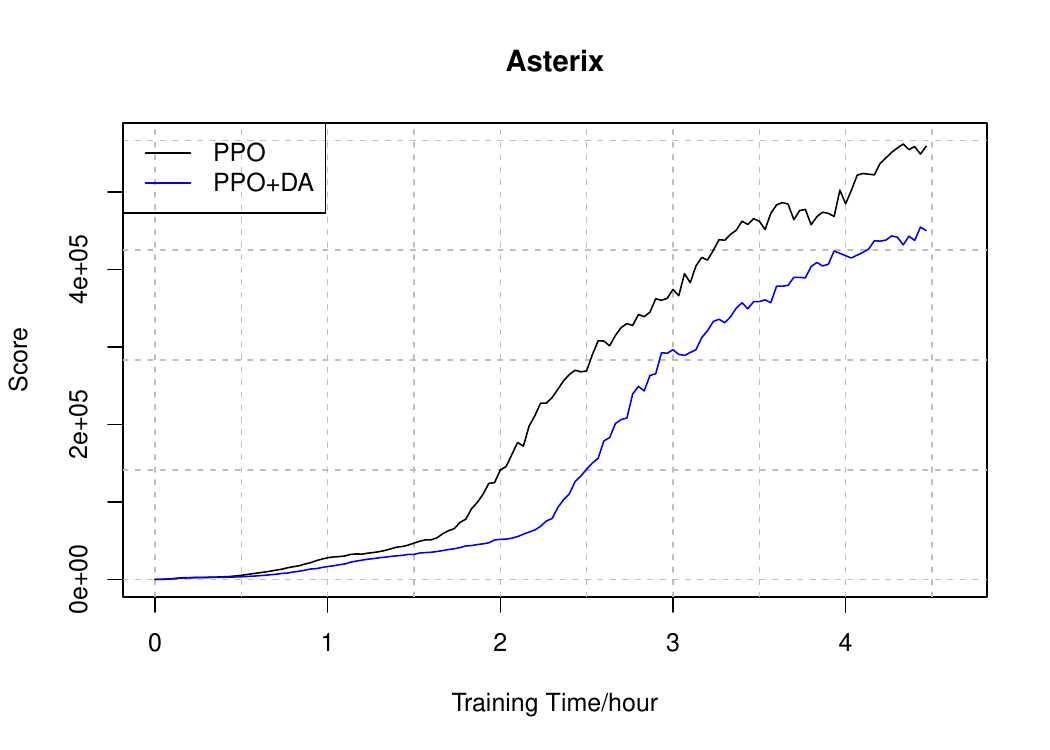}
\includegraphics[width=0.245 \textwidth]{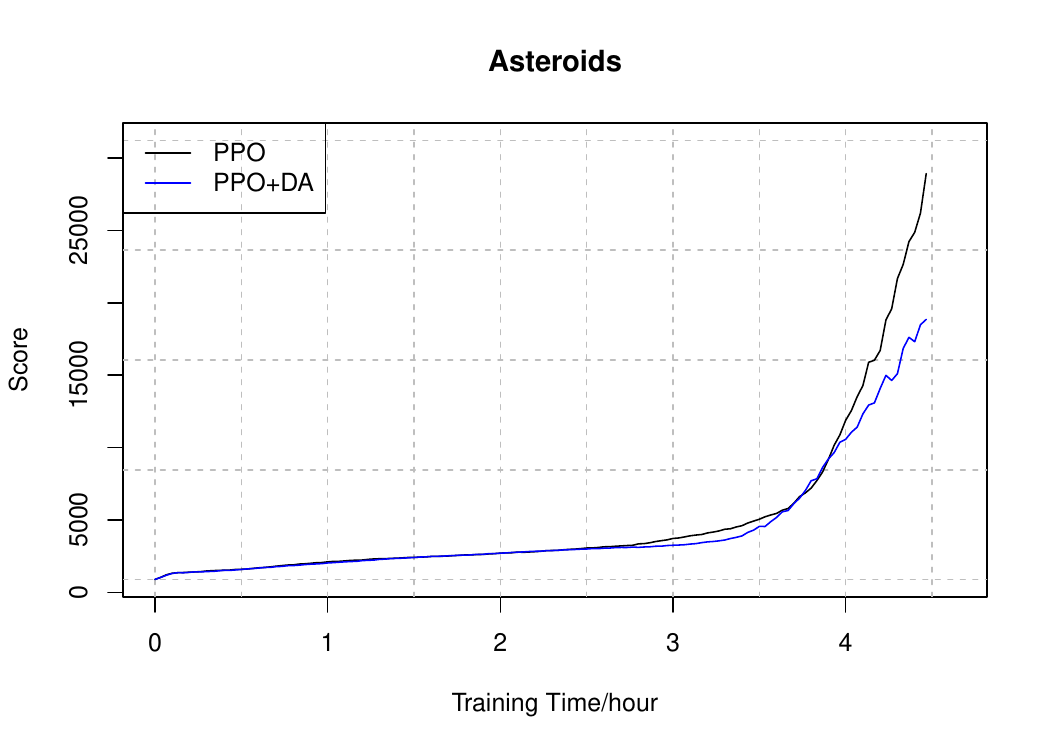}
\includegraphics[width=0.245 \textwidth]{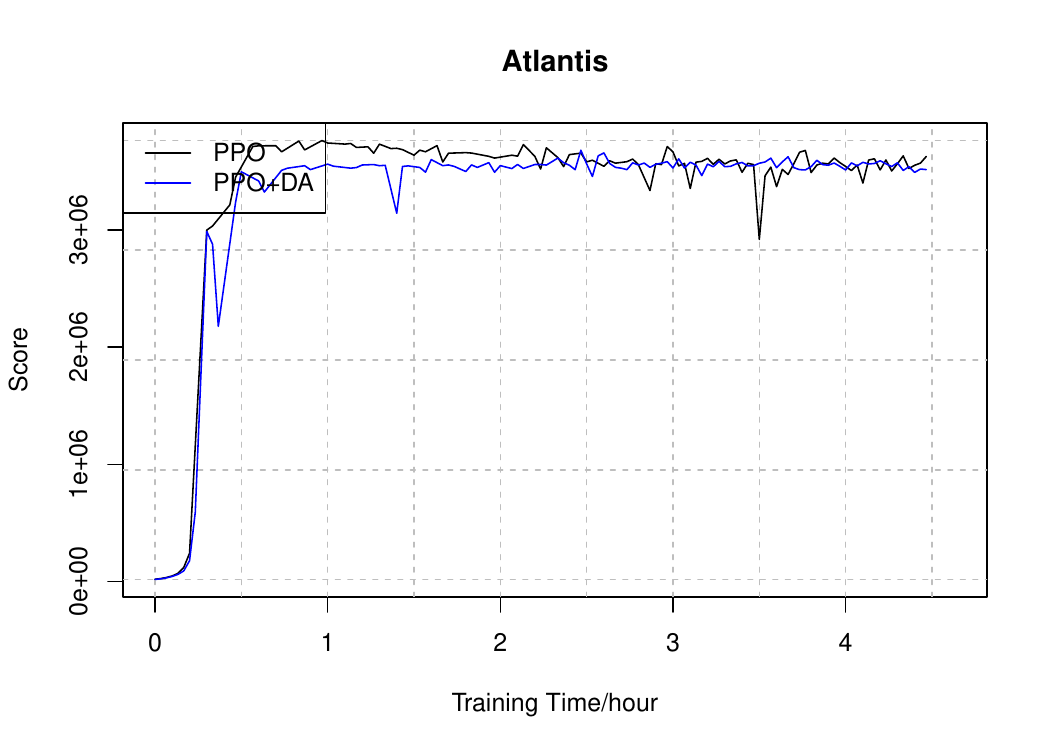}
\includegraphics[width=0.245 \textwidth]{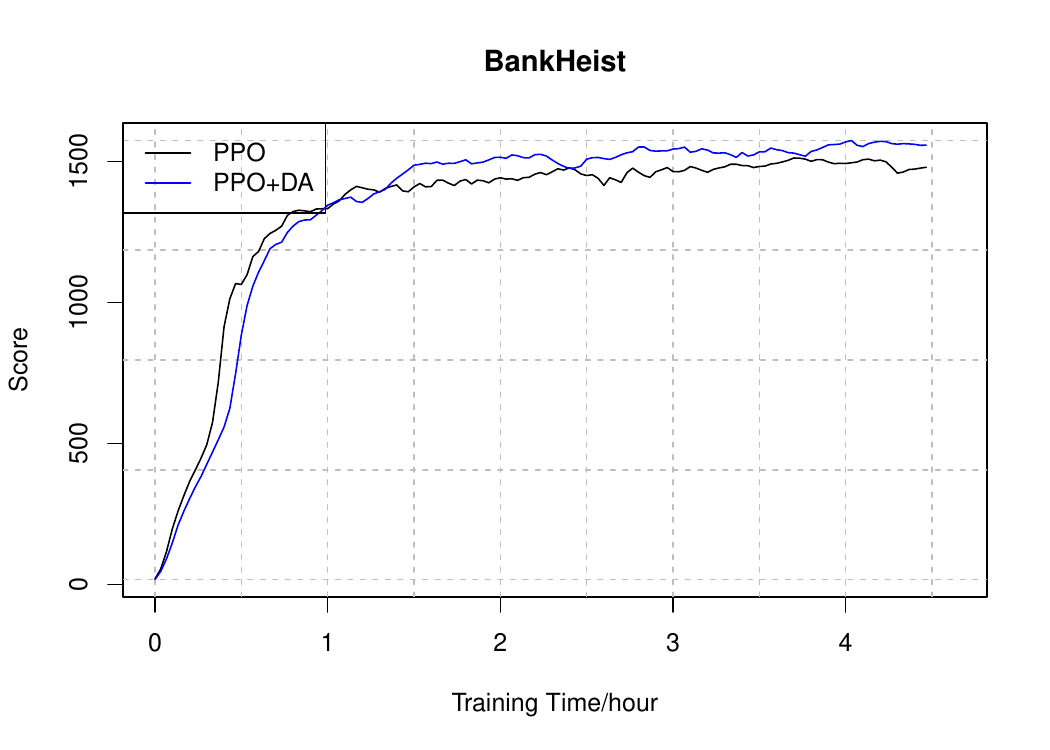}
\includegraphics[width=0.245 \textwidth]{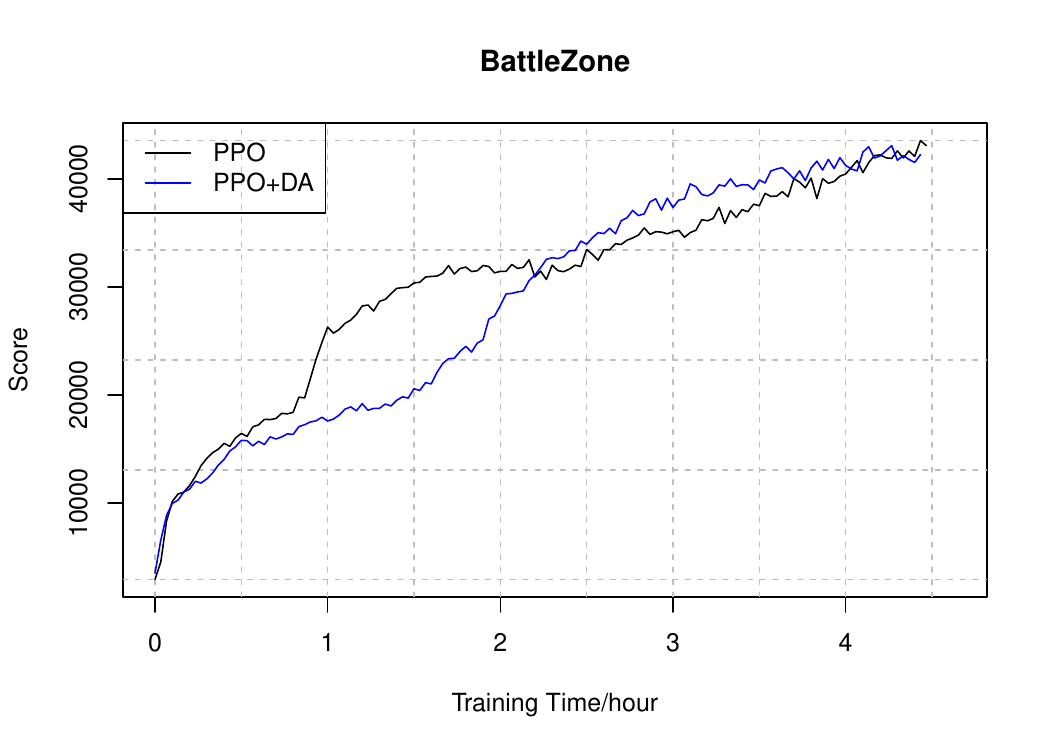}
\includegraphics[width=0.245 \textwidth]{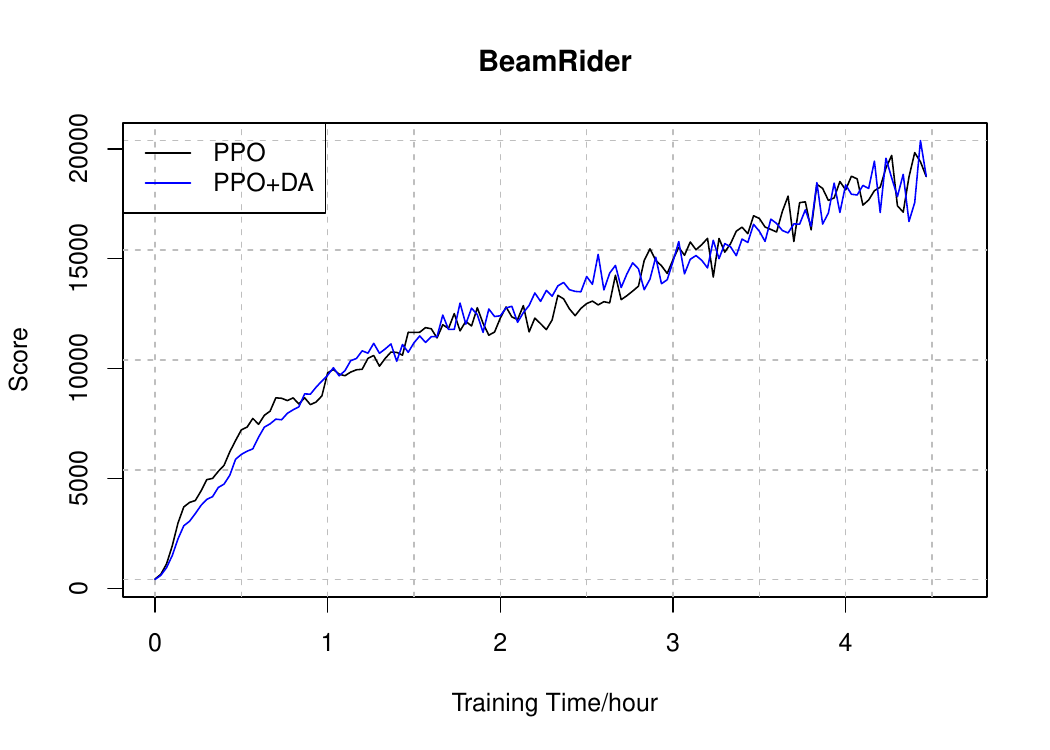}
\includegraphics[width=0.245 \textwidth]{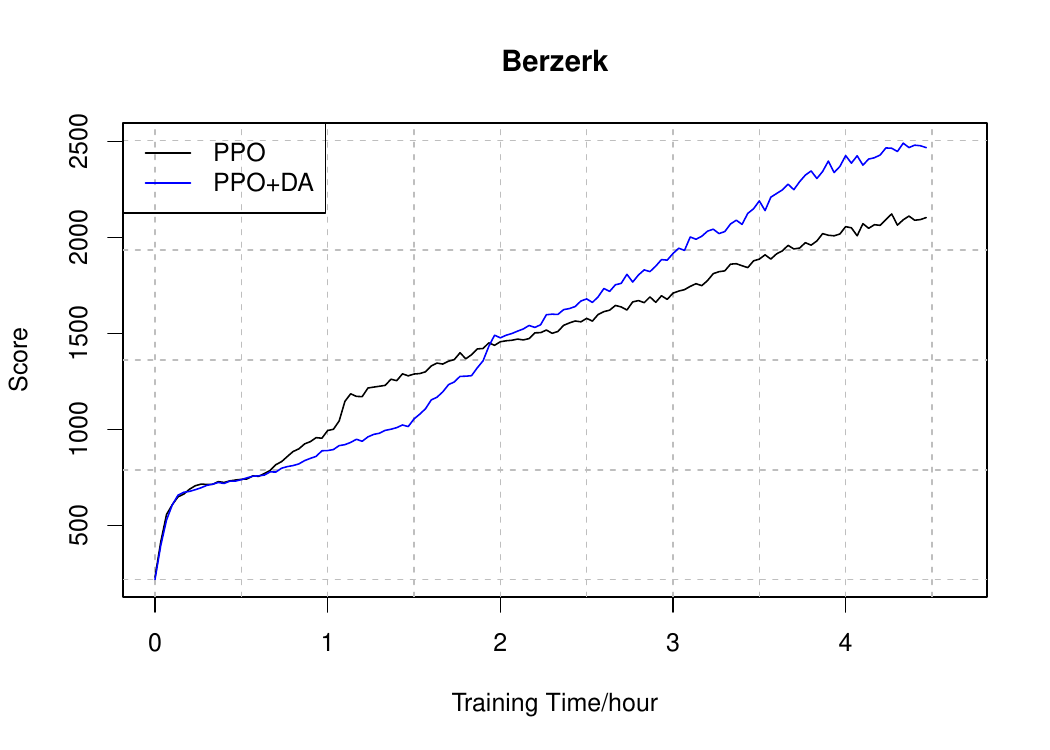}
\includegraphics[width=0.245 \textwidth]{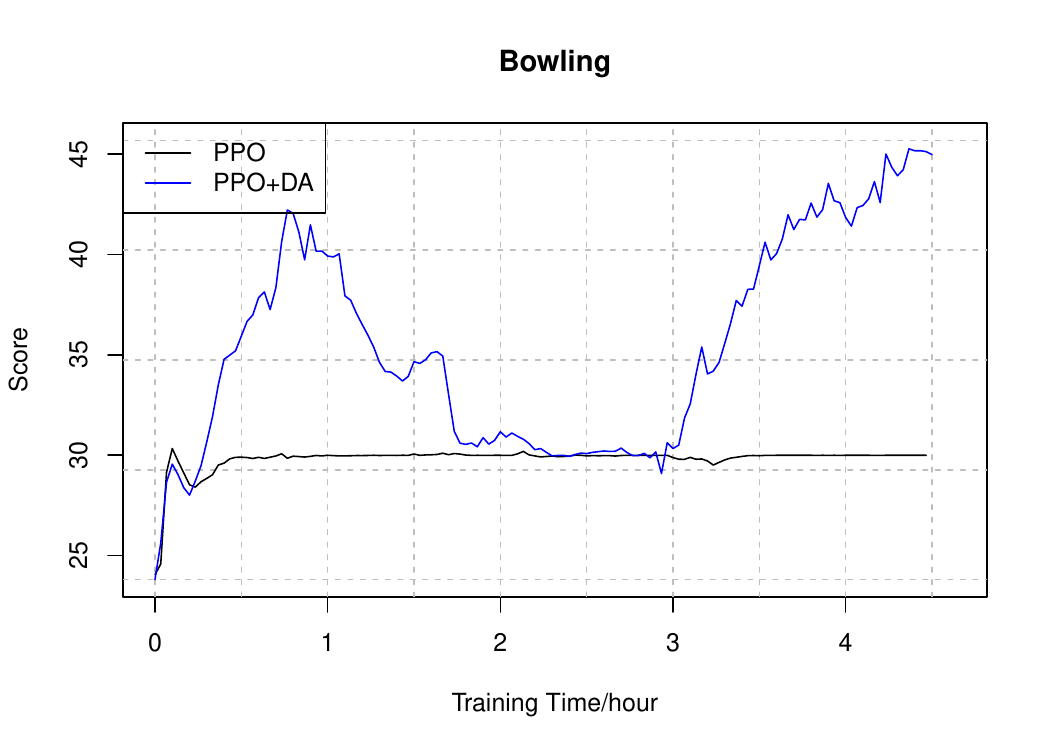}
\includegraphics[width=0.245 \textwidth]{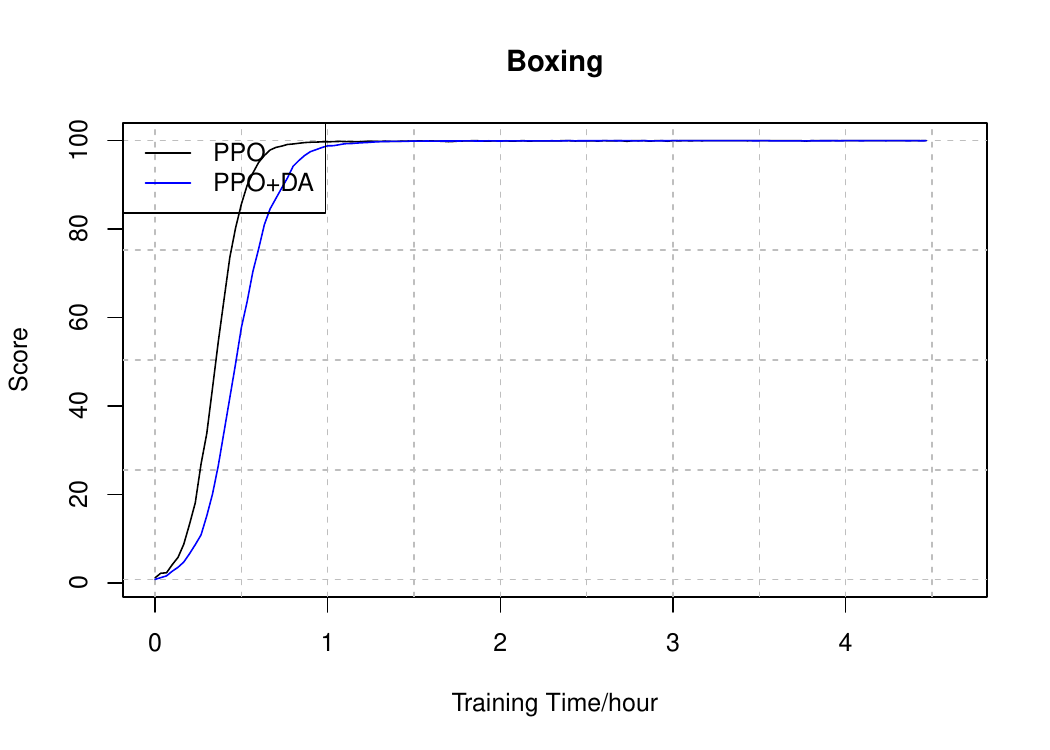}
\includegraphics[width=0.245 \textwidth]{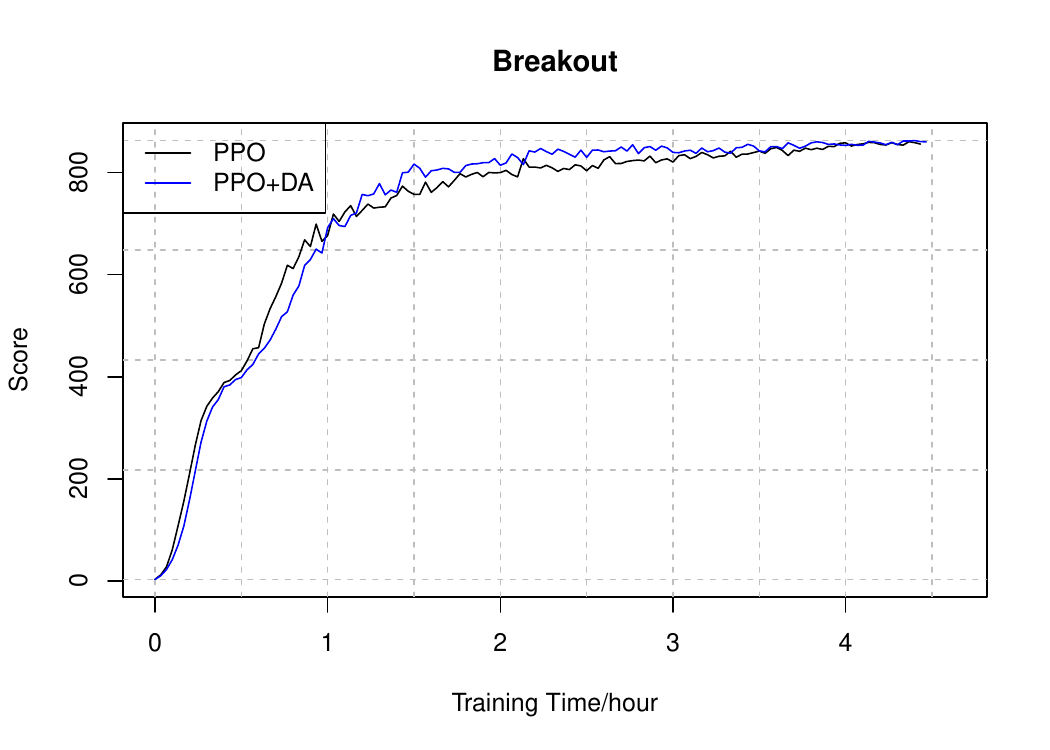}
\includegraphics[width=0.245 \textwidth]{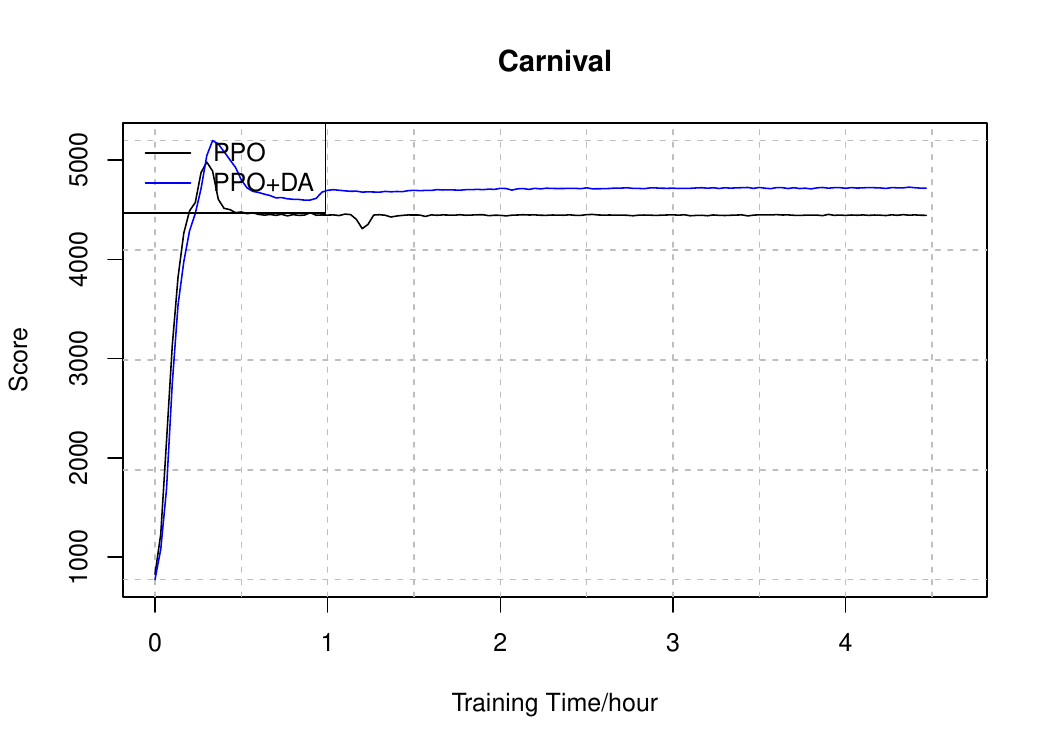}
\includegraphics[width=0.245 \textwidth]{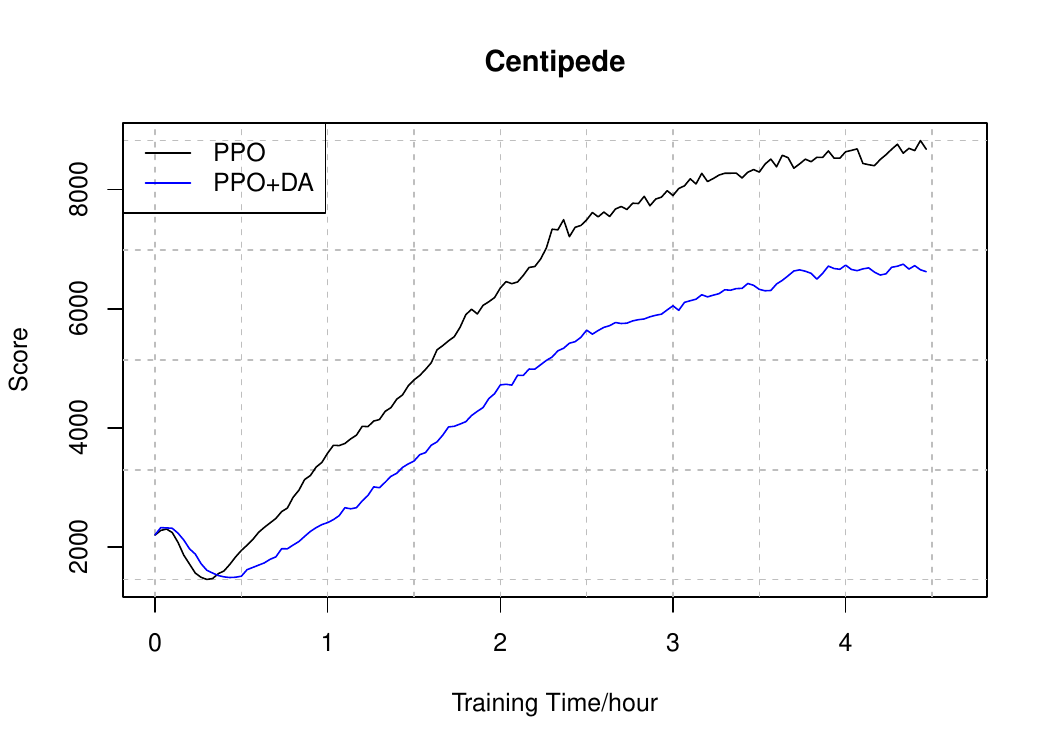}
\includegraphics[width=0.245 \textwidth]{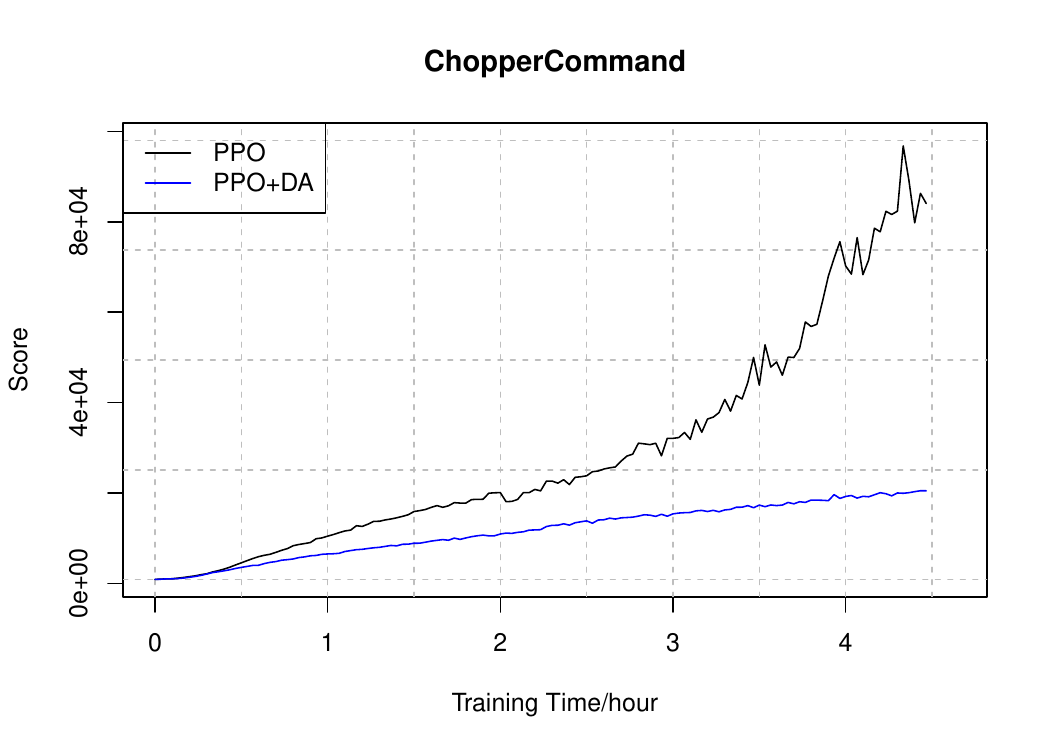}
\includegraphics[width=0.245 \textwidth]{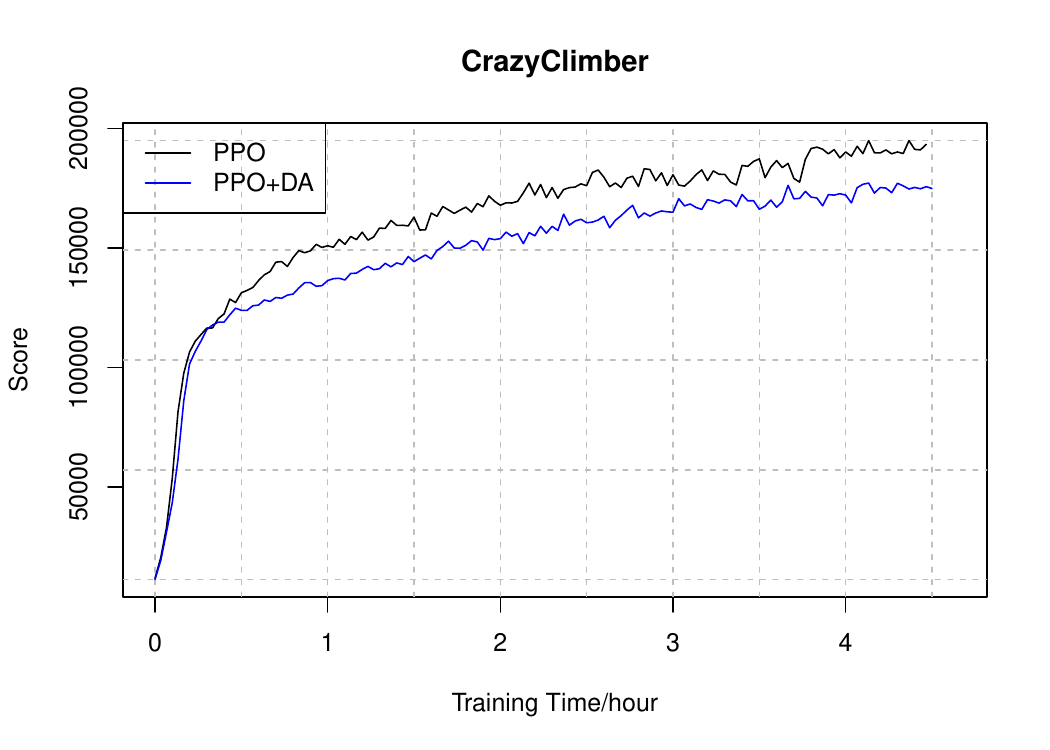}
\includegraphics[width=0.245 \textwidth]{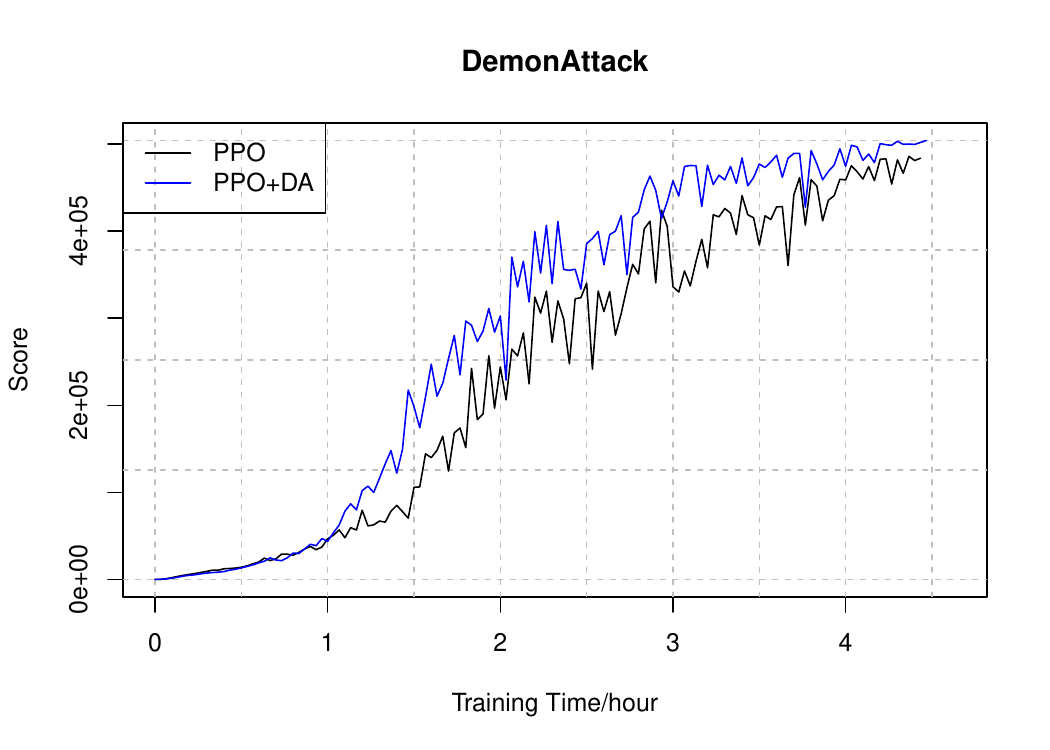}
\includegraphics[width=0.245 \textwidth]{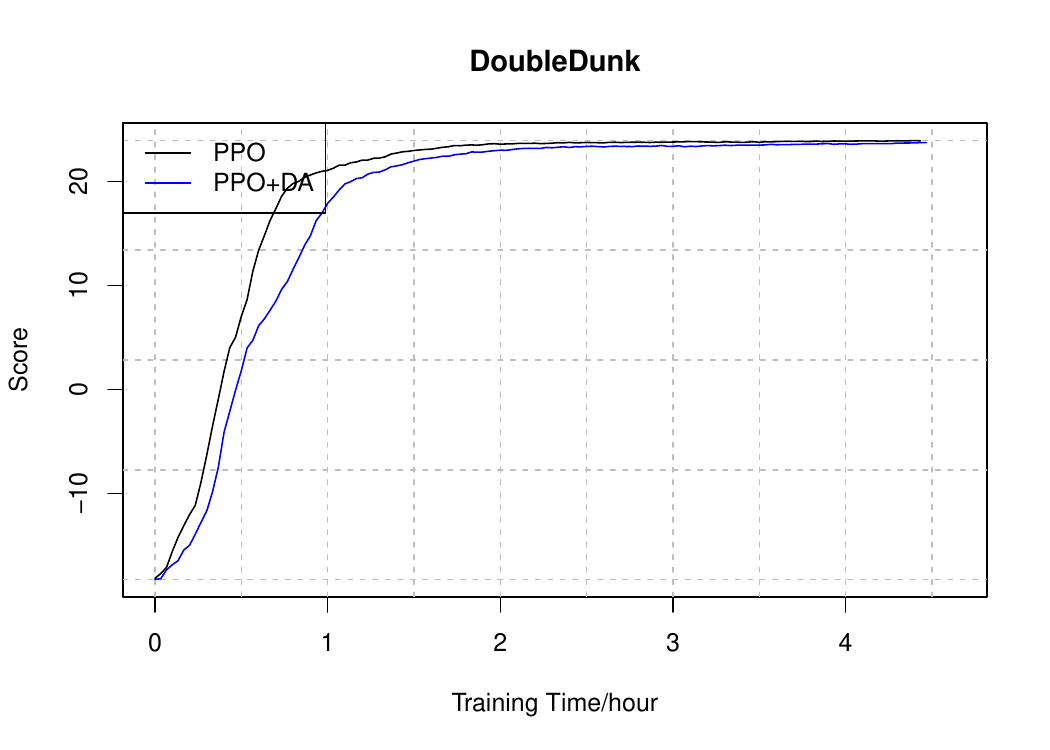}
\includegraphics[width=0.245 \textwidth]{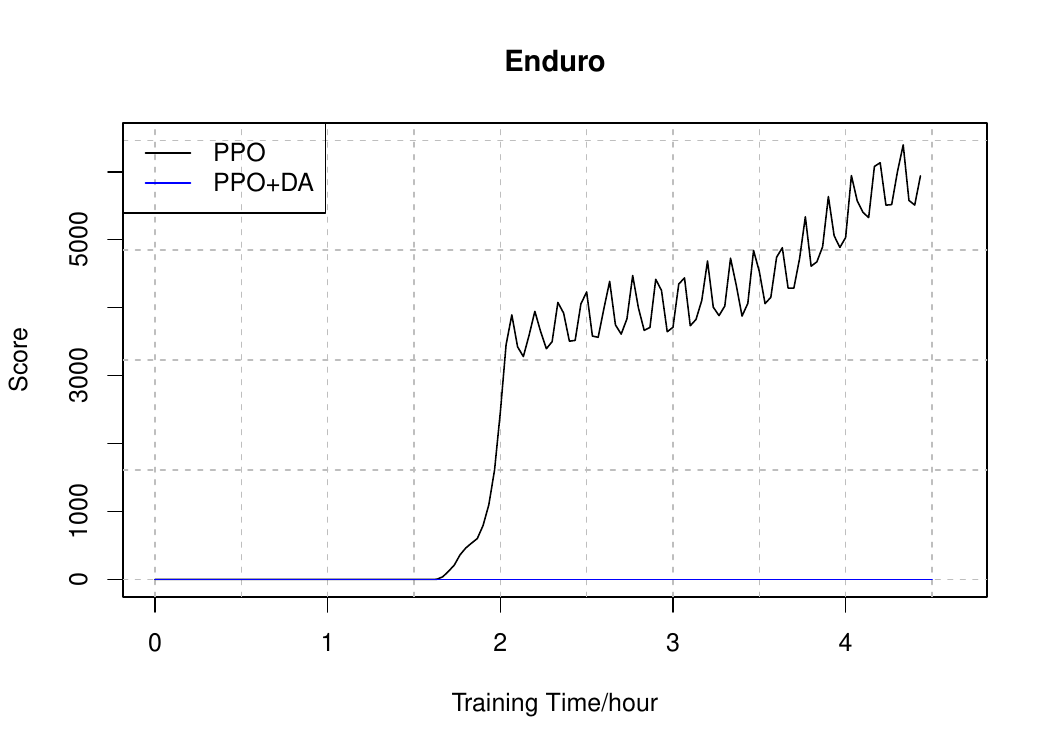}
\includegraphics[width=0.245 \textwidth]{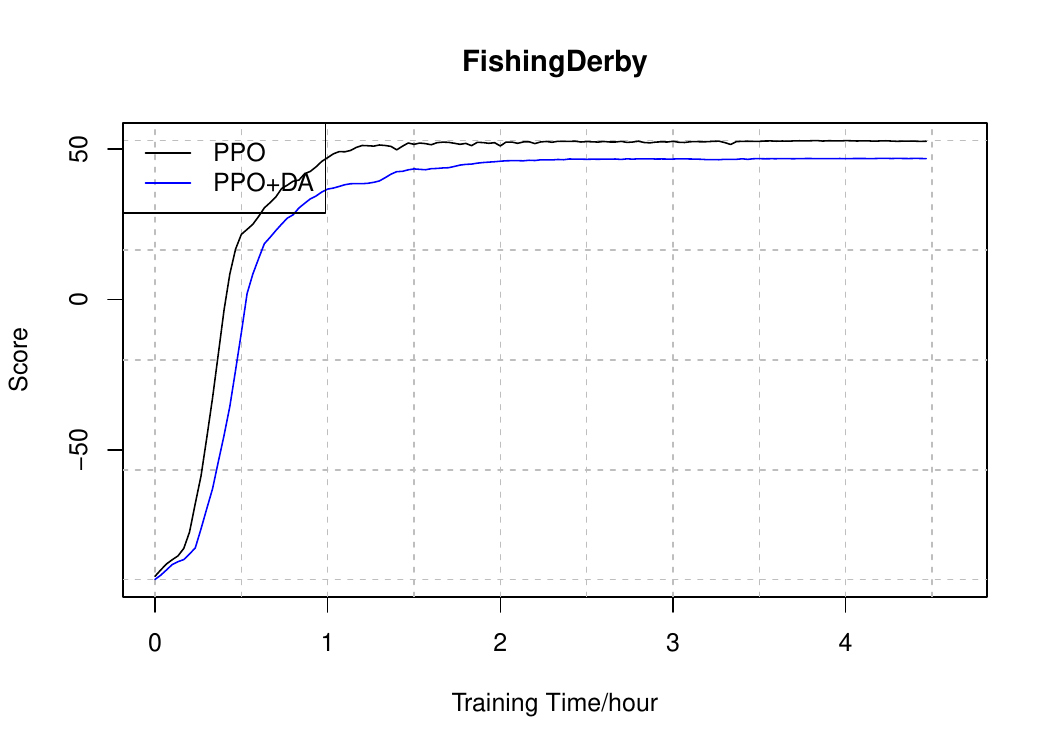}
\includegraphics[width=0.245 \textwidth]{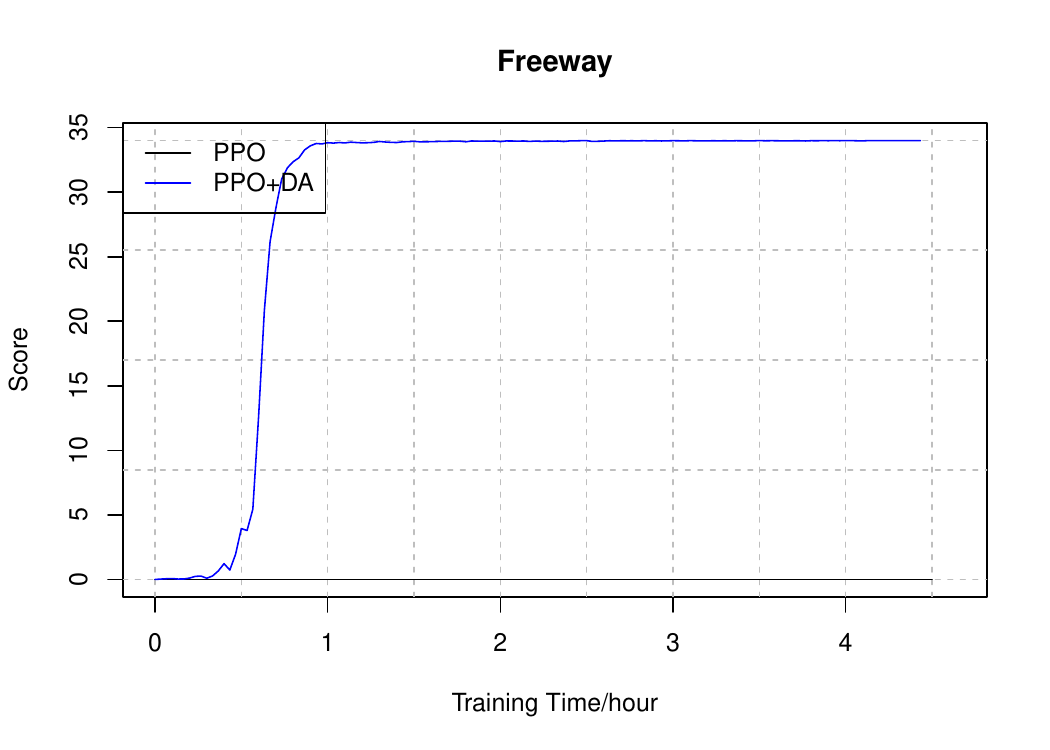}
\includegraphics[width=0.245 \textwidth]{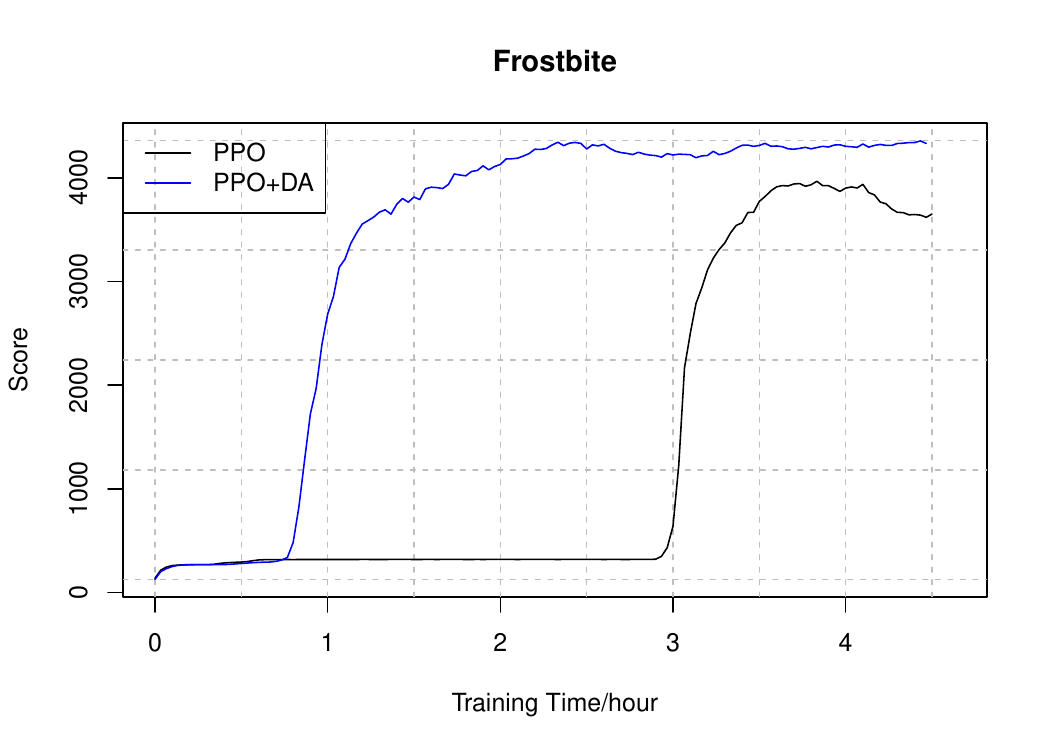}
\includegraphics[width=0.245 \textwidth]{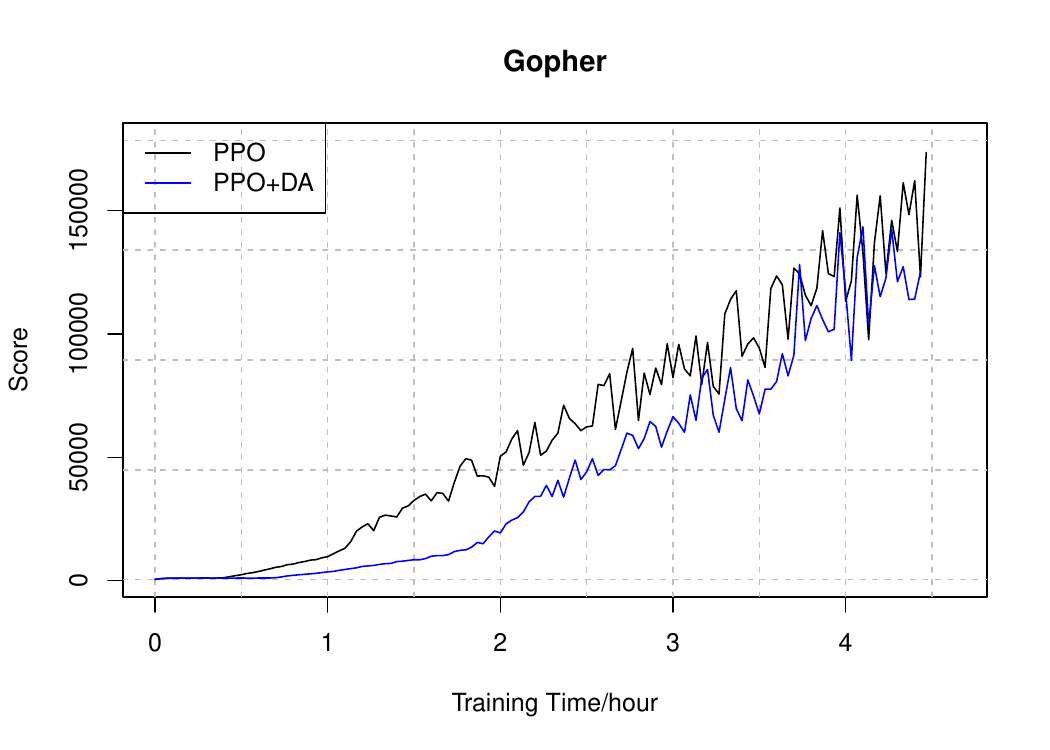}
\includegraphics[width=0.245 \textwidth]{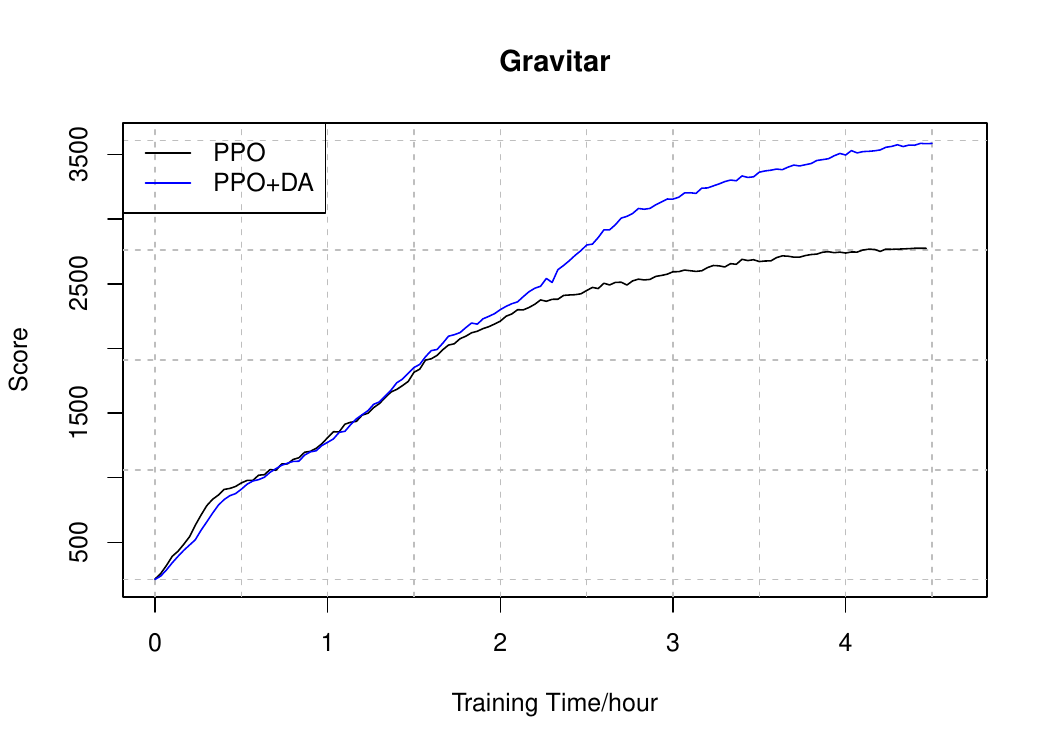}
\includegraphics[width=0.245 \textwidth]{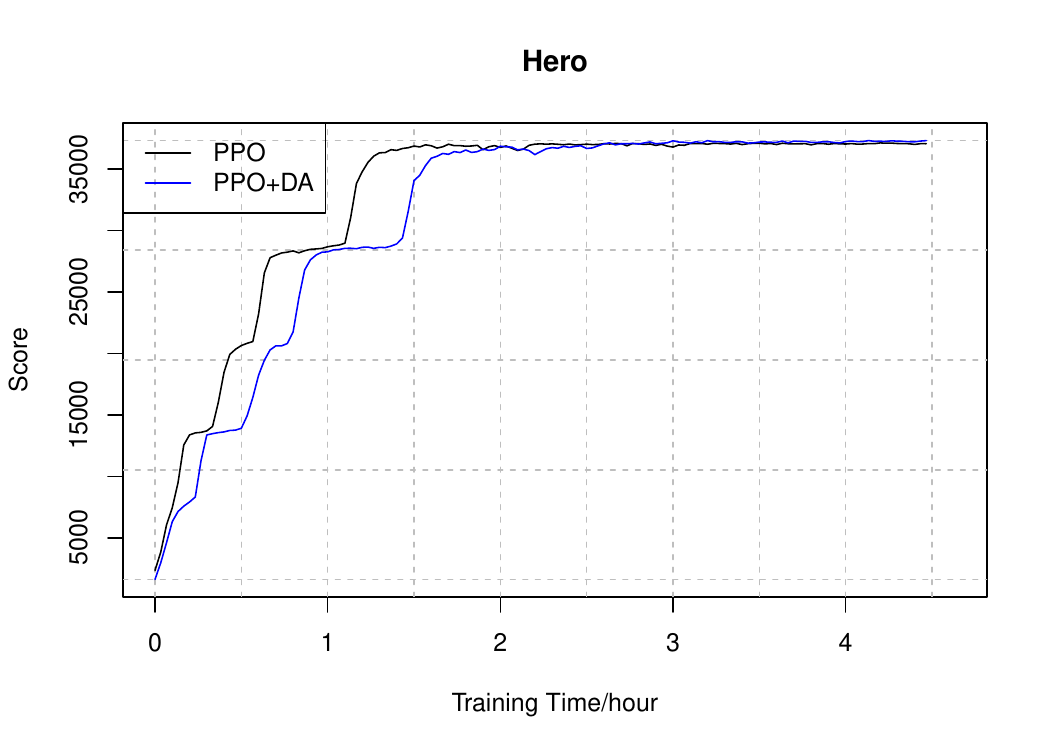}
\includegraphics[width=0.245 \textwidth]{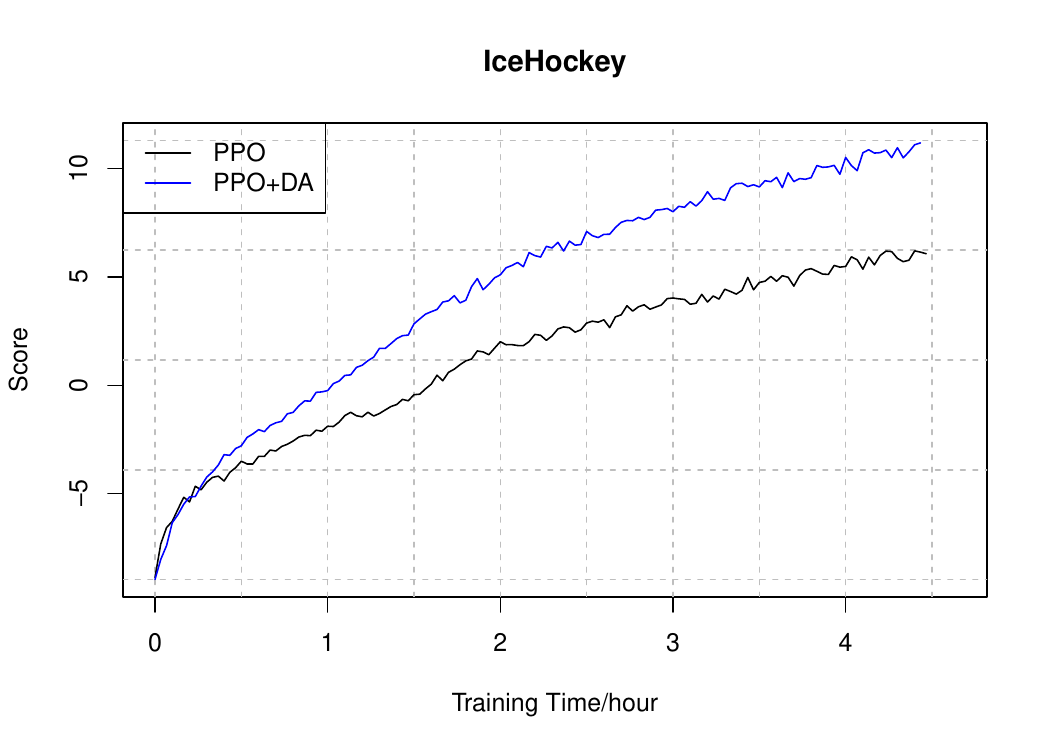}
\includegraphics[width=0.245 \textwidth]{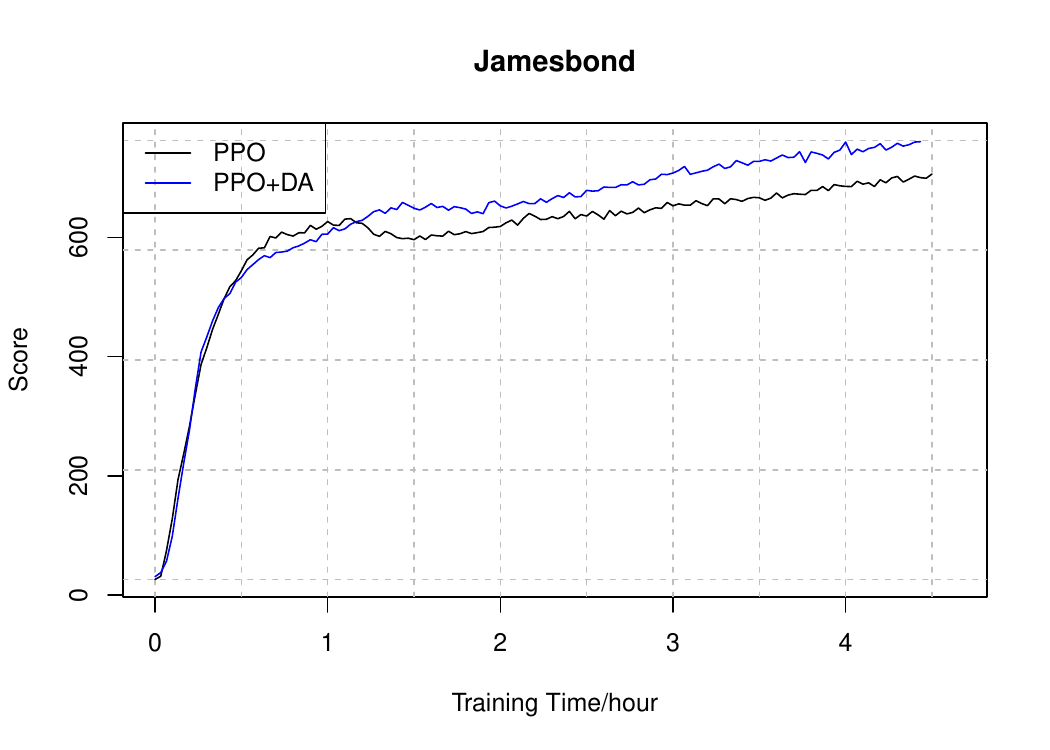}
\includegraphics[width=0.245 \textwidth]{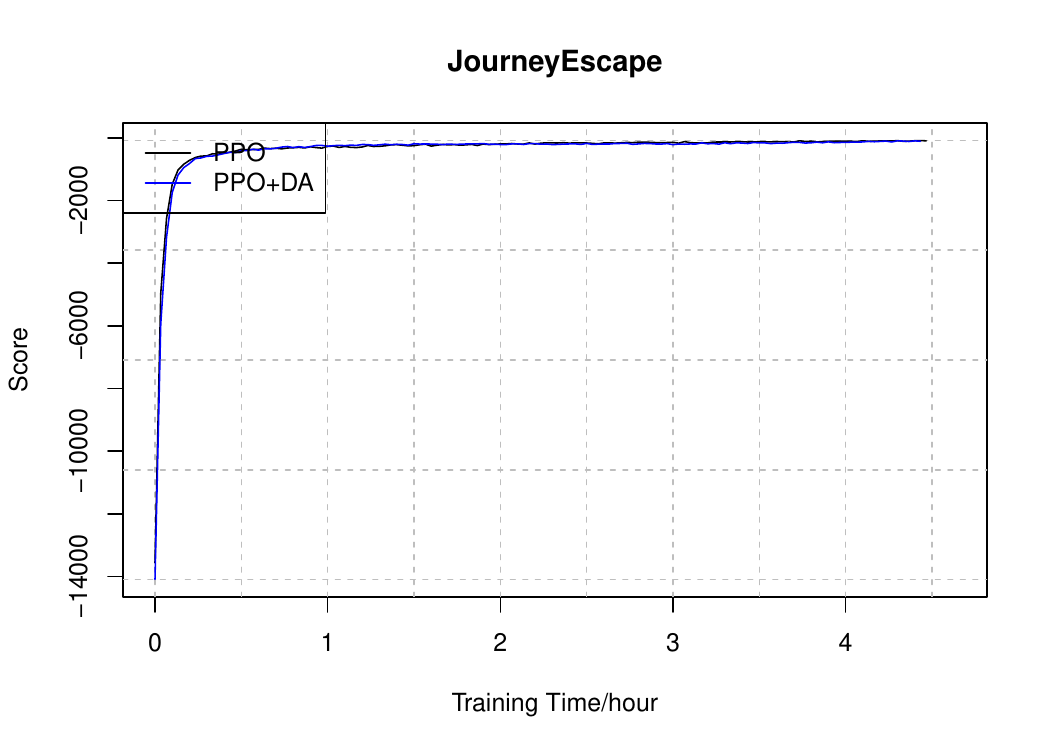}
\includegraphics[width=0.245 \textwidth]{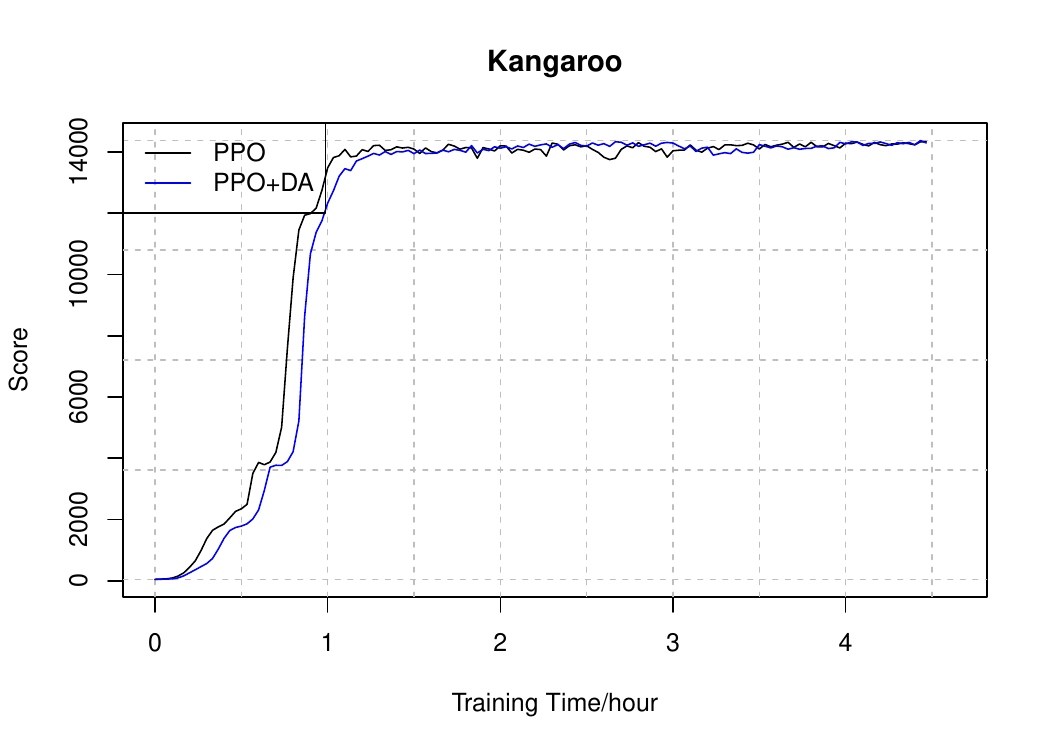}
\includegraphics[width=0.245 \textwidth]{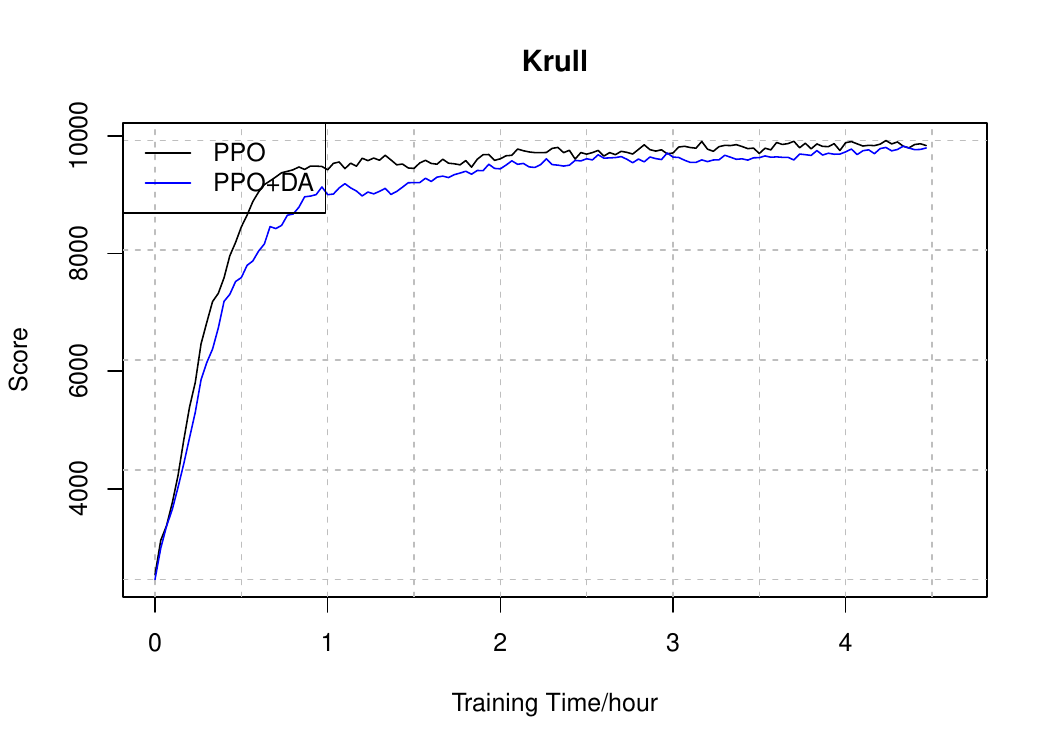}
\end{figure*}

\begin{figure*}[!t]
\includegraphics[width=0.245 \textwidth]{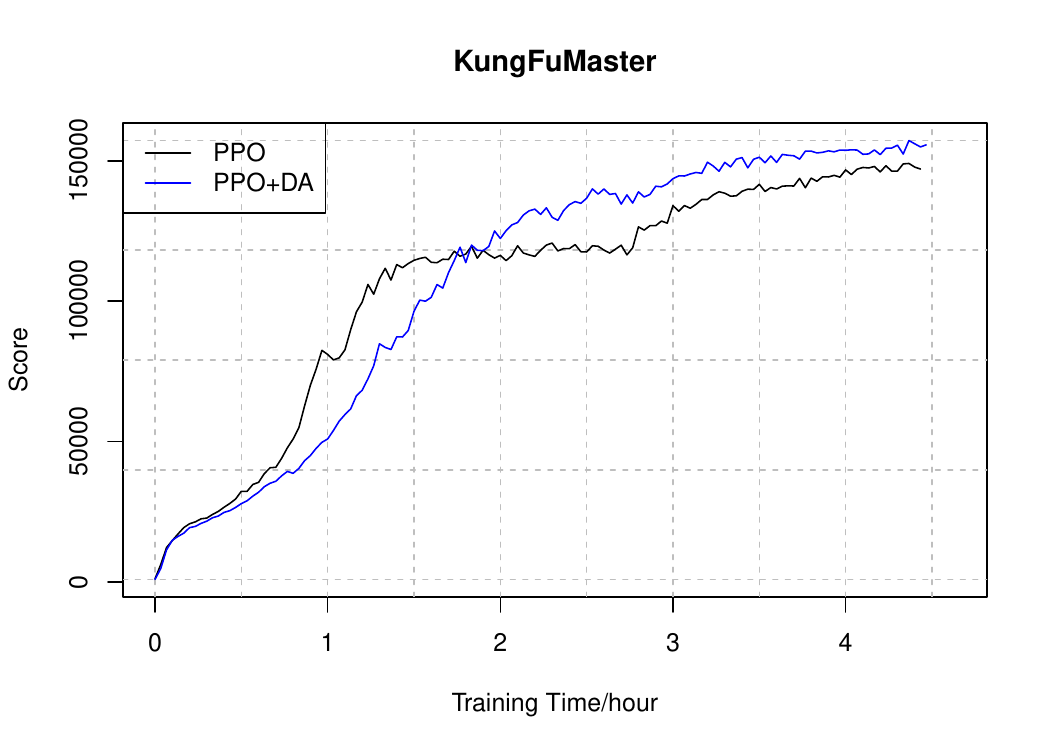}
\includegraphics[width=0.245 \textwidth]{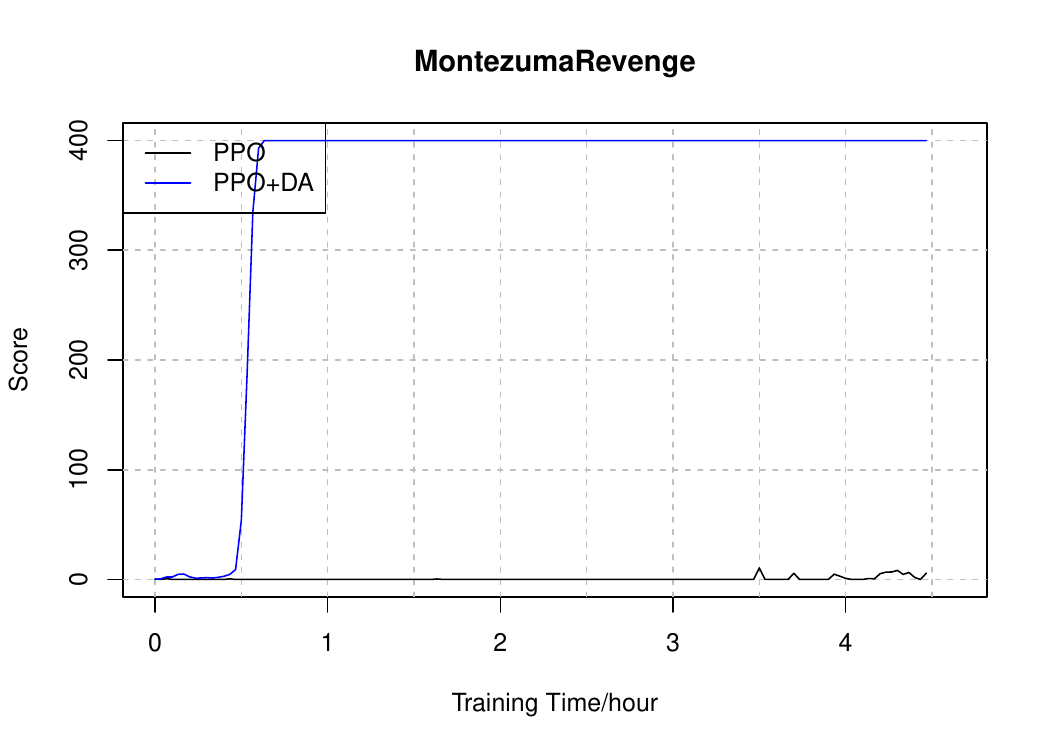}
\includegraphics[width=0.245 \textwidth]{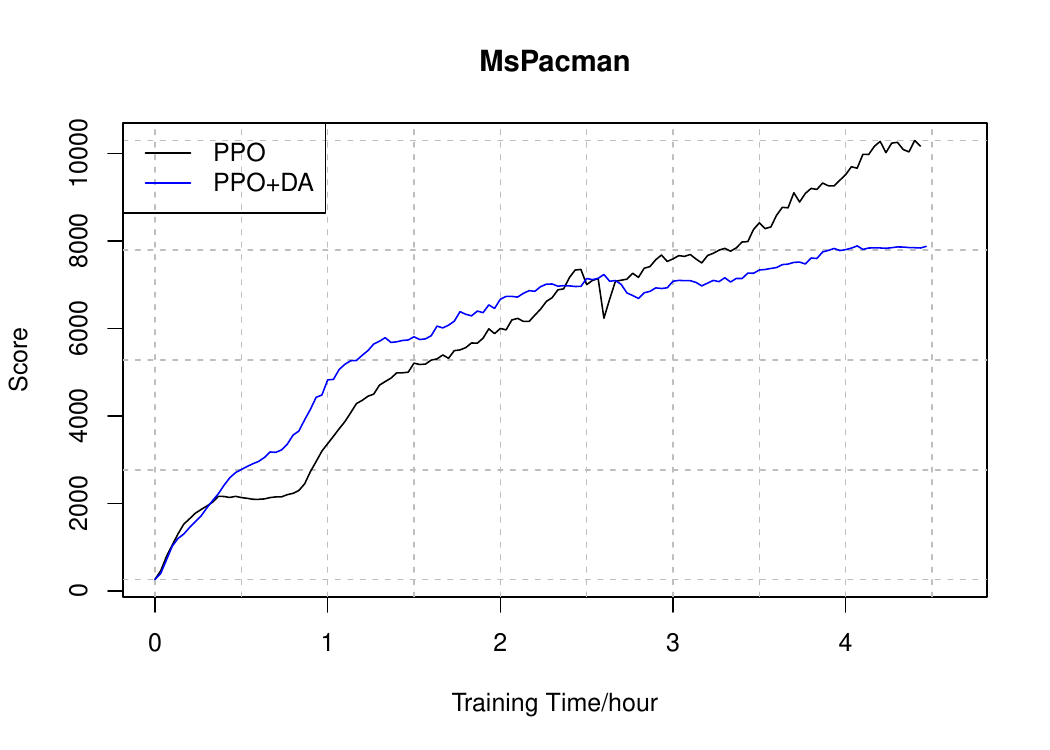}
\includegraphics[width=0.245 \textwidth]{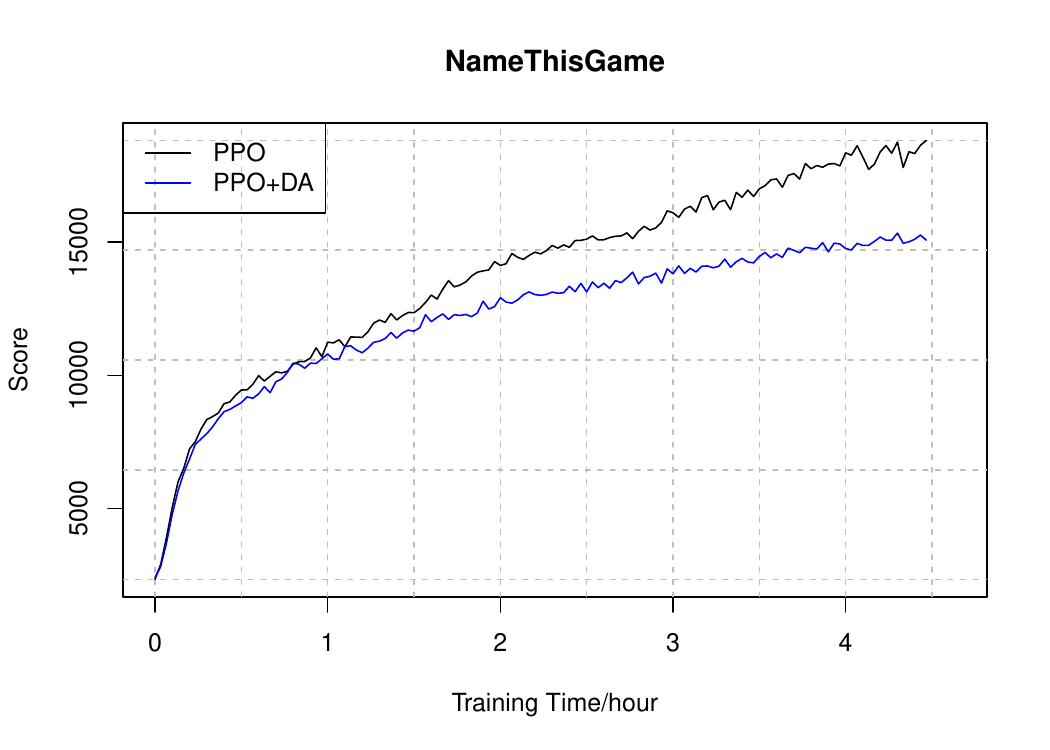}
\includegraphics[width=0.245 \textwidth]{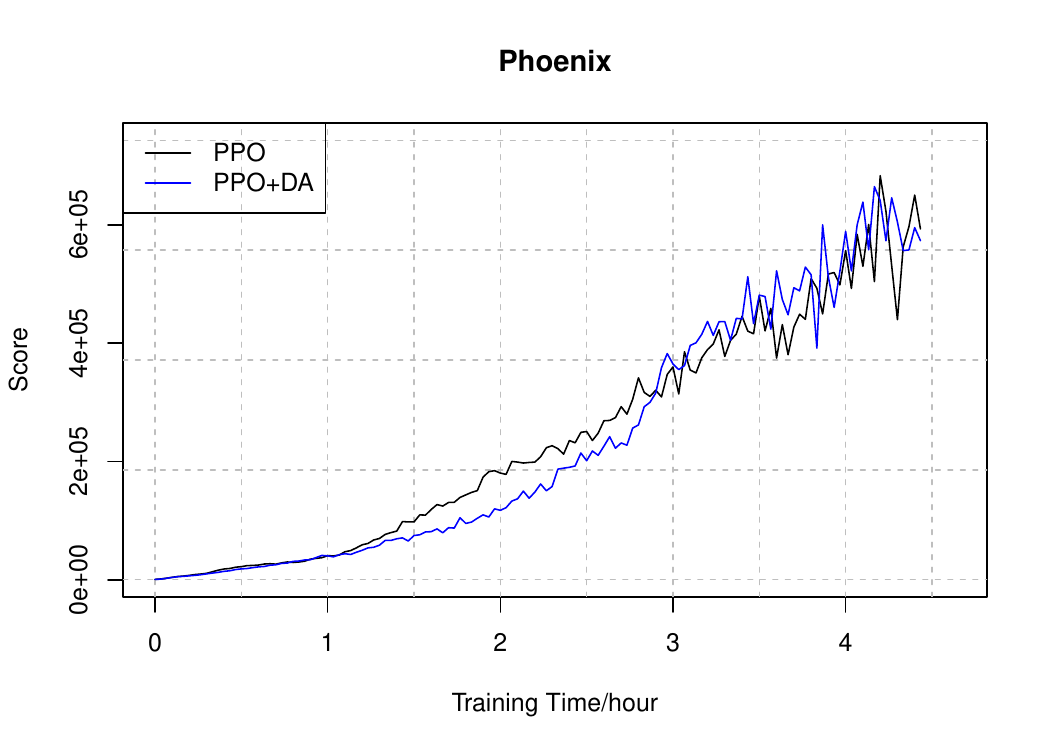}
\includegraphics[width=0.245 \textwidth]{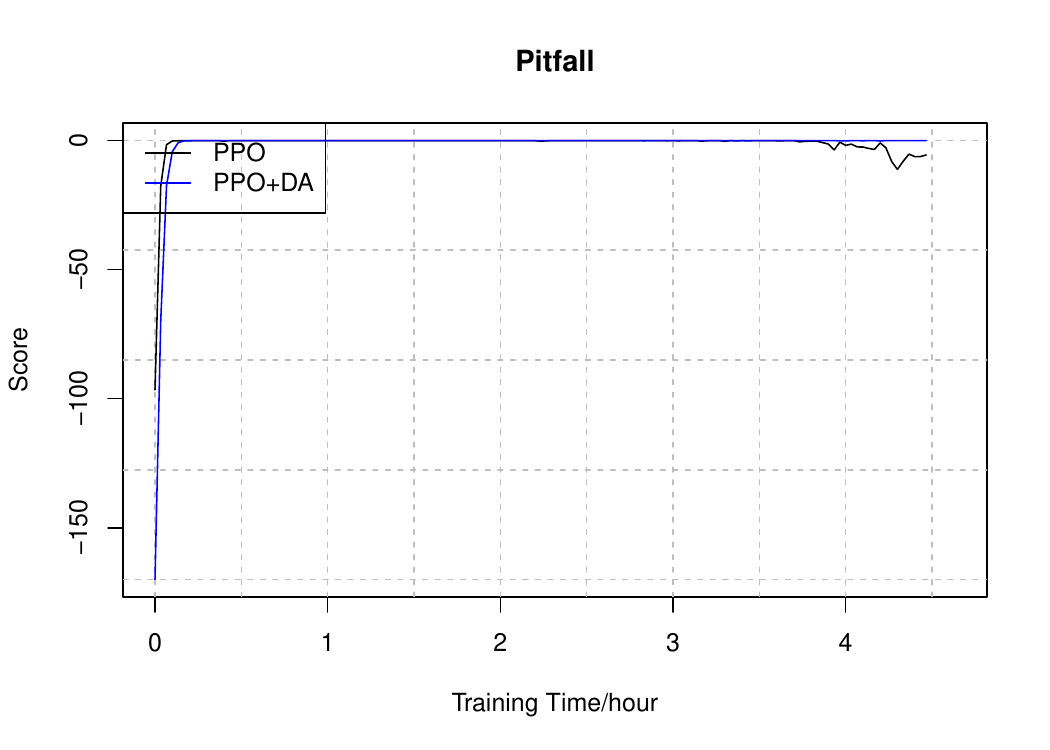}
\includegraphics[width=0.245 \textwidth]{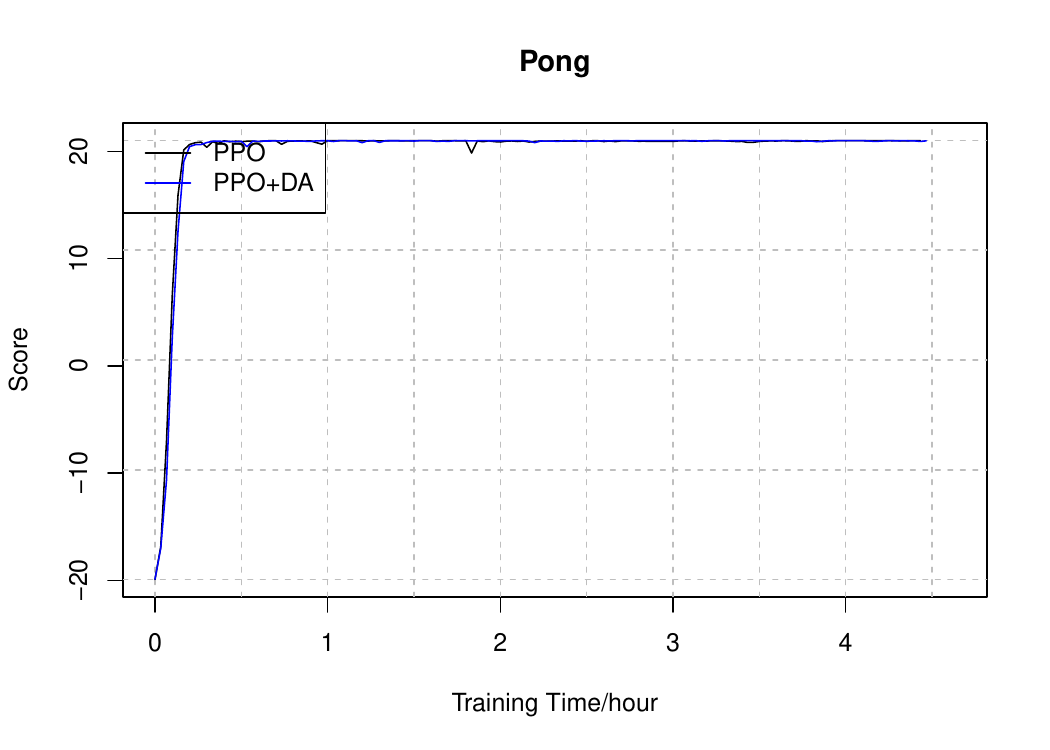}
\includegraphics[width=0.245 \textwidth]{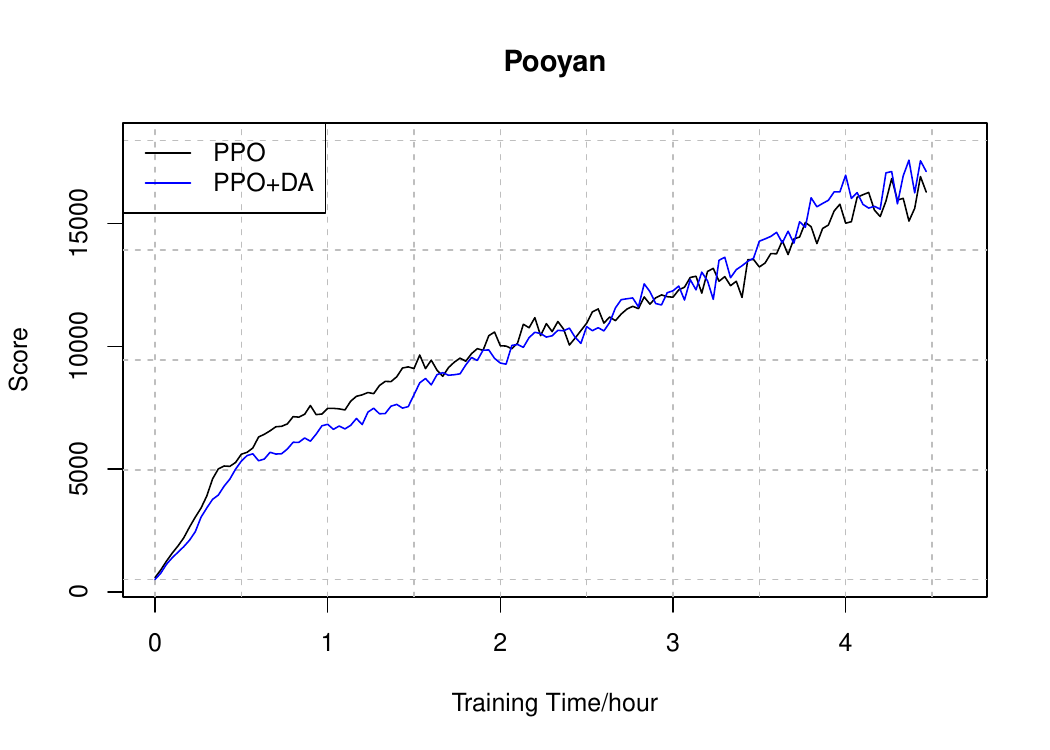}
\includegraphics[width=0.245 \textwidth]{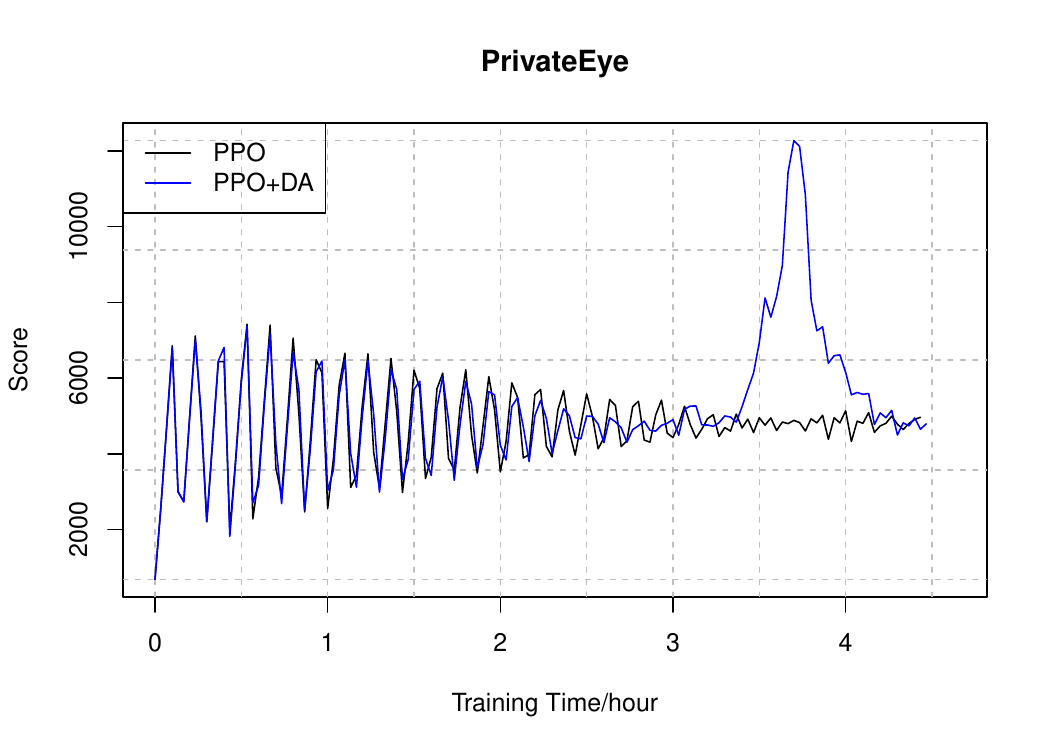}
\includegraphics[width=0.245 \textwidth]{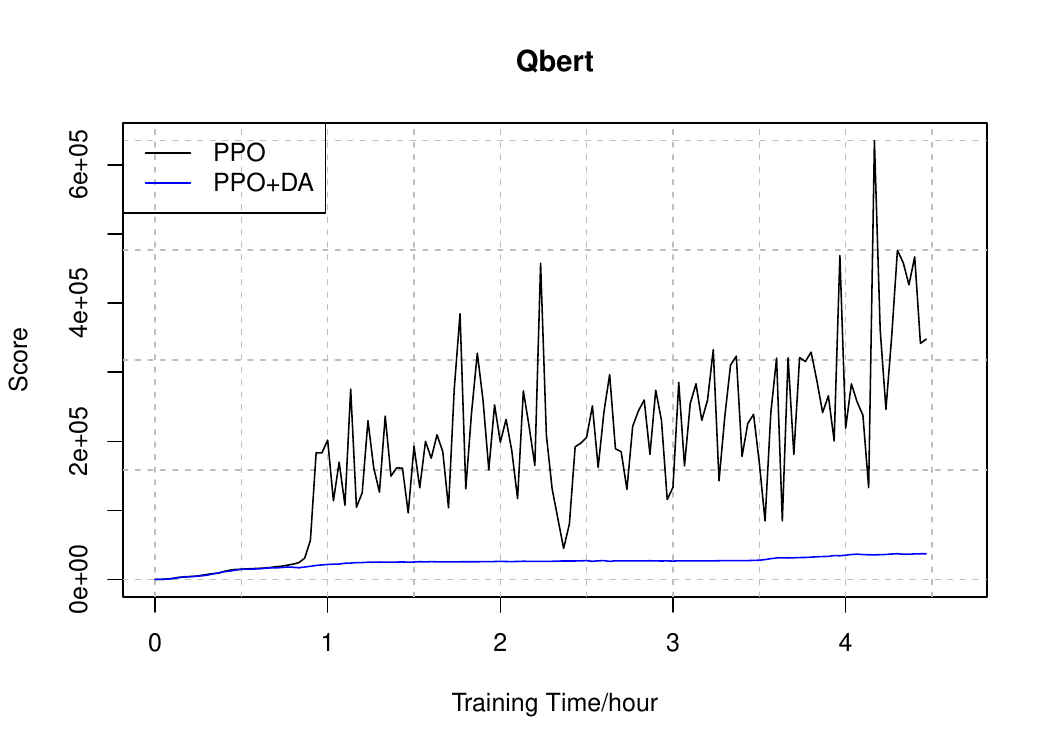}
\includegraphics[width=0.245 \textwidth]{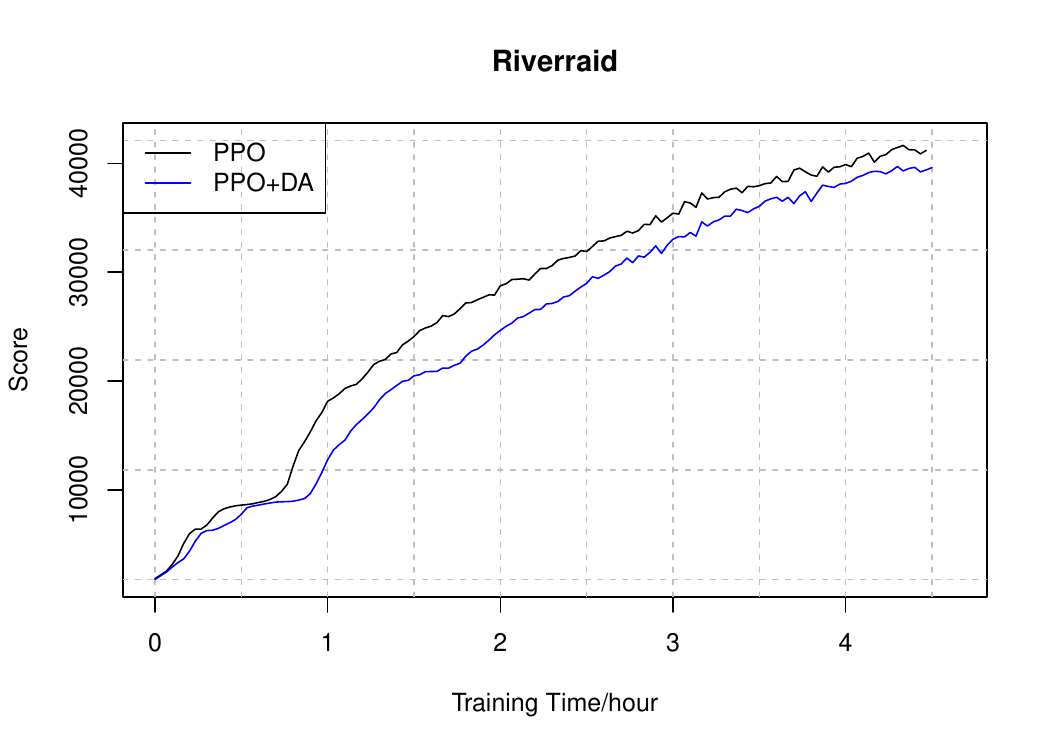}
\includegraphics[width=0.245 \textwidth]{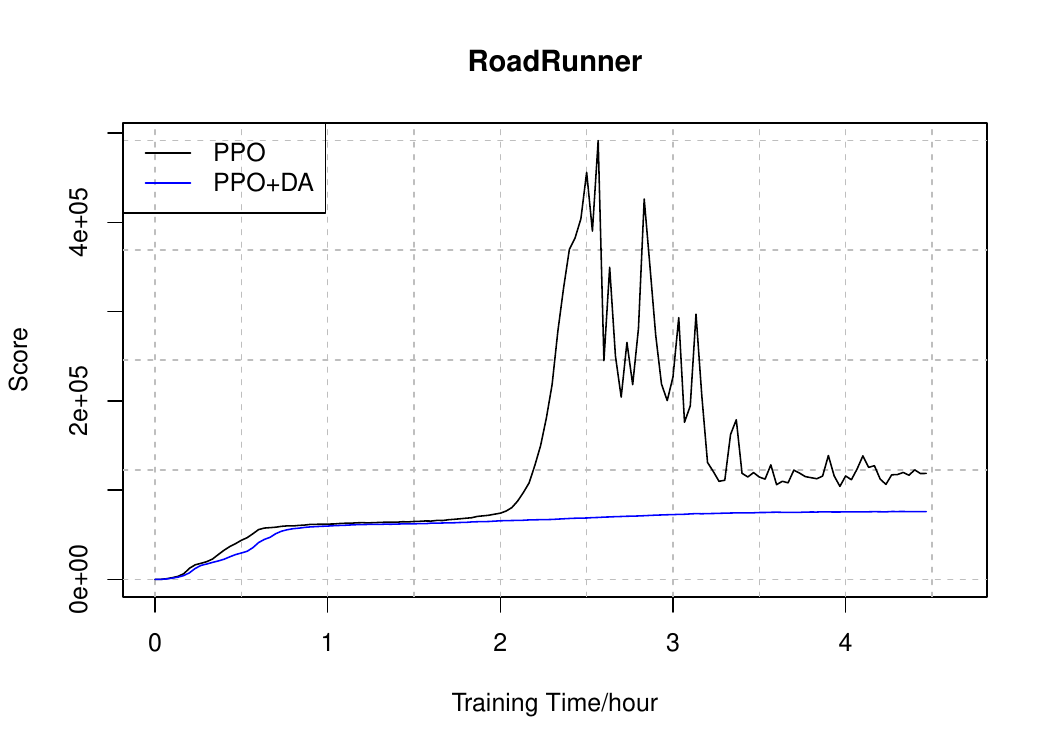}
\includegraphics[width=0.245 \textwidth]{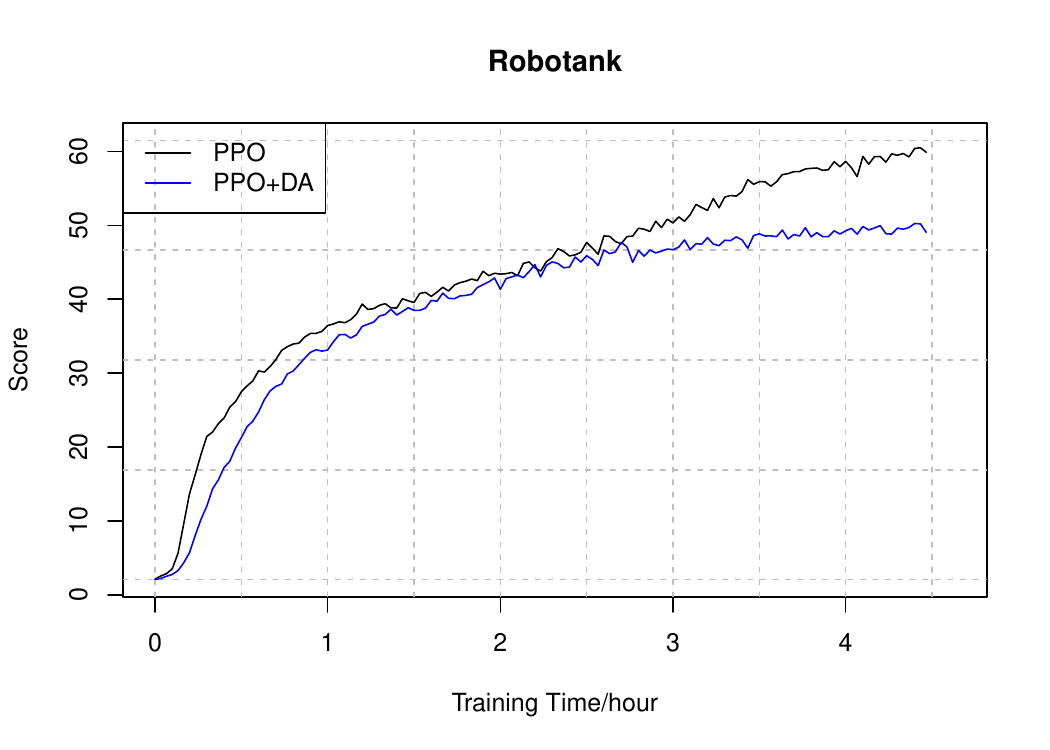}
\includegraphics[width=0.245 \textwidth]{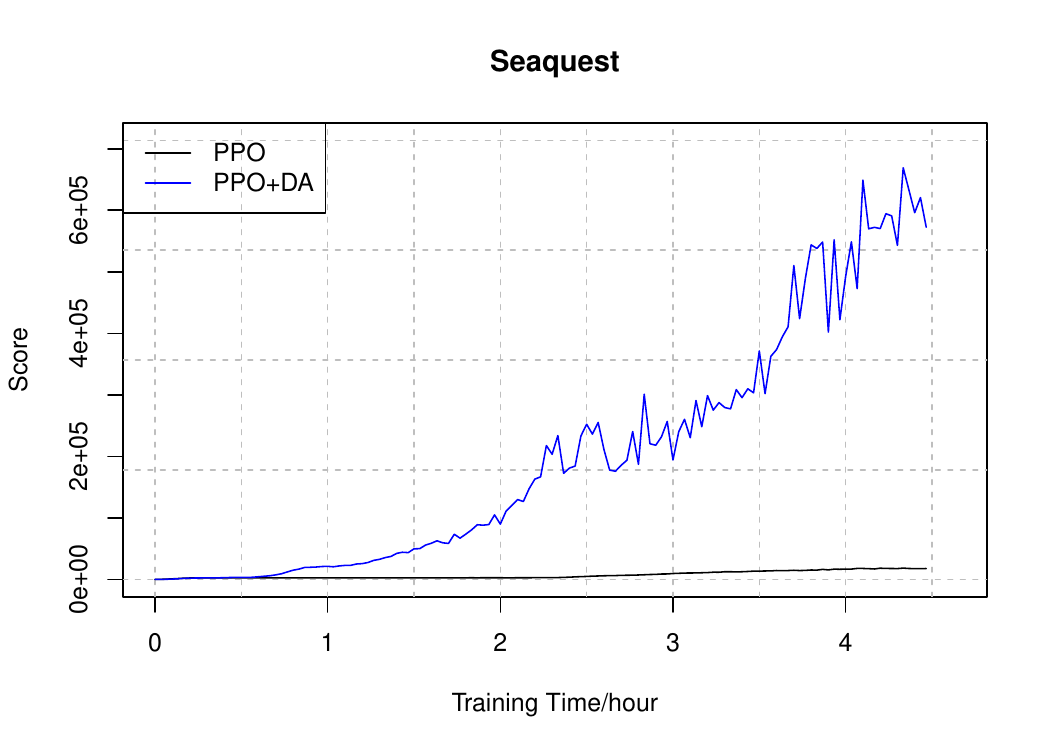}
\includegraphics[width=0.245 \textwidth]{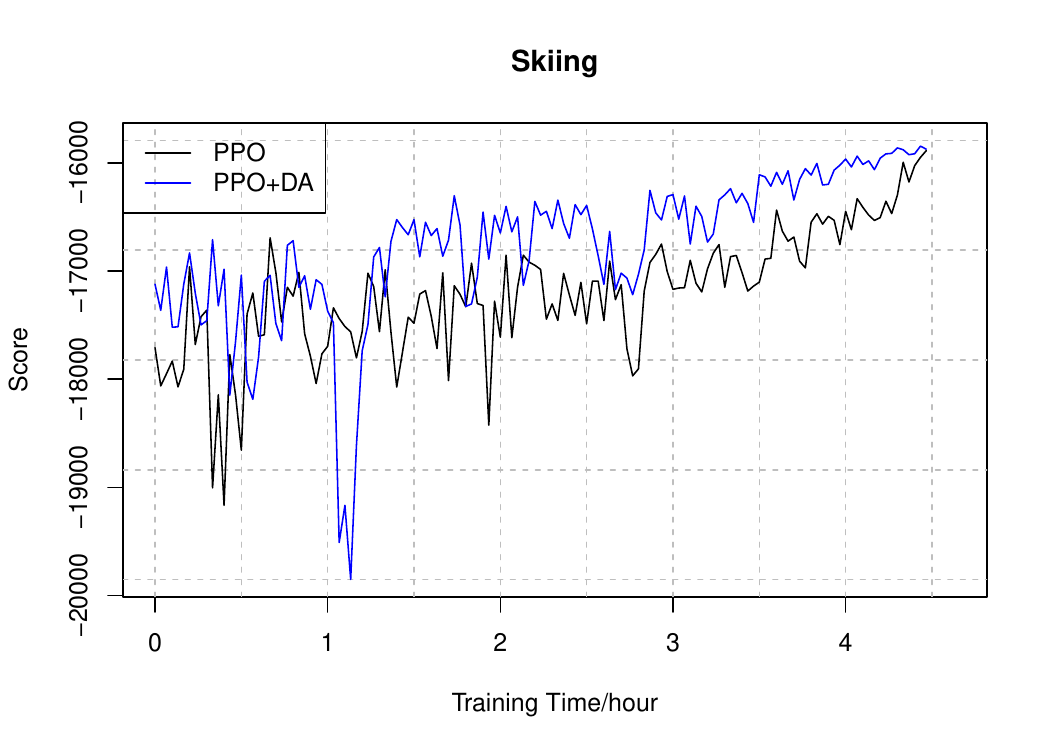}
\includegraphics[width=0.245 \textwidth]{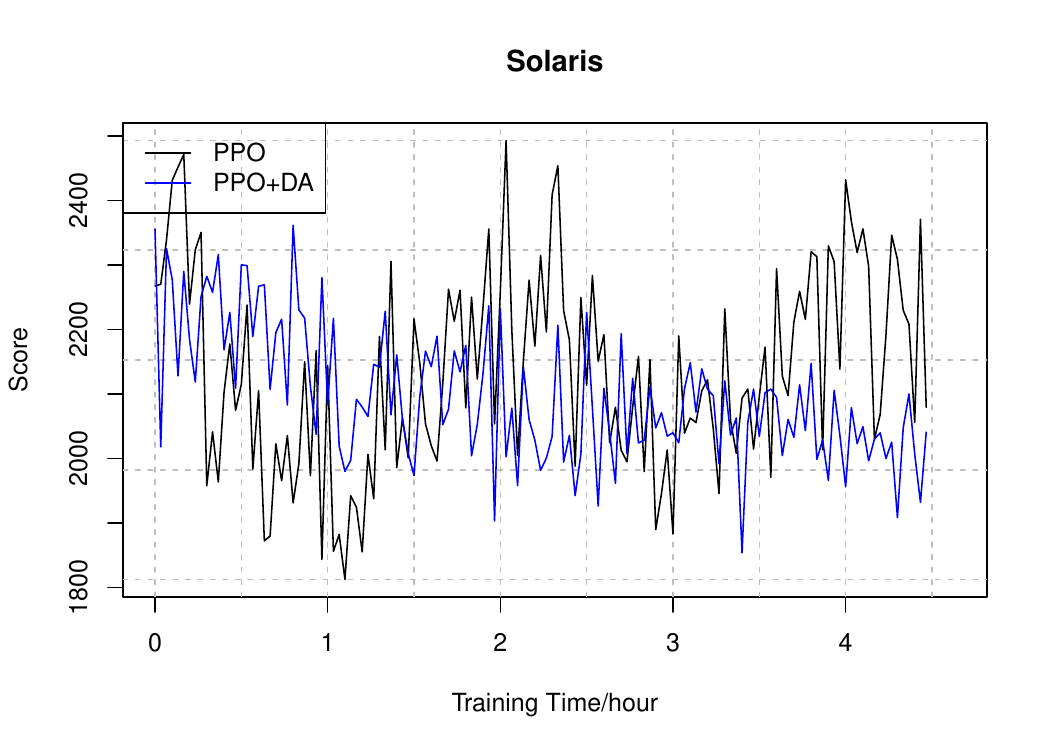}
\includegraphics[width=0.245 \textwidth]{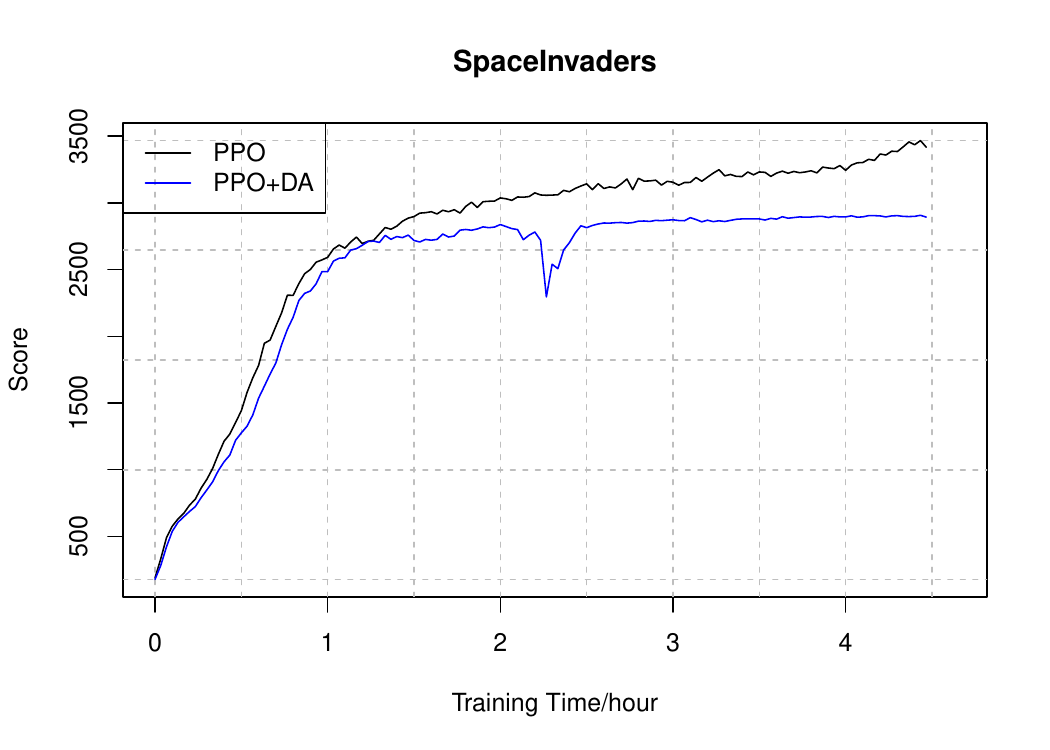}
\includegraphics[width=0.245 \textwidth]{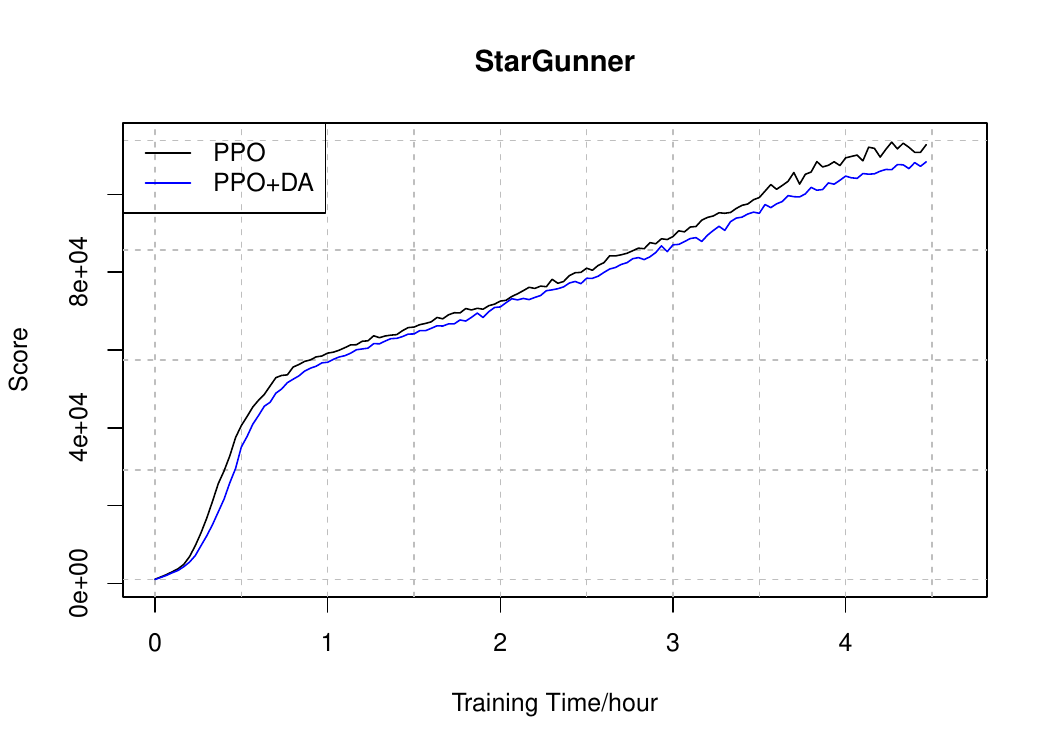}
\includegraphics[width=0.245 \textwidth]{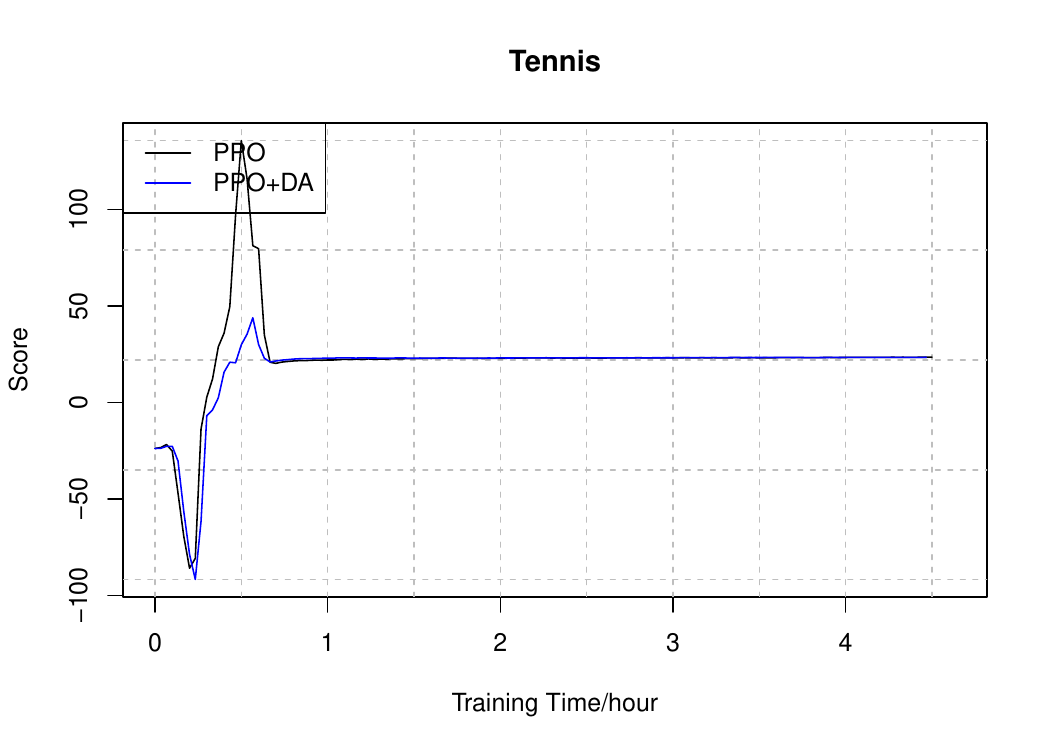}
\includegraphics[width=0.245 \textwidth]{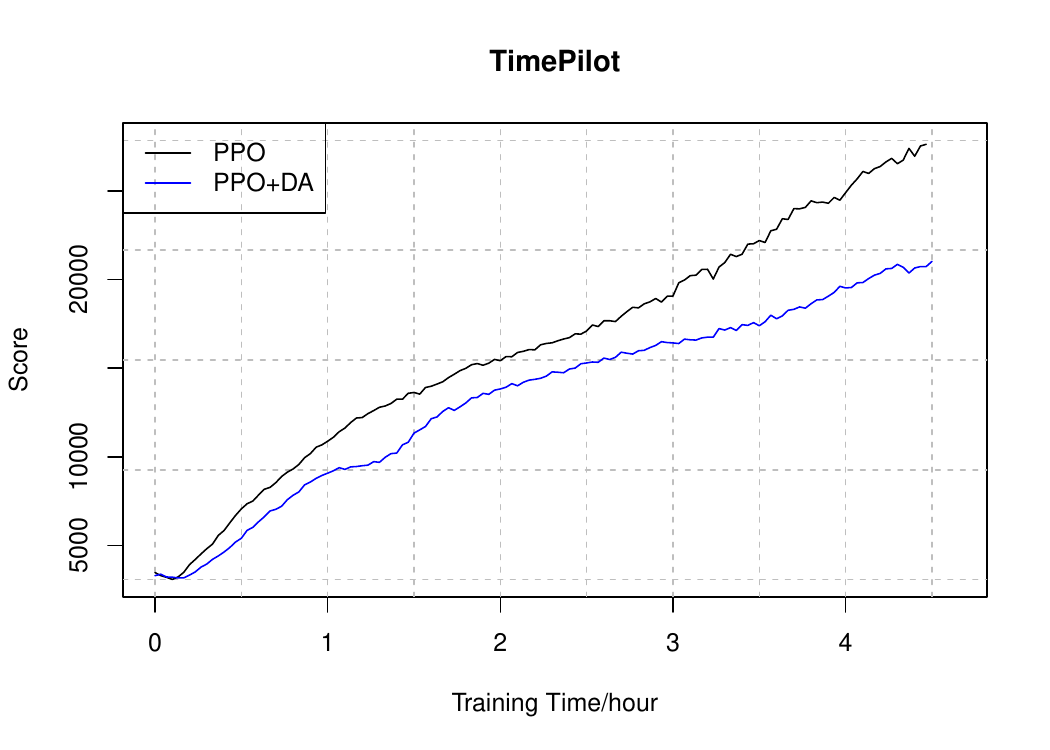}
\includegraphics[width=0.245 \textwidth]{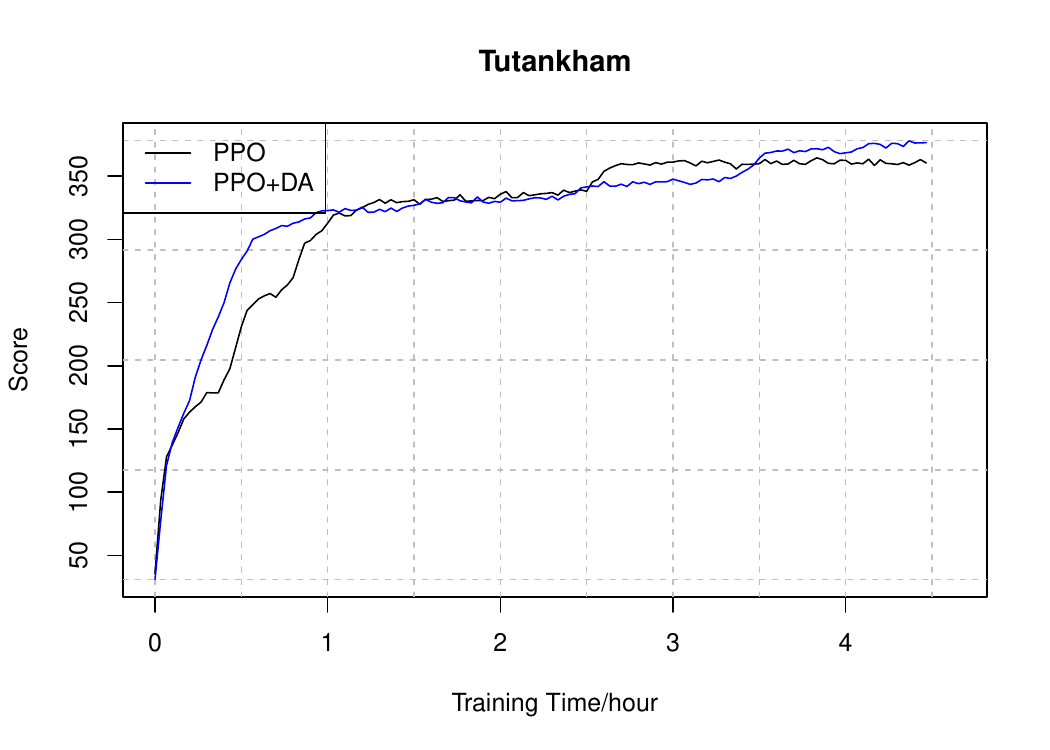}
\includegraphics[width=0.245 \textwidth]{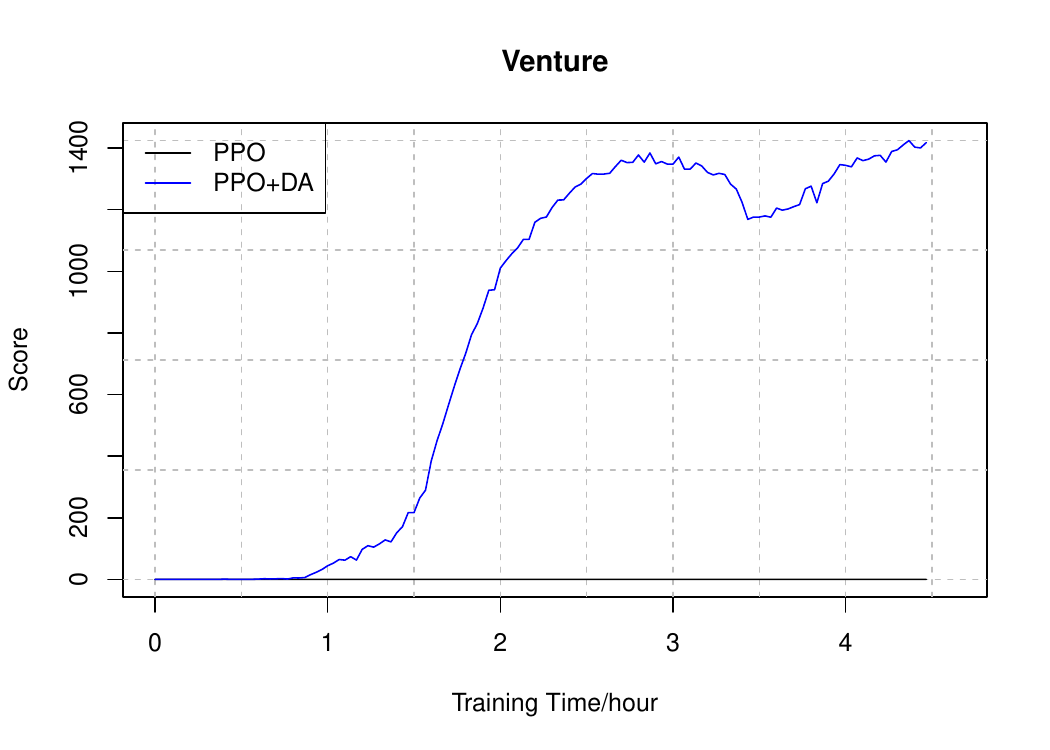}
\includegraphics[width=0.245 \textwidth]{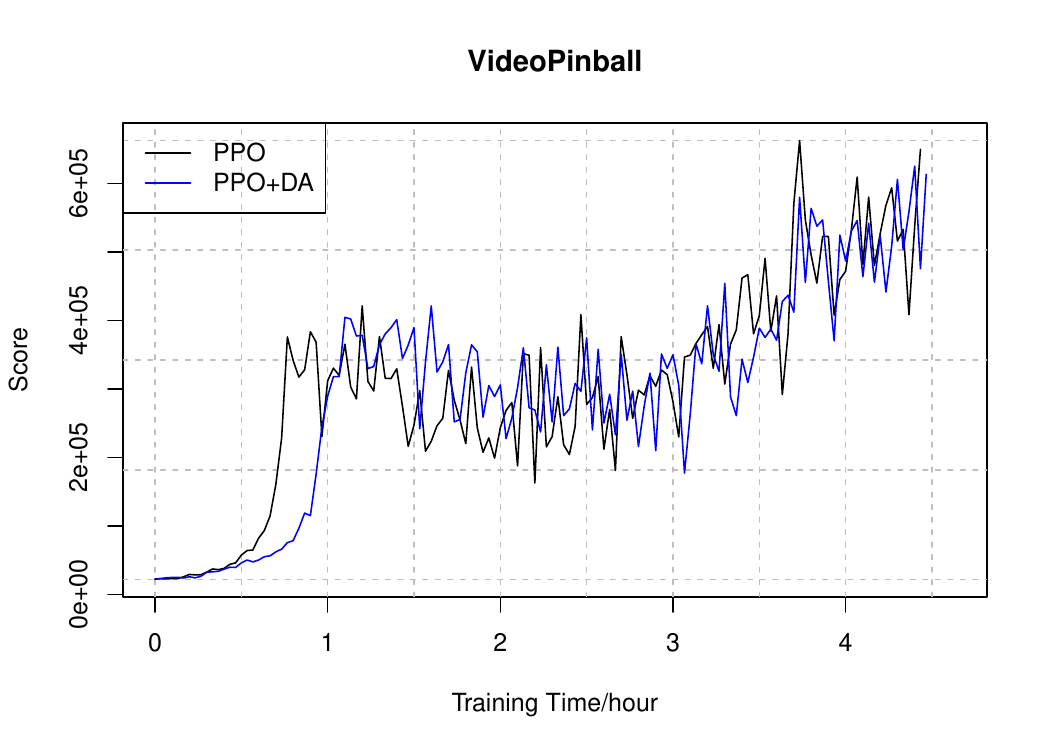}
\includegraphics[width=0.245 \textwidth]{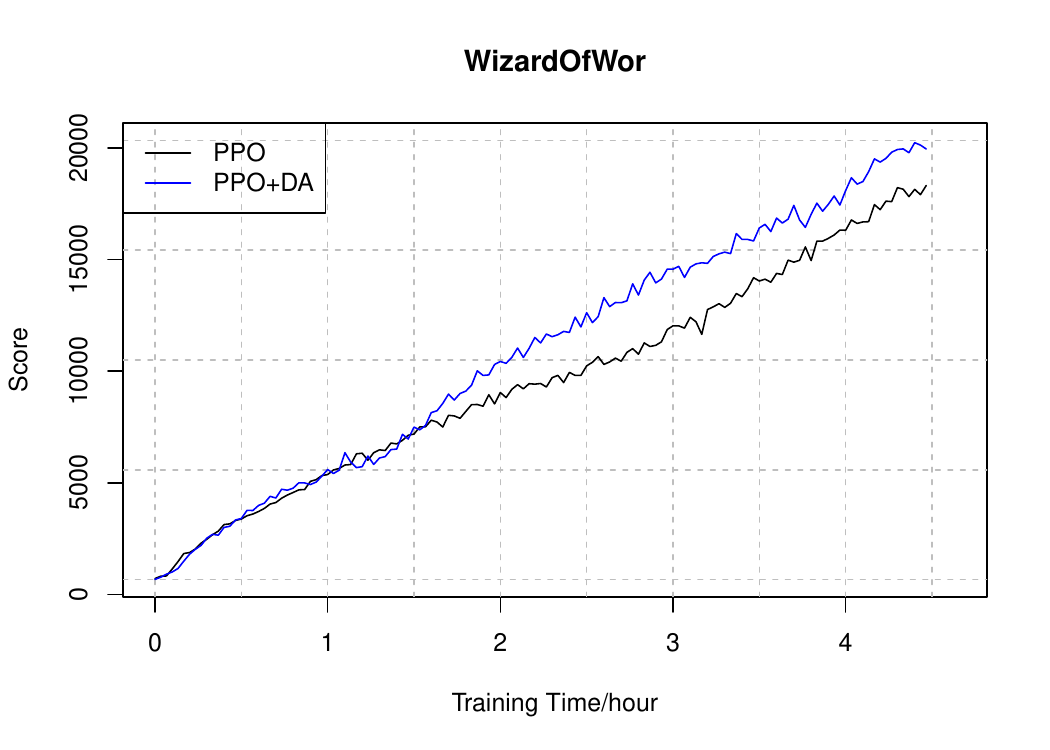}
\includegraphics[width=0.245 \textwidth]{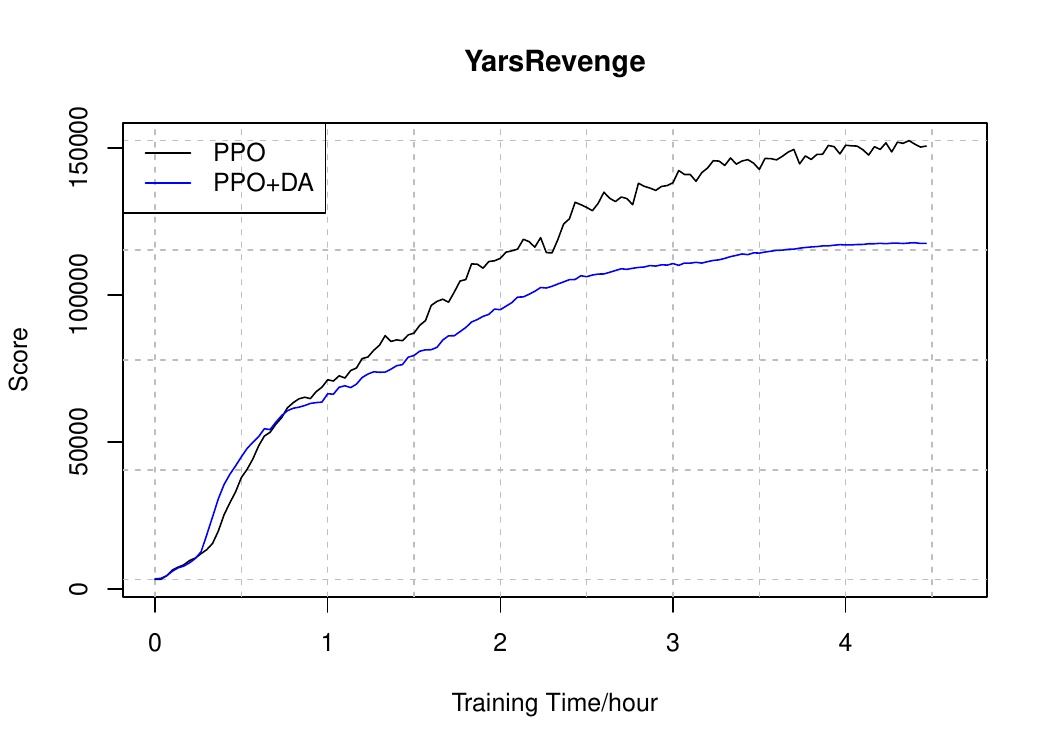}
\includegraphics[width=0.245 \textwidth]{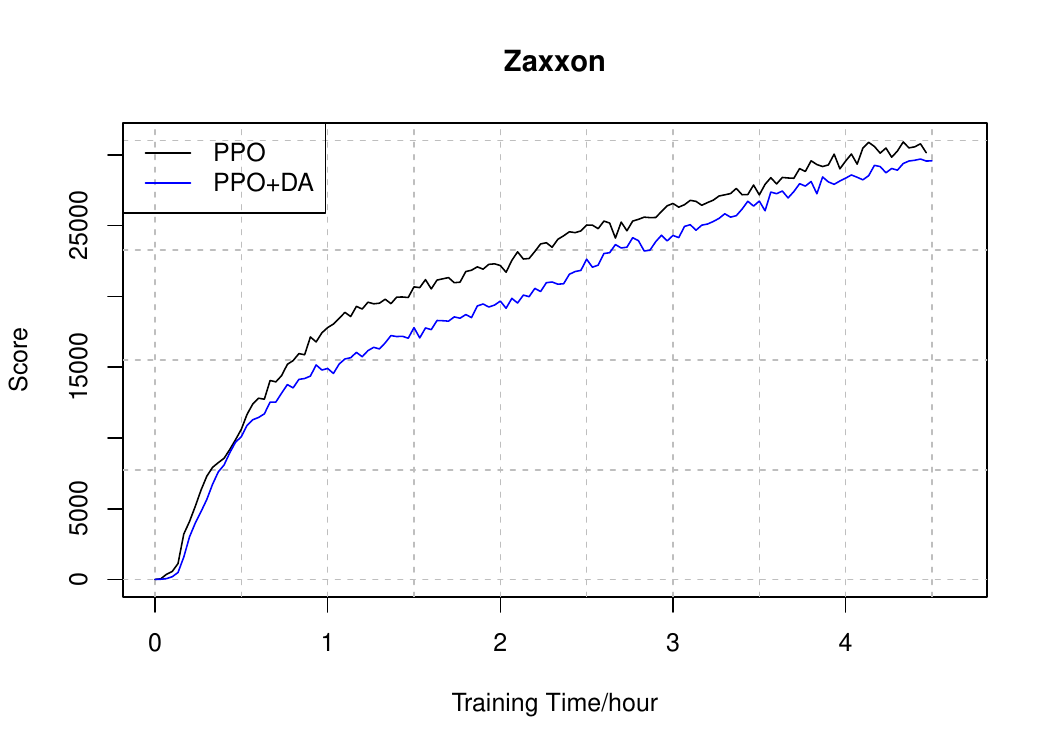}
  \caption{\red{Performance comparison of PPO+DA with PPO on 58 Atari games, with the number of used actors increased to 64 and running time increased to 4 hours.} }
  \label{fig:compare_a64}
\end{figure*}

\end{document}